\documentclass[times, review, 10pt]{elsarticle}

\usepackage{amssymb}
\usepackage{amsmath}

\usepackage{enumerate} 
\numberwithin{equation}{section} 
\usepackage{array} 
\graphicspath{{figures/Fig. 1/},{figures/Fig. 2/},{figures/Fig. 3/},{figures/Fig. 4/},{figures/Fig. 5/},{figures/Fig. 6/},{figures/Fig. 7/},{figures/Fig. 8/},{figures/Fig. 9/},{figures/Fig. 10/},{figures/Fig. 11/},{figures/Fig. 12/},{figures/Fig. 13/},{figures/Fig. 14/},{figures/Fig. 15/}} 

\usepackage{amsmath}
\usepackage{algorithm}
\usepackage{algorithmic}
\usepackage{booktabs}
\usepackage{subfigure}
\usepackage{multirow}
\usepackage{hyperref}

\usepackage[utf8]{inputenc}
\usepackage[T1]{fontenc}
\usepackage{lmodern}

\biboptions{sort&compress}

\usepackage{geometry}
\usepackage{hyperref}
\usepackage{xcolor}

\journal{arXiv}
\begin{document}
	
	\begin{frontmatter}

\title{Gaussian highpass guided image filtering}

 \author{Lei Zhao\fnref{}}
 \ead{202206021009@stu.cqu.edu.cn}
 
 \author{Chuanjiang He\corref{mycorrespondingauthor}}
 \cortext[mycorrespondingauthor]{Corresponding author}
 \ead{chuanjianghe@sina.com}
 
 \address{College of Mathematics and Statistics, Chongqing University, Chongqing, 401331, China}

\begin{abstract}
Guided image filtering (GIF) is a popular smoothing technique, in which an additional image is used as a structure guidance for noise removal with edge preservation. The original GIF and some of its subsequent improvements are derived from a two-parameter local affine model (LAM), where the filtering output is a local affine transformation of the guidance image, but the input image is not taken into account in the LAM formulation. In this paper, we first introduce a single-parameter Prior Model based on Gaussian (highpass/lowpass) Filtering (PM-GF), in which the filtering output is the sum of a weighted portion of Gaussian highpass filtering of the guidance image and Gaussian smoothing of the input image. In the PM-GF, the guidance structure determined by Gaussian highpass filtering is obviously transferred to the filtering output, thereby better revealing the structure transfer mechanism of guided filtering. Then we propose several Gaussian highpass GIFs (GH-GIFs) based on the PM-GF by emulating the original GIF and some improvements, i.e., using PM-GF instead of LAM in these GIFs. Experimental results illustrate that the proposed GIFs outperform their counterparts in several image processing applications.
\end{abstract}



\end{frontmatter}



\newpage

\section{Introduction}

Guided image filtering  \cite{he2012guided} is a popular smoothing technique, designed for smoothing image while preserving primary structures in the image under consideration. The basic idea  is
that an additional image (guidance image) is utilized as a structure guidance  to remove noise  while preserving edges  in the input image.
Up to now, the guided filtering has attracted more and more attention in the field of image processing, and has gained widespread application due to its low algorithmic complexity and superior performance. In fact, in image processing, computer vision and graphics, edge-preserving smoothing techniques including the guided filtering have been applied across a variety of tasks  \cite{goyal2020image}, e.g., image enhancement \cite{farbman2008edge,li2022adaptive}, tone mapping for high dynamic range (HDR) images \cite{durand2002fast,liang2018a}, image dehazing \cite{li2017single},noise removal\cite{he2022non}, signal smoothing\cite{ran2024iterative}, image texture removal smoothing \cite{zhang2014rolling,liu2022a}, image fusion\cite{ren2021infrared} \cite{li2013image} and joint upsampling \cite{kopf2007joint}.

In \cite{he2012guided}, He et al. first proposed a guided image filter (GIF), built on a prior model (local affine model, LAM). In the LAM, the filtering output at a pixel is an affine transformation of the guidance image in a local window containing the target pixel, while the affine transformations at different pixels in the local window share the identical affine coefficients. In the GIF, two local affine coefficients are estimated via linear ridge regression by taking into account the content of the guidance image and the constraint from the input image. Since the filtering output at pixel $p$, which is computed by LAM, is dependent of the center pixel in local window containing $p$, the filtering output has different values in different local windows containing the target pixel. He et al. \cite{he2012guided} adopted the arithmetic averaging strategy to address the overlapping windows problem involved in calculating the filtered output of GIF.

As stated in  \cite{he2012guided}, the GIF has naturally a fast and nonapproximate linear time algorithm, and has witnessed a series of new applications in computer vision and graphics due to its simplicity and efficiency. Despite its success, the GIF cannot preserve best sharp edges, thereby leading to halo artifacts caused by edge blurring in many applications. 
Two main reasons for this limitation are to utilize the constant regularization parameter for the cost function based on ridge regression and adopt the arithmetic averaging strategy to address the problem of overlapping windows involved in computing the filtered output. To address these issues, some subsequent improvements of the GIF have documented in the literature; see for example \cite{li2014weighted,kou2015gradient,zhang2022robust,sun2019weighted,ochotorena2019anisotropic,yin2019side,yuan2024weighted}.

In \cite{li2014weighted}, Li et al. introduced a weighted guided image ﬁlter (WGIF) by leveraging an edge-aware weighting in place of the constant regularization parameter in the original GIF. Kou et al. \cite{kou2015gradient} proposed a gradient domain guided image filter (GGIF) with a new edge-aware weighting by incorporating an explicit first-order edge-aware constraint. The edge-aware weighting and the constraint can both make the filter preserve edges better. In \cite{zhang2022robust}, Zhang and He presented a robust double-weighted guided image filter (RDWGIF) by incorporating the robust edge-aware weighting and the mollifier used in Sobolev space into the cost function of GIF. On the other hand, the authors in \cite{sun2019weighted,ochotorena2019anisotropic}  advocated using the weighted averaging instead of the arithmetic averaging in the final step of guided filtering to achieve higher edge-preserving performance, thereby reducing halo artifacts effectively. For that, Sun et al. \cite{sun2019weighted} proposed a weighted guided image filter with steering kernel (SKWGIF), where the normalized local steering kernel of the guidance image is leveraged as weighting in their weighted averaging. Ochotorena and Yamashita \cite{ochotorena2019anisotropic} proposed a novel guided filter called anisotropic guided filter (AnisGF), which utilizes weighted averaging to achieve maximum diffusion while preserving strong edges. The weights are optimized based on local variances, thereby facilitating robust anisotropic filtering. In addition, the authors in \cite{yin2019side,yuan2024weighted} suggested that typical filtering scheme of placing local window center at the target pixel being processed is also a cause for generating edge blurring and halo artifacts. Based on this observation, the side window is introduced \cite{yin2019side} and used \cite{yuan2024weighted} in calculating the filtered output of GIF.

On the other hand, the principle of guided filtering is to transfer the structure of the guidance image to the filtered output. However, guided filtering usually performs poorly when there are inconsistent structures between the guidance image and the input one. To address this issue, the authors in \cite{shen2015mutual,ham2017robust,guo2017mutually} proposed the global approaches built on iterative algorithms, i.e., mutual-structure joint filtering (MSJF) \cite{shen2015mutual}, static/dynamic filtering (SDF) \cite{ham2017robust}, and mutually guided filtering (muGF) \cite{guo2017mutually}. In addition, guided filtering has recently advanced by learning-based ones \cite{li2016deep,shi2021unsharp,zhong2023deep}, which directly predict the filtered output by means of feature fusion from the guidance and target images, and do not rely on hand-crafted functions. However, learning-based guided filters require a large amount of data and time for training, and the implicit way of structure-transferring may fail to transfer the desired edges and  suffer from transferring undesired content to the target image \cite{shi2021unsharp}.

The guided image filters in \cite{li2014weighted,kou2015gradient,zhang2022robust,sun2019weighted,ochotorena2019anisotropic,yin2019side,yuan2024weighted} are built on an identical local prior model with two parameters (i.e., LAM). Despite their success in application, LAM-based GIFs present two intrinsic limitations in theory. By definition, the local affine transformation in LAM shares the identical affine coefficients in a local window, which makes LAM suffer from content-blindness in the sense of treating identically all pixels in the local window containing the target pixel, overlooking the distinctions between different pixels (edge or flat ones) in the local window. It is most likely for the LAM-based GIFs to poorly preserve some sharp edges because both edges and flat regions coexist within the local window. On the other hand, since LAM does not take into account the input image in the model formulation, LAM-based GIFs cannot provide an explicit understanding of the mechanism of transferring the guidance structure to the filtered output (see Table \ref{tab:comparison}). Yin et al. \cite{yin2019side} and Yuan et al. \cite{yuan2024weighted} have paid attention to the issue of local windows crossing edges, and thus adopted the side window strategy for guided filtering to preserve sharp edge. This strategy helps mitigate the issue to some extent; however, it will fail when the window radius is so large that the side window still crosses edges. Consequently, LAM-based GIFs still face the challenge in preserving sharp edges; it motivates us to explore a novel local prior model for guided filtering to cope with this challenge.

In this paper, we first introduce a novel local Prior Model involving a single parameter, based on Gaussian (highpass and lowpass) Filtering (PM-GF) for guided filtering, where the filtering
output is a weighted portion of Gaussian highpass ﬁltered guidance image plus Gaussian smoothed version of the input image. The PM-GF   takes into account the guidance structure (determined by Gaussian highpass filtering) and low frequency component of input image (determined by Gaussian lowpass filtering) simultaneously, thereby providing a clear explanation of the mechanism behind the structure transferring. Based on PM-GF, we propose several Gaussian highpass guided image filters (GH-GIFs) by imitating the original GIF  \cite{he2012guided} and some of its subsequent improvements, i.e., using PM-GF in place of LAM in these GIFs. 
Finally, we conduct a series of experiments to compare the PM-GF-based GIFs and the LAM-based ones in some applications: edge-aware smoothing, denoising, detail enhancement, HDR images tone mapping, image dehazing, and texture removal smoothing. Experimental results show that the PM-GF-based GIFs all outperform the LAM-based ones in terms of subjective assessment and objective evaluation.

The main contributions of this work are two-folds:
\begin{enumerate}[\textbullet]
\item We introduce a novel local prior model (i.e., PM-GF) with a single parameter for guided filtering, where the filtering output is a weighted portion of Gaussian highpass filtered guidance image plus Gaussian smoothed version of the input image. The PM-GF reveals the mechanism of how guided filtering transfers the guidance structure (determined by Gaussian highpass filtering) into the filtering output.  
\item We propose several Gaussian highpass guided image filters (GH-GIFs) by imitating the original GIF and some of its subsequent improvements. In particular, the LAM in the GIFs is replaced with the PM-GF in the GH-GIFs. Comprehensive comparison experiments show that PM-GF-based GIFs all outperform their counterparts in several image applications, subjectively and objectively.
\end{enumerate}

This paper is structured as follows. Section 2 briefly reviews the original guided image filter and some of its improvements. Section 3 presents our local prior model, as well as several GH-GIFs. 
Comprehensive comparison experiments are provided in Section 4. This paper is concluded in Section 5.

\section{Guided image filtering}
In this section, we briefly review the original GIF \cite{he2012guided} and some of its subsequent improvements \cite{li2014weighted,kou2015gradient,zhang2022robust,sun2019weighted,ochotorena2019anisotropic}, which will be employed in our comparison experiments.

\subsection{Local affine model (Prior model)}
Let $(I,G)$ denote an image pair, where $I$ and $G$ are the input image to be filtered and the guidance image, respectively.
He et al. \cite{he2012guided} first proposed the GIF for image smoothing, in which the filtering output is calculated by considering the content of the guidance image. The prior model for GIF is a local affine model (LAM), which assumes that the filtering output $O$ is an affine transformation of the guidance image $G$ in a local window $\omega_{p'}$ centered at pixel $p'$, as follows: 
\begin{equation}\label{eq.2.1}
	O_{p'}\big(p\big)=a_{p'}G\big(p\big)+b_{p'},\forall p\in\omega_{p'},
\end{equation}
where the coefficients $a_{p'}$ and $b_{p'}$ are  constant in the window $\omega_{p'}$ (square window of  radius $r$). The subscript in $O_{p’}$ is used to indicate that the $O_{p’}$ value is associated with $p'$. 

The idea behind equation \eqref{eq.2.1} is that the pixel $p$ in the window $\omega _{p'}$ shares the identical  coefficients $a_{p'}$ and $b_{p'}$. Obviously, the LAM treats identically pixels from distinct regions (edges and flat ones) within the window $\omega _{p'}$, regardless of whether the window $\omega _{p'}$ contains edge pixels or not. The assumption of constant coefficients is not valid when the window $\omega _{p'}$ contains the edges and flat patch simultaneously. Moreover, the LAM does not provide an intuitive comprehending of how to transfer the guidance structure to the filtering output, and does not explicitly consider the content of the input image. 

\subsection{Definition of GIF}
In the original GIF \cite{he2012guided}, the affine coefficients $a_{p'}$ and $b_{p'}$ are obtained by minimizing the following cost function:
\begin{equation}\label{eq.2.2}
E({a_{p'}},{b_{p'}}) = \sum\limits_{p \in {\omega _{p'}}} {\left( {{{\left( {{a_{p'}}G(p) + {b_{p'}} - I(p)} \right)}^2} + \varepsilon a_{p'}^2} \right)},
\end{equation}
where $I$  is the input image, and $\varepsilon >0$  is a regularization parameter, penalizing large $a_{p'}$. By least squares method, the optimal values of $a_{p'}$ and $b_{p'}$ are computed as:
\begin{equation}\label{eq.2.3}
	a_{p'}=\frac{\mu_{G\circ I}\left(p'\right)-\mu_G\left(p'\right)\mu_I\left(p'\right)}{\sigma_G^2\left(p'\right)+\varepsilon},
\end{equation}
and
\begin{equation}\label{eq.2.4}
	b_{p'}=\mu_I\left(p'\right)-a_{p'}\mu_G\left(p'\right),
\end{equation}
where $G \circ I$ denotes the pixelwise product of $G$ and $I$, and $\mu_X(p^{\prime})$ and $\sigma_X^2(p^{\prime})$  are the mean and variance of image $X$ in the window $\omega_{p'}$, respectively.

The edge-preserving filtering property of the GIF can be explained intuitively as follows \cite{he2012guided}. For the convenience of theoretical analysis, He et al. \cite{he2012guided} only considered the self-guided case where $G(p)=I(p)$, although the general case remains valid in practice. In this case, it is easily derived from \eqref{eq.2.3} and \eqref{eq.2.4} that
\begin{equation}\label{eq.2.5}
	a_{p'}=\frac{\sigma_I^2(p')}{\sigma_I^2(p')+\varepsilon},~b_{p'}=\left(1-a_{p'}\right)\mu_I\left(p'\right).
\end{equation}
Based on \eqref{eq.2.5}, the parameter $\varepsilon$ can be interpreted as a threshold, utilized to discriminate the “high variance” patches and “flat patches” in the input image. Specifically, a patch with variance $\sigma_I^2(p^{\prime})\ll\varepsilon $ is regarded as “flat patch” and so is smoothed (due to $O_{p’}(p)\approx\mu_I(p^{\prime}),~\forall p\in\omega_{p^{\prime}}$), whereas a patch with variance $\sigma_I^2(p^{\prime})\gg\varepsilon $ is considered as a “high variance” patch/an edge and thus is preserved (due to $O_{p’}(p)\approx I(p),~\forall p\in\omega_{p^{\prime}}$). In other words, the parameter $\varepsilon$ in \eqref{eq.2.2} serves as a threshold, which greatly decides the extent to which “high variance” patches/edges are preserved in the filtered output.

However, the parameter $\varepsilon$ is identical for all  windows ($\omega_{p^{\prime}},~\forall p^{\prime}$), and must be ﬁne-tuned for a specific image. If the  $\varepsilon$  value is too large, the filtering process over-smooths the input image and therefore sharp edges cannot be well preserved. In contrast, if the  $\varepsilon$  value is too small, the filtering process will yield a filtered image similar to the original one. This is one reason why the GIF \cite{he2012guided} cannot well preserve sharp edges and so suffers from halo artifacts around some edges.

On the other hand, the filtering output $O_{p'}(p)$  at pixel $p\in\omega_{p^{\prime}}$  depends upon the pixel $p'$; in other words, $O_{p'}(p)$  has different values when it is computed in different windows of  $\omega_{p^{\prime}}$, where $p^{\prime}\in\omega_p$ . He et al. \cite{he2012guided} used the arithmetic averaging to address this problem of overlapping windows. Specifically, the final output $O(p)$  of the GIF is computed by means of the following formula:
\begin{equation}
	O\left(p\right)=\frac{1}{\mid\omega\mid}\sum_{p^{\prime}\in\omega_{p}}O_{p^{\prime}}(p)=\mu_{a}(p)G\left(p\right)+\mu_{b}(p),~\forall p,
\end{equation}
where  $\mu_a(p)$ and $\mu_b(p)$  are the mean values of  $a_{p'}$ and  $b_{p'}$  in window $\omega_{p}$  , respectively:
\begin{equation}
	\mu_a(p)=\frac{1}{\mid\omega\mid}\sum_{p^{\prime}\in\omega_p}a_{p^{\prime}},\mu_b(p)=\frac{1}{\mid\omega\mid}\sum_{p^{\prime}\in\omega_p}b_{p^{\prime}},
\end{equation}
where  $\mid\omega\mid$ is the number of elements in $\omega_{p}$.

The mean operator (lowpass filter) computes the coefficients  $\mu_a(p)$ and $\mu_b(p)$  of the ﬁltered output $O(p)$  by averaging all values of  $a_{p'}$ and  $b_{p'}$ which are calculated in distinct but overlapping windows containing  $p$. The isotropy of mean operator assigns the same weight (i.e., ${1}/{\mid\omega\mid}$) to all pixels in window $\omega_{p}$, disregarding spatial distance between the pixels and the target pixel $p$. When larger sizes of local windows are employed, this lowpass filtering effect is further enforced, making more “high variance” patches/edges be smoothed in the filtering process. This is another reason why the GIF \cite{he2012guided} cannot well preserve edges and thus suffers from halo artifacts around some edges to some extent.

\subsection{Some improvements}
As noted above, the GIF \cite{he2012guided} cannot well preserve sharp edges and thus suffers from halo artifacts caused by edge blurring. Two main reasons for this phenomenon are to utilize the constant parameter  $\varepsilon$ and adopt the arithmetic averaging strategy to address the problem of overlapping windows involved in computing the ﬁltered output. To address these issues, several improved GIFs have been developed in the literature; see for example\cite{li2014weighted,kou2015gradient,zhang2022robust,sun2019weighted,ochotorena2019anisotropic}.

In \cite{li2014weighted,kou2015gradient,zhang2022robust}, the constant parameter $\varepsilon$  in \eqref{eq.2.2} is replaced by an adaptive variable parameter  $\varepsilon w(p')$. The function $w(p')$  is the edge-aware weight (EAW), having the property that it is very large when the pixel  $p'$ is in a flat patch, while it is very small when the pixel $p'$  locates in an edge patch. 

In \cite{li2014weighted}, EAW is defined by the variance of the guidance image in a $3\times 3$  local window; specifically, EAW  $w_1(p')$ is defined by the formula:

\begin{equation}\label{eq.2.8}
	\frac1{w_1(p^{\prime})}=\frac1N\sum_{p=1}^N\frac{\sigma_{G,1}^2(p^{\prime})+\tau}{\sigma_{G,1}^2(p)+\tau},\quad\forall p^{\prime},
\end{equation}
where $\tau>0$  is a small constant, $\sigma_{G,1}^2$  is the variance of the guidance image $G$ in a  $3\times 3$  window, and $N$  is the number of pixels in $G$. We can rewrite the EAW  $w_1(p')$ as the following form:
\begin{equation}\label{eq.2.9}
	w_1(p')=\frac{H_{1,G}}{\sigma_{G,1}^2(p')+\tau},H_{1,G}=1\Bigg/\left(\frac{1}{N}\sum_{p=1}^{N}\frac{1}{\sigma_{G,1}^2(p)+\tau}\right),
\end{equation}
where  $H_{1,G}$ denotes the harmonic mean of  $\sigma_{G,1}^2(p')+\tau$ over the image $G$. It can easily be seen from \eqref{eq.2.9} that the value of  $w_1(p')$ is usually smaller than $1$ if  $p'$ is at an edge and larger than $1$ if $p'$ is in a flat area. Therefore, smaller weights are assigned to pixels at edges than those pixels in flat areas by the weight $w_1(p')$.

In \cite{kou2015gradient}, Kou et al. first introduced a multiscale EAW $w_2(p')$  by multiplying the local variances of both  $3\times 3$ and $(2r+1) \times (2r+1)$  windows of all pixels in the guidance image. Similar to  $w_1(p')$ above,  $w_2(p')$ is rewritten as:
\begin{equation}\label{eq.2.10}
	w_2(p')=\frac{H_{2,G}}{\chi\left(p'\right)+\tau},H_{2,G}=1\Bigg/\left(\frac{1}{N}\sum_{p=1}^{N}\frac{1}{\chi(p)+\tau}\right),
\end{equation}
where  $\chi\left(p'\right)=\sigma_{G,1}(p')\sigma_{G,r}(p')$, $\sigma_{G,r}(p')$ is the local variance of the image $G$ in a $(2r+1) \times (2r+1)$  window, and  $H_{2,G}$ is the harmonic mean of $\chi\left(p'\right)+\tau$  over the image $G$. The multi-scale EAW  $w_2(p')$ can better separate edges of an image from ﬁne details of the image than the single-scale EAW  $w_1(p')$. As shown in \cite{kou2015gradient}, the multiscale EAW  $w_2(p')$ detect edges more accurately while enhancing ﬁne details better compared to the EAW  $w_1(p')$, with negligible increment of the computation time. 

On the other hand, Kou et al. \cite{kou2015gradient} introduced a novel cost function as follows:
\begin{equation}\label{eq.2.11}
	E(a_{p^{\prime}},b_{p^{\prime}})=\sum_{p\in\omega_{p^{\prime}}}\left(\left(a_{p^{\prime}}G(p)+b_{p^{\prime}}-I(p)\right)^2+\varepsilon w_2(p^{\prime})\left(a_{p^{\prime}}-\gamma_{p^{\prime}}\right)^2\right),
\end{equation}
where $\gamma_{p^{\prime}}$  is defined as:
\begin{equation*}
	\gamma_{p^{\prime}}=1-\frac1{1+e^{\eta(\chi(p^{\prime})-\bar{\chi})}},
\end{equation*}
where  $\bar{\chi}$ is the mean value of all  $\chi(p)$, and $\eta$  is calculated via  $4/(\bar{\chi}-\min\chi(p))$. The analysis and experimental results in \cite{kou2015gradient} shows that the edges can be better preserved by GGIF  \cite{kou2015gradient} than by GIF \cite{he2012guided} and  WGIF\cite{li2014weighted}. Besides, the GGIF is less sensitive to the selection of  $\varepsilon$.

In \cite{zhang2022robust}, Zhang and He first introduced Maximum Neighbor Difference (MND), which can be aware of different edges and textures in the image. Then they utilized the MND to define a robust EAW as follows:
\begin{equation}\label{eq.2.12}
	{w_3}(p') = \left\{ \begin{array}{l}
		{\left( {1 - {{\left( {\frac{{{{{\mathop{\rm MND}\nolimits} }_G}\left( {p'} \right)}}{{cS\left( G \right)}}} \right)}^2}} \right)^2},~{\rm{ }}{\mathop{\rm if}\nolimits} \left| {{{{\mathop{\rm MND}\nolimits} }_G}\left( {p'} \right)} \right| < cS\left( G \right) \\ 
		~~~~~~~~~~~~~0, ~~~~~~~~~~~~~~~~{\mathop{\rm if}\nolimits} \left| {{{{\mathop{\rm MND}\nolimits} }_G}\left( {p'} \right)} \right| \ge cS\left( G \right) \\ 
	\end{array} \right.,
\end{equation}
where the scale $c>0$  is an integer, and the image-dependent scale $S(G)$  is estimated by Median Absolute Deviation (MAD) in robust statistics:
\begin{equation}\label{eq.2.13}
	S\left(G\right)=1.4826\cdot\underset{p\in G}{\operatorname*{median}}\left\{\left|{\operatorname*{MND}}_{G}\left(p\right)-\underset{p\in G}{\operatorname*{median}}\left\{{\operatorname*{MND}}_{G}\left(p\right)\right\}\right|\right\},
\end{equation}
where the function $\operatorname*{median}\{\cdot\}$ is performed over all locations in $G$ (guidance image). As shown in \cite{zhang2022robust}, REAW $w_3(p')$  can capture structures of various scales from the guidance image.
Finally, they introduced the following double-weighted cost function:
\begin{equation}\label{eq.2.14}
	E(a_{p^{\prime}},b_{p^{\prime}})=\sum_{p\in\omega_{p^{\prime}}} \left( M_{p^{\prime},p}\left(a_{p^{\prime}}G(p)+b_{p^{\prime}}-I(p)\right)^2+\epsilon w_3(p^{\prime})a_{p^{\prime}}^2\right),
\end{equation}
where  $M_{p^{\prime},p}$ is the ﬁlter kernel associated with a molliﬁer used in Sobolev space theory. In addition, the filtered output of RDWGIF \cite{zhang2022robust} is computed by the following formula:
\begin{equation}\label{eq.2.15}
	O\left(p\right)=G\left(p\right)\sum_{p^{\prime}\in\omega_p}M_{p,p^{\prime}}a_{p^{\prime}}+\sum_{p^{\prime}\in\omega_p}M_{p,p^{\prime}}b_{p^{\prime}},
\end{equation}
which is different from \cite{he2012guided,li2014weighted,kou2015gradient}.

In \cite{sun2019weighted} and \cite{ochotorena2019anisotropic}, the authors adopted the weighted mean operator to compute the filtered output  $O(p)$, as follows:
\begin{equation}\label{eq.2.16}
	O\left(p\right)=\tilde{a}_pG\left(p\right)+\tilde{b}_p,~\forall p,
\end{equation}
where
\begin{equation}\label{eq.2.17}
	\tilde{a}_p=\sum_{p'\in\omega_p}w_{pp'}a_{p'},~\tilde{b}_p=\sum_{p'\in\omega_p}w_{pp'}b_{p'},
\end{equation}
are the weighted mean values of $a_{p'}$  and  $b_{p'}$ in window  $\omega_{p}$, respectively. 
In \cite{sun2019weighted}, the weights  $w_{pp'}$ are described by the local steering kernel depended on the guidance image which make full use of the edge direction and thus can better preserve edges and suppress the halo artifacts simultaneously. In \cite{ochotorena2019anisotropic}, the weights $w_{pp'}$  are computed by minimizing an objective function based on the local neighborhood variances, which make AnisGF \cite{ochotorena2019anisotropic} achieve strong anisotropic filtering while preserving strong edges in the filtered image.

The GIF \cite{he2012guided} and some subsequent improvements (e.g., \cite{li2014weighted,kou2015gradient,zhang2022robust,sun2019weighted,ochotorena2019anisotropic}), which are all based on the prior model LAM in \eqref{eq.2.1}, have been widely used in many image applications. Despite their success, the LAM makes these guided filters still cannot well preserve sharp edges, and therefore exhibit halo artifacts caused by edge blurring in the filtered outputs. Moreover, the LAM does not provide an intuitive understanding of how the LAM-based GIFs transfer the guidance structure into the filtered output, and does not explicitly take into account the input image in the LAM formulation. Therefore, the LAM-based GIFs still face the challenge in suppressing halo artifacts caused by edge blurring. This observation motivates us to explore a novel prior model for guided filtering to overcome the limitations of the  LAM.

\section{Gaussian highpass guided image filtering (GH-GIF)}
In this section, we first introduce a novel prior model (Prior Model based on Gaussian (highpass and lowpass) Filtering, PM-GF) for guided filtering, with a comparison between the PM-GF and the LAM. Then we propose a new guided image filter based on PM-GF (i.e., GH-GIF), as well as some  improvements.

\subsection{Prior model based on Gaussian (highpass and lowpass) filtering (PM-GF)}
Given an image pair $(I, G)$ consisting of the input image $I$ to be filtered and the guidance image $G$, we proceed to the derivation of the PM-GF for guided filtering. 

Recall that the original GIF \cite{he2012guided} is made up of \eqref{eq.2.1}, \eqref{eq.2.3} and \eqref{eq.2.4} in the previous section. By substituting $b_{p'}$ in \eqref{eq.2.4} into the formula \eqref{eq.2.1}, the filtering output $O_{p^{\prime}}$ can be represented as
\begin{equation}\label{eq.3.1}
	O_{p^{\prime}}\left(p\right)=a_{p^{\prime}}\left(G\left(p\right)-\mu_G\left(p^{\prime}\right)\right)+\mu_I\left(p^{\prime}\right),~\forall p\in\omega_{p^{\prime}}.
\end{equation}
Note that the local means $\mu_G(p')$  and $\mu_I(p')$  are both dependent upon $p'$ rather than $p$; that is, both $\mu_G(p')$  and $\mu_I(p')$ are constant in the window $\omega_{p'}$. 
Inspired by formula \eqref{eq.3.1}, we suggest a novel local prior model as follows:
\begin{equation}\label{eq.3.2}
	O_{p^{\prime}}(p)=\alpha_{p^{\prime}}\left(G(p)-\bar{G}(p)\right)+\bar{I}(p),~\forall p\in\omega_{p^{\prime}},
\end{equation}
where the parameter $\alpha_{p'}$ is assumed to be constant in the window $\omega_{p'}$, and $\bar G$ and $\bar I$ denote the Gaussian lowpass filtering outputs of $G$ and $I$, respectively. To simplify notation, here we use $\bar X$ to represent the Gaussian lowpass filtering output of image $X$. Note that both $\bar G(p)$ and $\bar I(p)$ are dependent of $p$ rather than $p'$, i.e., they are not constant in the window $\omega_{p'}$.

Now we analyse the difference image  $G(p)-\bar{G}(p)$ in the frequency domain. Let ${\mathcal{F}}$  be the Fourier transformation, and $H(u,\nu)$ be the 2-D frequency domain Gaussian transfer function where $u$ and $v$ denote the frequency variables. Then we have
\begin{equation}\label{eq.3.3}
	\begin{aligned}
		\mathcal{F}\left(G-\bar{G}\right)& =\mathcal{F}\left(G-\mathcal{F}^{-1}(H)*G\right)  \\
&=\mathcal{F}\left(G\right)-\mathcal{F}\left(\mathcal{F}^{-1}\left(H\right)\right)\cdot\mathcal{F}\left(G\right)=\left(1-H\right)\mathcal{F}(G),
	\end{aligned}
\end{equation}
where $1-H(u,\nu)$  is the transfer function of Gaussian highpass filter. As a result, the difference image $G(p)-\bar{G}(p)$  is indeed the filtering output of Gaussian highpass filter with the transfer function  $1-H(u,\nu)$. For this reason, the prior model \eqref{eq.3.2} is referred to as Prior Model based on Gaussian (highpass and lowpass) Filtering (PM-GF for short).

\begin{figure}[!htbp]
	\centering
	\subfigure[]{\includegraphics[width=0.3\linewidth]{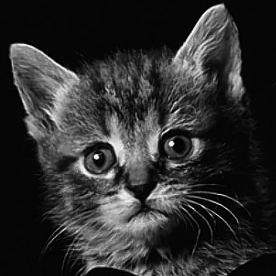}}
	\subfigure[]{\includegraphics[width=0.3\linewidth]{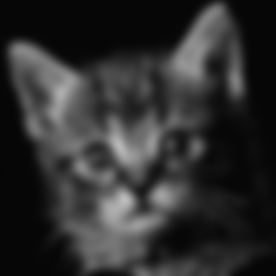}}
	\subfigure[]{\includegraphics[width=0.3\linewidth]{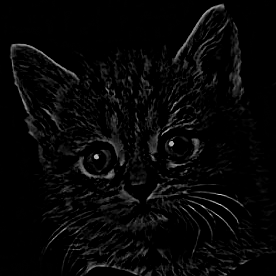}}	
	\caption{(a) $G$ \& $I$. (b) $\bar G$ and $\bar I$. (c)  $G- \bar G$.}
	\label{fig:3.1-0}
\end{figure}

We show an example of $G$ in Fig. \ref{fig:3.1-0}(a). Fig. \ref{fig:3.1-0}(b) denotes the Gaussian lowpass filtering of $G$ (i.e., $\bar{G}$), and Fig. \ref{fig:3.1-0}(c) is the difference image $G-\bar{G}$. It can be observed in Fig. \ref{fig:3.1-0}(c) that the difference image $G(p)-\bar{G}(p)$ highlights structures (edge lines) and other discontinuities in the image while de-emphasizing regions of slowly varying intensities.

In the following, we give a comparison between LAM \eqref{eq.2.1}  and PM-GF \eqref{eq.3.2}.

\begin{figure}[!htbp]
	\centering
	\subfigure[]{\includegraphics[width=0.2\linewidth]{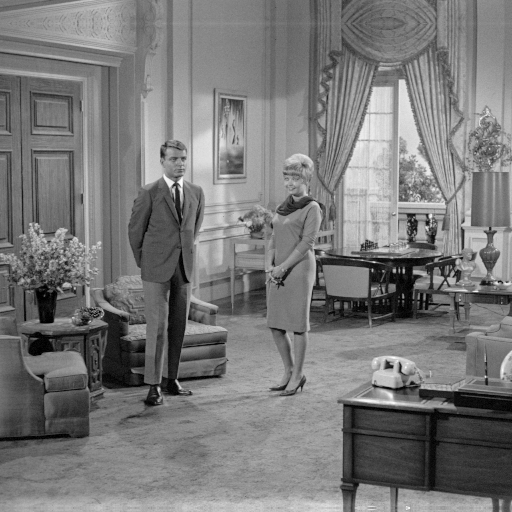}}
	\subfigure[]{\includegraphics[width=0.2\linewidth]{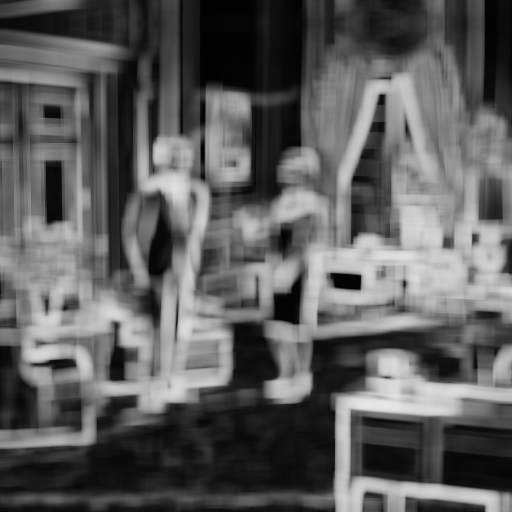}}
	\subfigure[]{\includegraphics[width=0.2\linewidth]{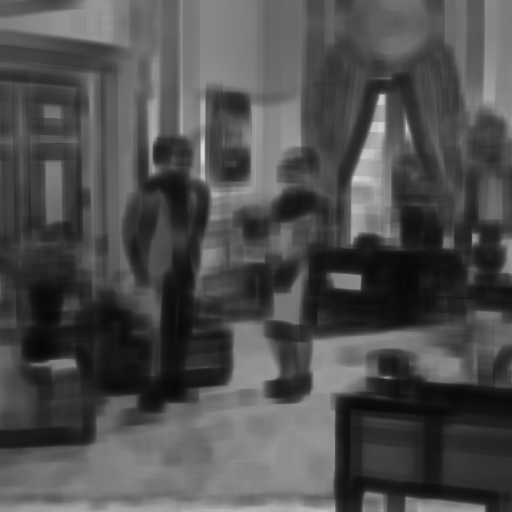}}
	\subfigure[]{\includegraphics[width=0.2\linewidth]{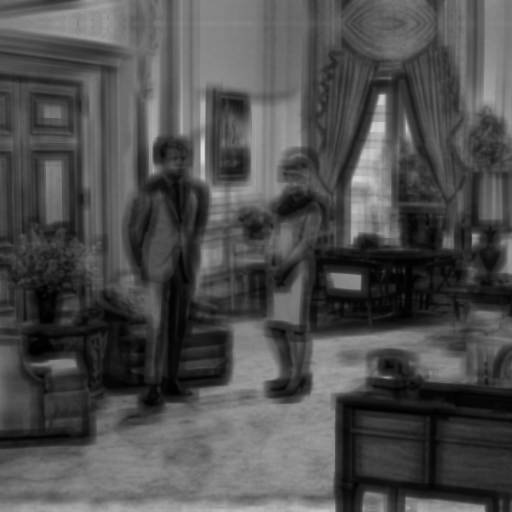}}	
	\caption{Comparion of $b_{p'}$ in \eqref{eq.2.1} and $\beta_{p'}$ in \eqref{eq.3.5-1} in the self-guided case. (a) $G$ \& $I$. (b) $a_{p'}$. (c) $b_{p'}$. (d) $\beta_{p'}$}
	\label{fig:3.1-1}
\end{figure}

Firstly, in \eqref{eq.2.1}, each pixel in window $\omega_{p'}$ is assigned the same scale $ a_p'$ and the offset $ b_p'$, whether it is from a flat region or edge one. This might result in edge blurring when the window contains  flat and edge pixels simultaneously.
In order to facilitate comparison, we rewrite \eqref{eq.3.2} in the following form:
\begin{equation}
	O(p)=\alpha_{p^{\prime}}G(p)+\beta_{p'}(p),~\forall p\in\omega_{p^{\prime}},
	\label{eq.3.5-1}
\end{equation}
where $\beta_{p'}(p)=\bar{I}(p)-\alpha_{p^{\prime}}\bar{G}(p)$. 
The offset $\beta_{p'}(p)$ corresponds to the costant $b_{p'}$  in \eqref{eq.2.1}, but it varies across different positions of $p$ in  window $\omega_{p^{\prime}}$. For visual comparison, images of both offsets are shown in Fig. \ref{fig:3.1-1}(c)(d), where 
the optimal values for $a_{p'}$ and $b_{p'}$ are calculated by \eqref{eq.2.3}, while  $\alpha_{p'} $ is crudely set to $a_{p'}$ in  \eqref{eq.2.3}. Although such an $a_{p'}$ is not the optimal value of $\alpha_{p'}$ for PM-GF (see \eqref{eq.3.8} below for the optimal), it can be observed from Fig. \ref{fig:3.1-1}(c)(d) that $b_{p'}$ exhibits block artifacts, and $\beta_{p'}$ better preserves edges and more details than $b_{p'}$. 

Secondly, we make a comparison between the filtered outputs for guided filters derived from \eqref{eq.3.1} and \eqref{eq.3.2}.
The filtering outputs $O_{p'}(p)$  in \eqref{eq.3.1} and \eqref{eq.3.2} are both dependent on the pixel  $p'$; i.e., $O_{p'}(p)$  has different value when it is computed in different windows of  $\omega_{p'}$, where  $p\in \omega_{p'}$. To simplify the analysis, the arithmetic averaging is adopted to address this problem of overlapping windows. Specifically, the ﬁltered output $O(p)$ is calculated by the arithmetic average of $O_{p'} (p)$ (with respect to $ p'$) in the window $\omega (p)$ at pixel $p$. Therefore, the filtered output of the guided filter based on \eqref{eq.3.1}/\eqref{eq.3.2} is estimated by
\begin{equation}\label{eq.3.4}
	\begin{aligned}
		O(p)& =\frac1{\mid\omega\mid}\sum_{p^{\prime}\in\omega_p}O_{p^{\prime}}(p)  \\
		&=\frac1{\mid\omega\mid}\sum_{p^{\prime}\in\omega_p}a_{p^{\prime}}\left(G\left(p\right)-\mu_G\left(p^{\prime}\right)\right)+\frac1{\mid\omega\mid}\sum_{p^{\prime}\in\omega_p}\mu_I(p^{\prime})
	\end{aligned}
\end{equation}
for the guided filter based on \eqref{eq.3.1}, and
\begin{equation}\label{eq.3.5}
	\begin{aligned}
		O(p)& =\frac1{\mid\omega\mid}\sum_{p^{\prime}\in\omega_p}O_{p^{\prime}}(p)  \\
		&=\frac1{\mid\omega\mid}\sum_{p^{\prime}\in\omega_p}\alpha_{p^{\prime}}\left(G\left(p\right)-\bar{G}\left(p\right)\right)+\bar{I}\left(p\right) \\
		&=\bar{\alpha}_p\left(G(p)-\bar{G}(p)\right)+\bar{I}(p)
	\end{aligned}
\end{equation}
for the guided filter based on \eqref{eq.3.2}, where $\bar{\alpha}_p$ is the mean value of $\alpha_{p'}$  in window $\omega (p)$.

\begin{figure}[!htbp]
	\centering
	\subfigure{\includegraphics[width=1\linewidth]{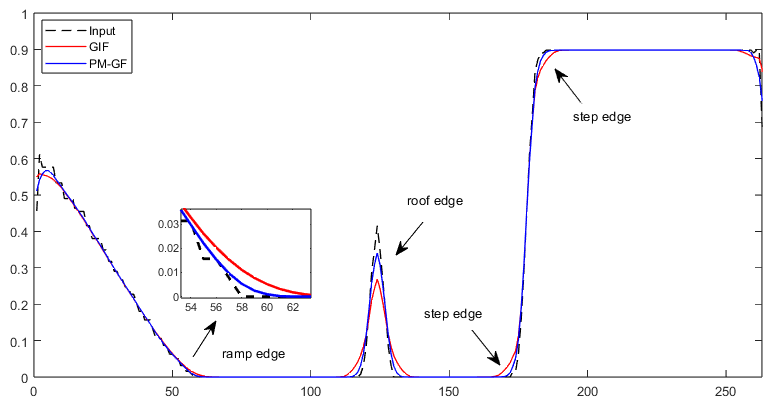}}
	\caption{The $100th$ horizontal intensity profile of Fig. \ref{fig:3.3}(a) and filtered output \eqref{eq.3.4} (GIF) and  \eqref{eq.3.5} (PM-GF) with $\alpha_p’=a_p’$, given in \eqref{eq.2.3}.}
	\label{fig:3.1-2}
\end{figure}

In comparison to \eqref{eq.3.4}, the filtered output \eqref{eq.3.5} in theory leads to a better and more intuitive understanding of how the guided ﬁlter transfers the guidance structure (determined by Gaussian highpass filtering) into the filtered output. In practice, the filtered output of \eqref{eq.3.5} is also more superior to that of \eqref{eq.3.4} in terms of edge preservation. To verify this assert, Fig. \ref{fig:3.1-2} shows the filtered outputs of \eqref{eq.3.4} and \eqref{eq.3.5}, where $a_{p’}$ in \eqref{eq.3.4} is chosen as the optimal value given in \eqref{eq.2.3} where $r=4$, $\varepsilon=0.16$, while $\alpha_{p'}$ in \eqref{eq.3.5} is directly set to $a_{p’}$ in \eqref{eq.2.3}; see \eqref{eq.3.8} for the optimal $\alpha_{p'}$ value. We can observe that the filtered output of \eqref{eq.3.5} really preserves edges better than that of \eqref{eq.3.4}.

\subsection{Definition of Gaussian highpass guided image filter}
In our PM-GF formulation, the filtering output $O_{p'}$ is computed by formula \eqref{eq.3.2}, where $\alpha_{p'}$ is an undetermined constant parameter. To obtain the optimal value of $\alpha_{p'}$, we choose to consider the ridge regression as follows:
\begin{equation}\label{eq.3.7}
	\min_{\alpha_{p^{\prime}}}E(\alpha_{p^{\prime}})=\sum_{p\in\omega_{p^{\prime}}}\left(\left(\alpha_{p^{\prime}}\left(G(p)-\bar{G}(p)\right)+\bar{I}(p)-I(p)\right)^2+\lambda\alpha_{p^{\prime}}^2\right),
\end{equation}
where $\lambda>0$ is a regularization parameter. According to ridge regression,
the term $\lambda\alpha_{p^{\prime}}^2$ is added to deal with the overfitting problem involved in  minimization of $L^2$-distance between $O_{p’}$ and $I$ in $\omega_{p'}$. Moreover, in the self-guided case where $G(p) = I(p)$, if it were ignored, we would obtain the trivial solution $\alpha_{p'}= 1$, indicating the filtered output is exactly the input image itself.

Differentiating $E(\alpha_{p^{\prime}})$ with respect to $\alpha_{p'}$ and making the derivative vanish, yields:
\begin{equation}\label{eq.3.8}
	\alpha_{p^{'}}=\frac{\frac{1}{\mid\omega\mid}\sum_{p\in\omega_{p^{'}}}\left(G(p)-\bar{G}(p)\Big)\Big(I(p)-\bar{I}(p)\right)}{\frac{1}{\mid\omega\mid}\sum_{p\in\omega_{p^{'}}}\left(G(p)-\bar{G}(p)\right)^{2}+\lambda},~\forall p^{\prime}.
\end{equation}
Note that the first quantity in the denominator is not the variance of $G$ in window $\omega_{p'}$. 
By formula \eqref{eq.3.5}, the filtered output of our GH-GIF is estimated by the formola:
\begin{equation}\label{eq.3.9}
	O(p)=\bar{\alpha}_p\left(G(p)-\bar{G}(p)\right)+\bar{I}(p),
\end{equation}
where the weight  $\bar{\alpha}_p$ is the mean value of  $\alpha_p$ in the window $\omega (p)$, namely,
\begin{equation}\label{eq.3.10}
	\bar{\alpha}_p=\frac{1}{\mid\omega\mid}\sum_{p^{\prime}\in\omega_p}\alpha_{p^{\prime}},
\end{equation}
where the parameter  $\alpha_{p'}$ is given in \eqref{eq.3.8}.  

In conclusion, for given image pair $(I, G)$, the definition of our GH-GIF is summarized in \eqref{eq.3.8}, \eqref{eq.3.10} and \eqref{eq.3.9}; see also Algorithm \ref{algorithm}. For comparison,  Table \ref{tab:comparison} shows the similarities and differences between the original GIF \cite{he2012guided} and our GH-GIF.

\begin{algorithm}
	\caption{ GH-GIF} 
	\hspace{0.05in} \vspace{0.05cm} 
	\textbf{Input:} input  $I$, guidance  $G$. 
	\begin{algorithmic}[1]
		\STATE  Compute $\bar{G}(p)$  and $\bar{I}(p)$  by Gaussian smoothing;			
		\STATE  Compute $\alpha_{p'}$  by Eq. \eqref{eq.3.8};
		\STATE  Compute $\bar \alpha_{p}$  by Eq. \eqref{eq.3.10};
		\STATE  Compute $O(p)$  by Eq. \eqref{eq.3.9}.
	\end{algorithmic}
	\hspace{0.02in} \vspace{0.1cm} \textbf{Output:} filtered output $O$. 
	\label{algorithm} 
\end{algorithm}

\begin{table}[!htbp]
	\centering
	\caption{Comparison between GIF \cite{he2012guided} and our GH-GIF. PM: Prior Model. UP: Undetermined parameter. CF: Cost Function. FC: Filter Coefficient. FO: Filtered Output.}
	\small
	\begin{tabular}{@{}p{0.6cm} p{7.2cm} p{7.2cm}@{}}
		\toprule
		& \textbf{GIF} \cite{he2012guided} & \textbf{GH-GIF} \\
		\midrule
		\textbf{PM} & 
		\(\begin{aligned}[t]
			O_{p'}(p) = a_{p'}G(p) + b_{p'}, \forall p \in \omega_{p'}
		\end{aligned}\) & 
		\(\begin{aligned}[t]
			O_{p^{\prime}}(p) &= \alpha_{p^{\prime}}\left(G(p) - \bar{G}(p)\right) + \bar{I}(p), \forall p \in \omega_{p^{\prime}}
		\end{aligned}\) \\
		\textbf{UP} & 
		$a_{p'},~ b_{p'}$& 
		$\alpha_{p^{\prime}}$  \\
		\textbf{CF} & 
		\(\begin{aligned}[t]
			E(a_{p^{\prime}},b_{p^{\prime}}) &= \sum_{p \in \omega_{p^{\prime}}} \left( \left( O_{p^{\prime}}(p) - I(p) \right)^2 + \varepsilon a_{p^{\prime}}^2 \right)
		\end{aligned}\) & 
		\(\begin{aligned}[t]
			E(\alpha_{p^{\prime}}) &= \sum_{p \in \omega_{p^{\prime}}} \left( \left( O_{p^{\prime}}(p) - I(p) \right)^2 + \lambda \alpha_{p^{\prime}}^2 \right)
		\end{aligned}\) \\
		\textbf{FC} & 
		\(\begin{aligned}[t]
			a_{p^{\prime}} &= \frac{\mu_{G \circ I}(p^{\prime}) - \mu_{G}(p^{\prime})\mu_{I}(p^{\prime})}{\sigma_{G}^2(p^{\prime}) + \varepsilon} \\
			b_{p'} &= \mu_{I}(p') - a_{p'}\mu_{G}(p')
		\end{aligned}\) 
		& 
		\(\begin{aligned}[t]
			\alpha_{p^{\prime}} &= \frac{\frac{1}{|\omega|} \sum_{p \in \omega_{p^{\prime}}} \left(G(p) - \bar{G}(p)\right)\left(I(p) - \bar{I}(p)\right)}{\frac{1}{|\omega|} \sum_{p \in \omega_{p^{\prime}}} \left(G(p) - \bar{G}(p)\right)^2 + \lambda}
		\end{aligned}\)
		\\
		\textbf{FO} & 
		\(\begin{aligned}[t]
			O(p) &= \frac{1}{|\omega|} \sum_{p^{\prime} \in \omega_{p}} O_{p^{\prime}}(p) \\
			&= \frac{1}{|\omega|} \sum_{p^{\prime} \in \omega_{p}} a_{p^{\prime}} \left(G(p) - \mu_G(p^{\prime})\right) + \mu_I(p^{\prime})
		\end{aligned}\) & 
		\(\begin{aligned}[t]
			O(p) &= \frac{1}{|\omega|} \sum_{p^{\prime} \in \omega_{p}} O_{p^{\prime}}(p) \\
			&= \frac{1}{|\omega|} \sum_{p^{\prime} \in \omega_{p}} \alpha_{p^{\prime}} \left(G(p) - \bar{G}(p)\right) + \bar{I}(p)
		\end{aligned}\) \\
		\bottomrule
	\end{tabular}
	\label{tab:comparison}
\end{table}

Finally, we discuss the influence of regularization parameter $\lambda$ upon edge preservation, as well as computational complexity of our GH-GIF.
Recall that the filtered output of our GH-GIF is estimated by the formula:
\begin{equation}\label{eq.3.11}
	O(p)=\bar{\alpha}_p\left(G(p)-\bar{G}(p)\right)+\bar{I}(p),
\end{equation}
where the weight $\bar{\alpha}_p$  is the mean value of $\alpha_{p'}$  in the window $\omega (p)$:
\begin{equation}\label{eq.3.12}
	\bar{\alpha}_{p}=\frac{1}{\mid\omega\mid}\sum_{p^{\prime}\in\omega_{p}}\alpha_{p^{\prime}},~ \alpha_{p^{'}}=\frac{\frac{1}{\mid\omega\mid}\sum_{p\in\omega_{p^{'}}}\left(G(p)-\bar{G}(p)\Big)\Big(I(p)-\bar{I}(p)\right)}{\frac{1}{\mid\omega\mid}\sum_{p\in\omega_{p^{'}}}\left(G(p)-\bar{G}(p)\right)^{2}+\lambda}.
\end{equation}
Obviously, the weight $\bar{\alpha}_p$ increases as the parameter $\lambda$ decreases, thereby increasing the contribution of  $G(p)-\bar{G}(p)$ in the process of transferring the guidance structure to the filtered output. In theory, the parameter $\lambda$ should be small enough to achieve the advantageous performance in preserving sharp edges. However, if $\lambda$ is too small, the weight $\bar{\alpha}_p$ may be too large in some flat patches in $G$, thereby introducing the artifacts (spurious details) in the filtered output. In many experiments, our observation is that the value of $\lambda=10^{-4}$ is almost at the extreme of what we can use without introducing some serious artifacts in the filtered output image.

\begin{table}[!htbp]
	\centering
	\caption{Average CPU time (second) of 100 runs.}
	\begin{tabular}{lllll}
		\toprule
		& Fig. \ref{fig:3.3}(a) & Fig.  \ref{fig:3.3}(b) & Fig.  \ref{fig:3.3}(c) & Fig.  \ref{fig:3.3}(d) \\
		\midrule
		GIF   & 0.0097 & 0.0343 & 0.0342 & 0.0345 \\
		GH-GIF & \textbf{0.0085} & \textbf{0.0236} & \textbf{0.0239} & \textbf{0.0240} \\
		\bottomrule
	\end{tabular}
	\label{tab:3.2}
\end{table}

\begin{figure}[!htbp]
	\centering
	\subfigure[]{\includegraphics[width=0.2\linewidth]{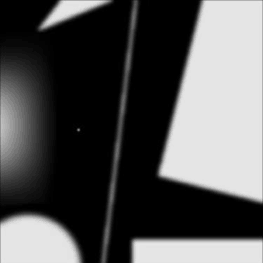}}
	\subfigure[]{\includegraphics[width=0.2\linewidth]{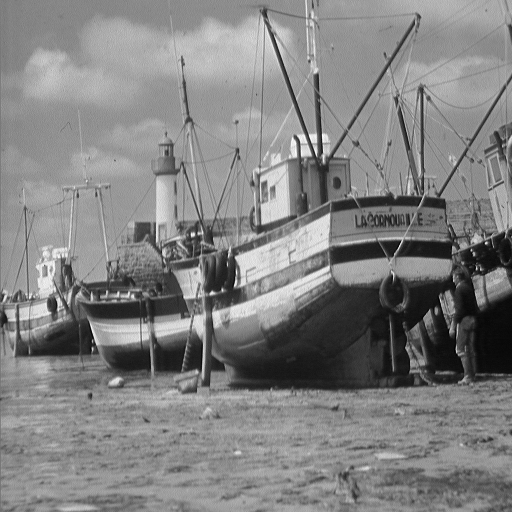}}
	\subfigure[]{\includegraphics[width=0.2\linewidth]{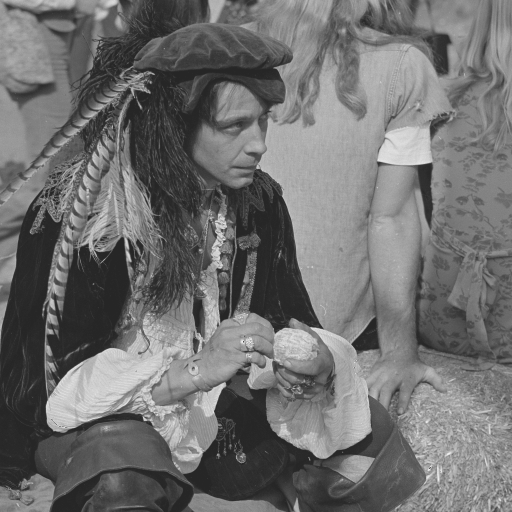}}
	\subfigure[]{\includegraphics[width=0.2\linewidth]{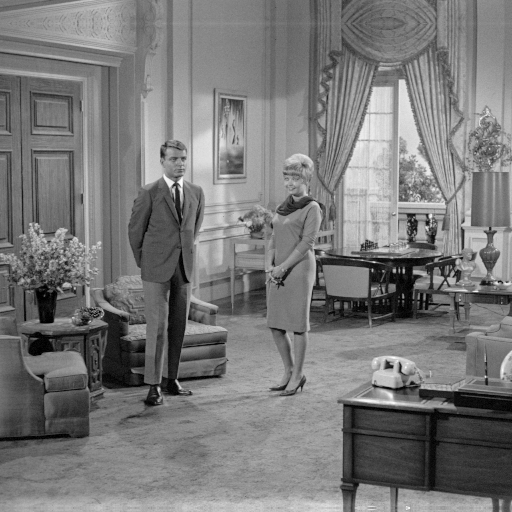}}	
	\caption{Test images. (a) $263 \times 263$. (b) $512 \times 512$. (c) $512 \times 512$. (d) $512 \times 512$.}
	\label{fig:3.3}
\end{figure}

For time complexity of Algorithm \ref{algorithm}, computational expense primarily relies on the calculation of  $\bar{G}(p)$, $\bar{I}(p)$, and  $\bar \alpha_p$, all having $O(N)$  time. Therefore, Algorithm \ref{algorithm} has a time complexity of $O(N)$, matching with the GIF algorithm \cite{he2012guided}. Even so, because our prior model (PM-GF) has only one parameter to be estimated, the proposed GH-GIF should be slightly faster than the GIF \cite{he2012guided} in theory. In fact, in all our experiments, we found that Algorithm \ref{algorithm} is really slightly faster than the GIF algorithm. In Table \ref{tab:3.2}, we list the CPU time of GIF and GH-GIF, performed on four images in Fig. \ref{fig:3.3}.

\subsection{Several improvements}
Some of existing guided image filters, such as GIF  \cite{he2012guided}, WGIF \cite{li2014weighted}, GGIF \cite{kou2015gradient}, SKWGIF \cite{sun2019weighted}, AnisGF \cite{ochotorena2019anisotropic}, and RDWGIF \cite{zhang2022robust}, are all driven from the local affine model (LAM) in \eqref{eq.2.1}. We use our PM-GF \eqref{eq.3.2} in place of the LAM in these filters to form new GIFs, which are referred to as GH-GIF, GH-WGIF, GH-GGIF, GH-SKWGIF, GH-AnisGF, GH-RDWGIF, respectively; they will be employed in our comparison experiments.

As an example, we choose to use GGIF \cite{kou2015gradient} to demonstrate the filtering process of GH-GGIF. Following \eqref{eq.2.11}, the cost function of GH-GGIF is defined as:
\begin{equation}\label{eq.3.13}
	E(\alpha_{p^{\prime}})=\sum_{p\in\omega_{p^{\prime}}}\left( \left(\alpha_{p^{\prime}}\left(G(p)-\bar{G}(p)\right)+\bar{I}(p)-I(p) \right)^{2}+\lambda w_{2}(p^{\prime}) \left(\alpha_{p^{\prime}}-\gamma(p^{\prime})\right)^{2}\right).
\end{equation}
By differentiating $E(\alpha_{p^{\prime}})$ with respect to $\alpha_{p'}$ and making the derivative vanish, we obtain the optimal value of $\alpha_{p'}$ as follows:
\begin{equation}\label{eq.3.14}
	\alpha_{p^{\prime}}=\frac{\frac{1}{\mid\omega\mid}\sum_{p\in\omega_{p^{\prime}}}\left(G(p)-\bar{G}(p)\right)\left(I(p)-\bar{I}(p)\right)+\lambda w_{2}(p^{\prime})\gamma\left(p^{\prime}\right)}{\frac{1}{\mid\omega\mid}\sum_{p\in\omega_{p^{\prime}}}\left(G(p)-\bar{G}(p)\right)^{2}+\lambda w_{2}(p^{\prime})}.
\end{equation}
As done in \cite{kou2015gradient}, the ﬁltered output $O(p)$ is estimated by the arithmetic average of $O_{p'} (p)$ (with respect to $ p'$) in the window $\omega (p)$ at pixel $p$. Therefore, the filtered output is calculated by
\begin{equation}\label{eq.3.15}
	O(p)=\frac{1}{\mid\omega\mid}\sum_{p'\in\omega_{p}}\alpha_{p'}\left(G(p)-\bar{G}(p)\right)+\bar{I}(p),~\forall p.
\end{equation}

\section{Experiments and applications}
In this section, we compare the performance of GH-GIF, GH-WGIF, GH-GGIF, GH-SKWGIF and GH-RDWGIF with their counterparts (GIF \cite{he2012guided}, WGIF \cite{li2014weighted}, GGIF \cite{kou2015gradient}, SKWGIF \cite{sun2019weighted}, RDWGIF \cite{zhang2022robust}) in some applications in terms of subjective assessment and objective evaluation.
Such applications include edge-aware smoothing, denoising, image detail enhancement, tone mapping of HDR images, image dehazing, and texture removal smoothing. In addition, in each experiment, the PM-GF-based GIFs use the same Gaussian filter parameters.
For simplicity, in all of those experiments, the window radius $r$ and the regularization parameter $\varepsilon$ for two types of guided filters are selected according to some of the original filters. (We always set $\lambda=0.1\varepsilon$ for PM-GF-based guided filters, where $\varepsilon $  is the regularization parameter of the original guided filter.) Such values of $r$ and $\varepsilon$ are not selected optimally in six real applications because we just want to compare two types of guided filters in the same parameter setting.

\begin{table}[H]
	\centering
	\caption{Average PSNR (dB) of different filters performed on dataset BSD68. Best values are highlighted in bold. (↑): the higher, the better.}
	\begin{tabular}{llllllll}
		\toprule
		\multicolumn{2}{c}{\multirow{2}[4]{*}{PSNR ↑}} & \multicolumn{2}{l}{$r=2$} & \multicolumn{2}{l}{$r=4$} & \multicolumn{2}{l}{$r=8$} \\
		\cmidrule{3-8}    \multicolumn{2}{c}{} & LAM & PM-GF & LAM & PM-GF & LAM & PM-GF \\
		\midrule
		$\epsilon=0.1^2$ & GIF   & 31.32  & \textbf{39.29 } & 30.20  & \textbf{39.00 } & 29.38  & \textbf{38.66 } \\
		& WGIF  & 31.59  & \textbf{39.41 } & 30.40  & \textbf{39.11 } & 29.52  & \textbf{38.75 } \\
		& GGIF  & 36.23  & \textbf{42.31 } & 34.69  & \textbf{42.31 } & 33.43  & \textbf{42.19 } \\
		& SKWGIF & 32.32  & \textbf{38.68 } & 31.13  & \textbf{38.46 } & 30.21  & \textbf{38.02 } \\
		& RDWGIF & 44.66  & \textbf{48.98 } & 43.41  & \textbf{49.32 } & 41.12  & \textbf{48.94 } \\
		$\epsilon=0.2^2$ & GIF   & 27.67  & \textbf{34.12 } & 26.08  & \textbf{33.70 } & 24.77  & \textbf{33.29 } \\
		& WGIF  & 28.04  & \textbf{34.39 } & 26.36  & \textbf{33.92 } & 24.97  & \textbf{33.46 } \\
		& GGIF  & 34.80  & \textbf{39.09 } & 32.55  & \textbf{38.82 } & 30.48  & \textbf{38.42 } \\
		& SKWGIF & 28.96  & \textbf{34.00 } & 27.17  & \textbf{33.58 } & 25.68  & \textbf{32.94 } \\
		& RDWGIF & 42.42  & \textbf{45.46 } & 40.76  & \textbf{45.74 } & 37.56  & \textbf{44.84 } \\
		$\epsilon=0.4^2$ & GIF   & 25.64  & \textbf{30.75 } & 23.80  & \textbf{30.43 } & 22.23  & \textbf{30.15 } \\
		& WGIF  & 25.93  & \textbf{31.06 } & 24.01  & \textbf{30.66 } & 22.38  & \textbf{30.32 } \\
		& GGIF  & 34.30  & \textbf{37.56 } & 31.62  & \textbf{37.11 } & 29.05  & \textbf{36.55 } \\
		& SKWGIF & 26.97  & \textbf{30.84 } & 24.83  & \textbf{30.39 } & 23.03  & \textbf{29.79 } \\
		& RDWGIF & 40.95  & \textbf{42.73 } & 38.93  & \textbf{43.07 } & 35.12  & \textbf{41.93 } \\
		\bottomrule
	\end{tabular}%
	\label{tab:4.1PSNR}%
\end{table}%

\subsection{Edge-aware smoothing}
The BSD68 dataset \footnote{\href{https://github.com/cszn/DnCNN/tree/master/testsets/BSD68}{https://github.com/cszn/DnCNN/tree/master/testsets/BSD68}} is used in this application, which consists of sixty-eight images, covering a wide range of scenes and contents, such as natural landscapes, urban environments, objects and textures.

We employ two full-reference metrics, PSNR and SSIM \cite{wang2004image}, to compare the performance of all test filters quantitatively; higher PSNR and SSIM values indicate better performance. Tables \ref{tab:4.1PSNR} and \ref{tab:4.1SSIM} list the PSNR and SSIM values for each filter across various parameter configurations, i.e., $r =2,~4,~8$ and $\varepsilon=0.1^2,~0.2^2,~0.4^2$. In both tables, LAM and PM-GF denote guided filters based on LAM \eqref{eq.2.1} and PM-GF \eqref{eq.3.2}, respectively. In addition, the guidance image is identical to the input one.

\begin{figure}[!htbp]
	\centering
	\subfigure[]{\includegraphics[width=0.18\linewidth]{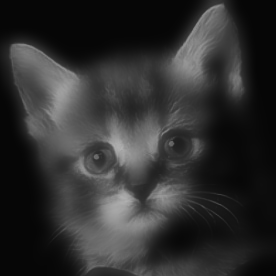}}
	\subfigure[]{\includegraphics[width=0.18\linewidth]{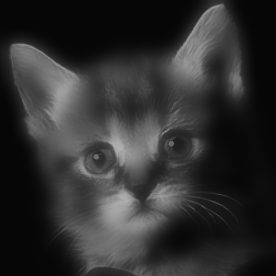}}
	\subfigure[]{\includegraphics[width=0.18\linewidth]{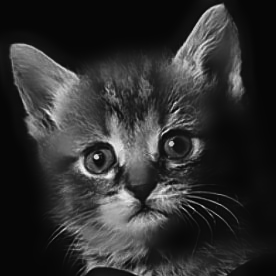}}
	\subfigure[]{\includegraphics[width=0.18\linewidth]{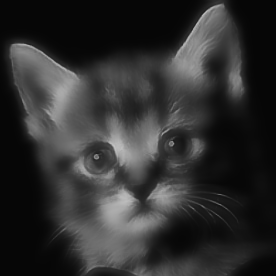}}	
	\subfigure[]{\includegraphics[width=0.18\linewidth]{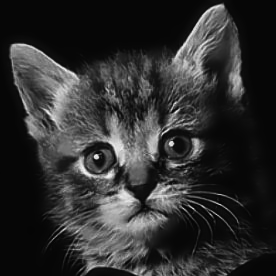}}
	\subfigure[]{\includegraphics[width=0.18\linewidth]{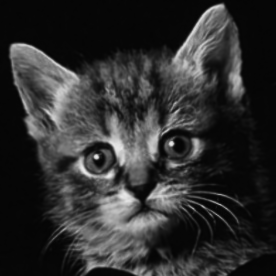}}
	\subfigure[]{\includegraphics[width=0.18\linewidth]{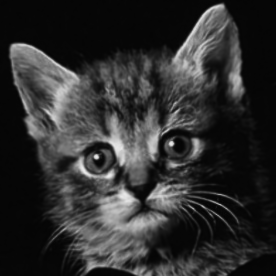}}
	\subfigure[]{\includegraphics[width=0.18\linewidth]{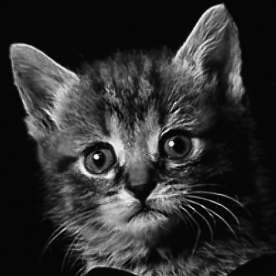}}
	\subfigure[]{\includegraphics[width=0.18\linewidth]{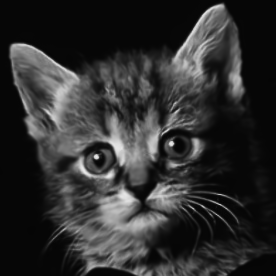}}	
	\subfigure[]{\includegraphics[width=0.18\linewidth]{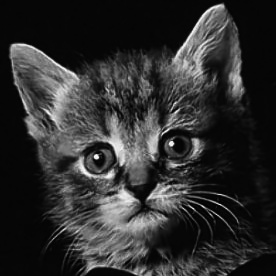}}
	\caption{Comparison of smoothing results by different filters. (a)-(e) GIF, WGIF, GGIF, SKWGIF, RDWGIF. (f)-(j) GH-GIF, GH-WGIF, GH-GGIF, GH-SKWGIF, GH-RDWGIF.}
	\label{fig:4.1}
\end{figure}

Fig. \ref{fig:4.1} displays the filtered outputs of different guided filters for the cat image used in \cite{he2012guided,sun2019weighted,zhang2022robust}, where $r=8$ and $\varepsilon=0.4^2$. It is evident that GH-GIF, GH-WGIF, GH-GGIF, GH-SKWGIF and GH-RDWGIF better preserve sharper edges compared to their counterparts, i.e.,GIF\cite{he2012guided}, WGIF \cite{li2014weighted}, GGIF \cite{kou2015gradient}, SKWGIF \cite{sun2019weighted}, and RDWGIF \cite{zhang2022robust}.

\begin{table}[H]
	\centering
	\caption{Average SSIM\cite{wang2004image} of different filters performed on dataset BSD68. Best values are highlighted in bold. (↑): the higher, the better.}
	\begin{tabular}{llllllll}
		\toprule
		\multicolumn{2}{c}{\multirow{2}[4]{*}{SSIM ↑}} & \multicolumn{2}{l}{$r=2$} & \multicolumn{2}{l}{$r=4$} & \multicolumn{2}{l}{$r=8$} \\
		\cmidrule{3-8}    \multicolumn{2}{c}{} & LAM & PM-GF & LAM & PM-GF & LAM & PM-GF \\
		\midrule
		$\epsilon=0.1^2$ & GIF   & 0.8656  & \textbf{0.9713 } & 0.8509  & \textbf{0.9719 } & 0.8634  & \textbf{0.9735 } \\
		& WGIF  & 0.8671  & \textbf{0.9711 } & 0.8517  & \textbf{0.9717 } & 0.8637  & \textbf{0.9733 } \\
		& GGIF  & 0.9237  & \textbf{0.9790 } & 0.9118  & \textbf{0.9800 } & 0.9191  & \textbf{0.9823 } \\
		& SKWGIF & 0.8794  & \textbf{0.9596 } & 0.8564  & \textbf{0.9595 } & 0.8627  & \textbf{0.9603 } \\
		& RDWGIF & 0.9873  & \textbf{0.9942 } & 0.9815  & \textbf{0.9943 } & 0.9775  & \textbf{0.9947 } \\
		$\epsilon=0.2^2$ & GIF   & 0.7702  & \textbf{0.9345 } & 0.7201  & \textbf{0.9336 } & 0.7120  & \textbf{0.9342 } \\
		& WGIF  & 0.7759  & \textbf{0.9351 } & 0.7254  & \textbf{0.9341 } & 0.7161  & \textbf{0.9347 } \\
		& GGIF  & 0.8992  & \textbf{0.9616 } & 0.8714  & \textbf{0.9621 } & 0.8670  & \textbf{0.9650 } \\
		& SKWGIF & 0.8031  & \textbf{0.9148 } & 0.7404  & \textbf{0.9130 } & 0.7191  & \textbf{0.9099 } \\
		& RDWGIF & 0.9826  & \textbf{0.9899 } & 0.9730  & \textbf{0.9899 } & 0.9621  & \textbf{0.9898 } \\
		$\epsilon=0.4^2$ & GIF   & 0.6986  & \textbf{0.8898 } & 0.6141  & \textbf{0.8875 } & 0.5755  & \textbf{0.8866 } \\
		& WGIF  & 0.7051  & \textbf{0.8920 } & 0.6206  & \textbf{0.8893 } & 0.5811  & \textbf{0.8882 } \\
		& GGIF  & 0.8889  & \textbf{0.9483 } & 0.8508  & \textbf{0.9478 } & 0.8346  & \textbf{0.9505 } \\
		& SKWGIF & 0.7462  & \textbf{0.8649 } & 0.6461  & \textbf{0.8609 } & 0.5882  & \textbf{0.8523 } \\
		& RDWGIF & 0.9792  & \textbf{0.9853 } & 0.9662  & \textbf{0.9852 } & 0.9476  & \textbf{0.9842 } \\
		\bottomrule
	\end{tabular}%
	\label{tab:4.1SSIM}%
\end{table}%

\subsection{Image Denoising}
We compare the denoising performance of GIFs based on LAM and PM-GF. In the experiments, we use all images from Set12 dataset \footnote{\href{https://github.com/cszn/FFDNet/tree/master/testsets/Set12}{https://github.com/cszn/FFDNet/tree/master/testsets/Set12}} containing twelve grayscale images, each depicting different scenes and objects. Noisy images were taken from the images in Set12 dataset by adding Gaussian noise (mean 0,  variance $(25/255)^2$). 

For each filter, we considered two cases: (1) the guidance is identical to the original clean image, and (2) the guidance is generated from the noisy image by performing Gaussian smoothing. In both cases, we set $r=4$ and $\varepsilon=0.2^2$, following SKWGIF \cite{sun2019weighted} and RDWGIF \cite{zhang2022robust}. Denoising performance is objectively evaluated by PSNR and SSIM\cite{wang2004image}.

\begin{figure}[!htbp]
	\centering
	\subfigure[]{\includegraphics[width=0.15\linewidth]{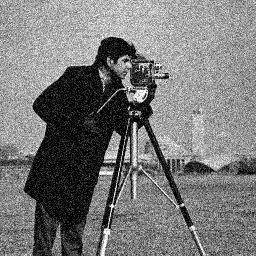}}
	\subfigure[]{\includegraphics[width=0.15\linewidth]{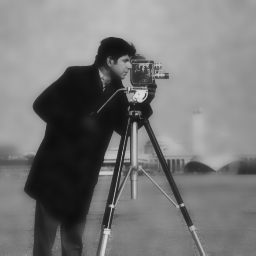}}
	\subfigure[]{\includegraphics[width=0.15\linewidth]{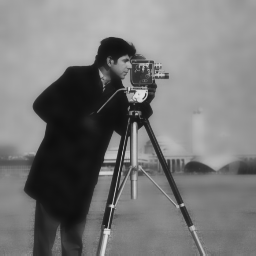}}
	\subfigure[]{\includegraphics[width=0.15\linewidth]{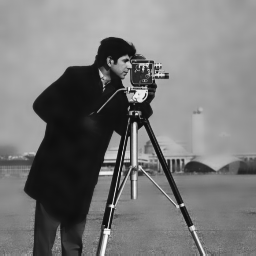}}
	\subfigure[]{\includegraphics[width=0.15\linewidth]{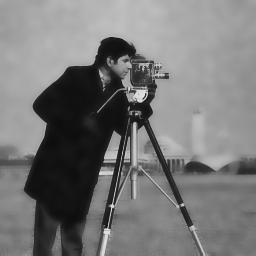}}	
	\subfigure[]{\includegraphics[width=0.15\linewidth]{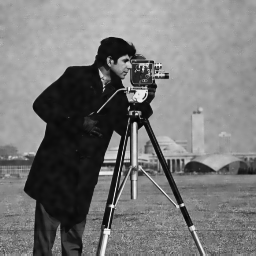}}
	\subfigure[]{\includegraphics[width=0.15\linewidth]{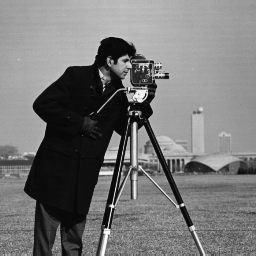}}
	\subfigure[]{\includegraphics[width=0.15\linewidth]{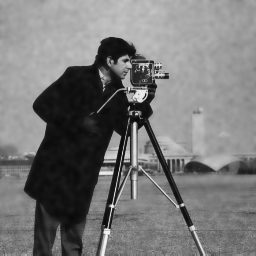}}
	\subfigure[]{\includegraphics[width=0.15\linewidth]{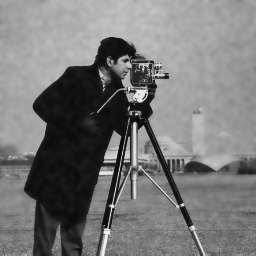}}
	\subfigure[]{\includegraphics[width=0.15\linewidth]{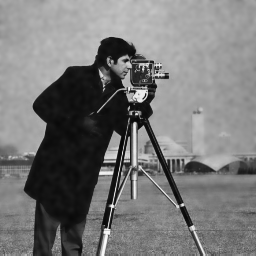}}
	\subfigure[]{\includegraphics[width=0.15\linewidth]{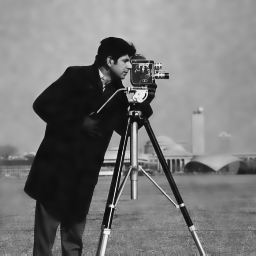}}	
	\subfigure[]{\includegraphics[width=0.15\linewidth]{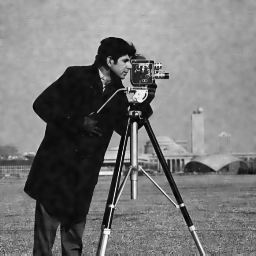}}
	\caption{Comparison of denoised results by different filters.  (a) Noisy image. (g) Guidance image. Denoised results: (b)-(f) GIF, WGIF, GGIF, SKWGIF, RDWGIF. (h)-(k) GH-GIF, GH-WGIF, GH-GGIF, GH-SKWGIF, GH-RDWGIF.}
	\label{fig:4.2}
\end{figure}

Table \ref{tab:4.2} presents the average PSNR and SSIM values across twelve noisy images, while Fig. \ref{fig:4.2} shows the denoising result of each filter for an image from Set12 dataset. Notably, PM-GF-based GIFs produce clearer architectural outlines in the background area of the image and demonstrate superior performance compared to their counterparts.

\begin{table}[!htbp]
	\centering
	\caption{Average PSNR (dB) and SSIM\cite{wang2004image} of different filters performed on Set12 dataset. Best values are highlighted in bold. (↑): the higher, the better.}
	\begin{tabular}{llllllll}
		\toprule
		& Metric & Model & GIF   & WGIF  & GGIF  & SKWGIF & RDWGIF \\
		\midrule
		\multicolumn{1}{p{4.19em}}{Case1:} & PSNR ↑ & LAM & 25.77 & 26.12 & 31.65 & 27.21 & 33.96 \\
		&       & PM-GF & \textbf{31.30} & \textbf{31.45} & \textbf{33.49} & \textbf{32.40} & \textbf{34.28} \\
		\cmidrule{3-8}          & SSIM ↑ & LAM & 0.7610 & 0.7671 & 0.8731 & 0.7858 & 0.9262 \\
		&       & PM-GF & \textbf{0.8792} & \textbf{0.8797} & \textbf{0.9036} & \textbf{0.8981} & \textbf{0.9331} \\
		\midrule
		\multicolumn{1}{p{4.19em}}{Case2} & PSNR ↑ & LAM & 24.00    & 24.10  & 26.86 & 24.71 & 27.15 \\
		&       & PM-GF & \textbf{26.86} & \textbf{26.91} & \textbf{27.46} & \textbf{27.12} & \textbf{27.64} \\
		\cmidrule{3-8}          & SSIM ↑ & LAM & 0.6945 & 0.697 & 0.7775 & 0.7123 & 0.7748 \\
		&       & PM-GF & \textbf{0.7718} & \textbf{0.7726} & \textbf{0.7824} & \textbf{0.7822} & \textbf{0.7852} \\
		\bottomrule
	\end{tabular}%
	\label{tab:4.2}%
\end{table}%

\subsection{Image detail enhancement}
 Single image detail enhancement is one of the application scenarios used to test the edge-preserving performance of a smoothing filter because it requires the filter to smooth details while preserving edges. In detail enhancement, an observed image $I$ is first smoothed to generate the base layer $\bar I$ (filter’s output) as well as the detail layer  $D = I- \bar I$. Then, the enhanced image is obtained by adding the  detail layer $kD $ boosted by a factor of $k >1$ back to the base layer $\bar I$. In equation form, the enhanced image is expressed as $	I_{en}=\bar I+k(I-\bar I)$.  The amplification factor $k$  is set to 5 in all the experiments in this section.

We evaluate the performance of guided filters on the widely-used image enhancement benchmarks, i.e., Kodak24 
\footnote{\href{http://www.r0k.us/graphics/kodak/}{http://www.r0k.us/graphics/kodak/}}, which contains twenty-four high-quality natural images of size $768 \times 512$. In the comparison experiments, following GGIF \cite{kou2015gradient}, we fix $r=16$ and $\varepsilon=0.1^2$ for all guided filters,  and the guidance image is the input one itself.  Evaluation metrics BIQI \cite{moorthy2010two} (used in \cite{li2014weighted,kou2015gradient,sun2019weighted,ochotorena2019anisotropic,zhang2022robust}), PIQE \cite{venkatanath2015blind}, and NIQE \cite{mittal2012making} are adopted to quantitatively assess detail enhancement quality. A higher BIQI value and lower PIQE \& NIQE values indicate higher enhancement quality.

\begin{figure}[!htbp]
	\centering		
	
	\includegraphics[width=0.15\linewidth]{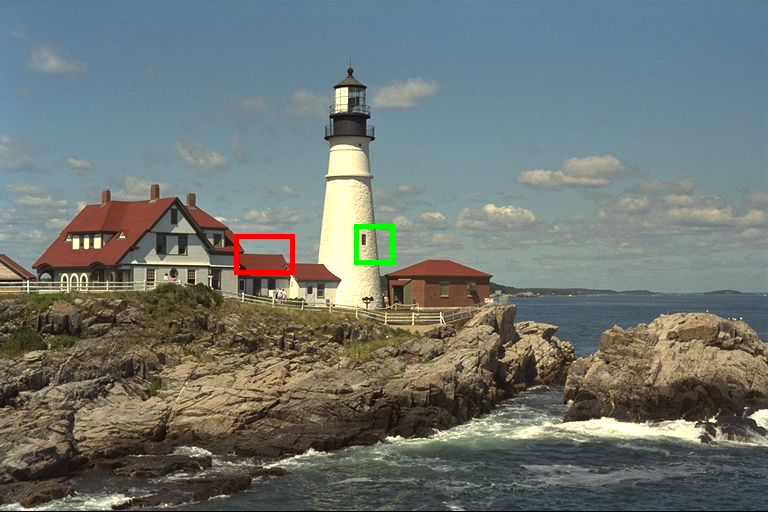}
	\includegraphics[width=0.15\linewidth]{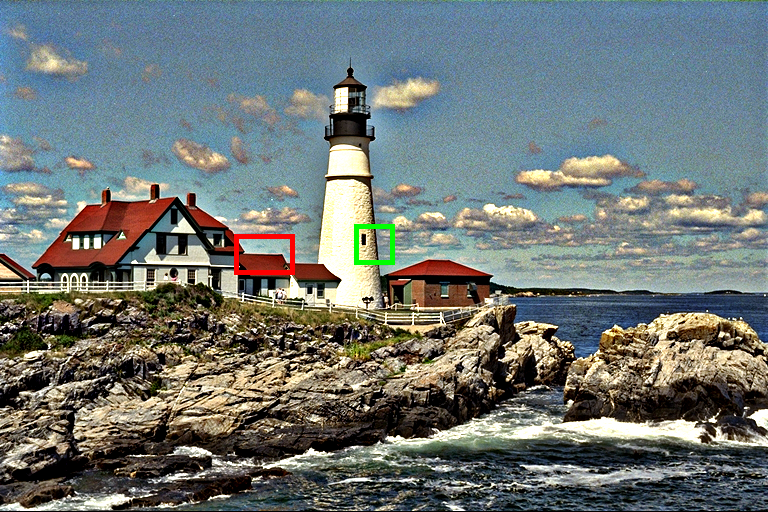}
	\includegraphics[width=0.15\linewidth]{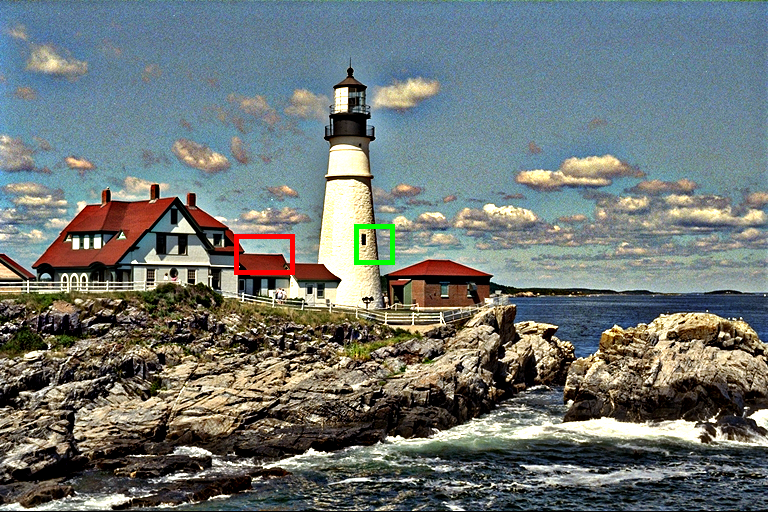}
	\includegraphics[width=0.15\linewidth]{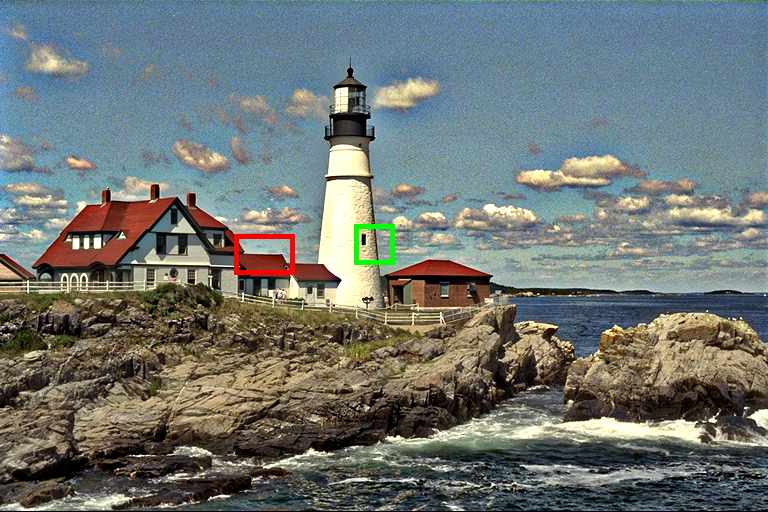}
	\includegraphics[width=0.15\linewidth]{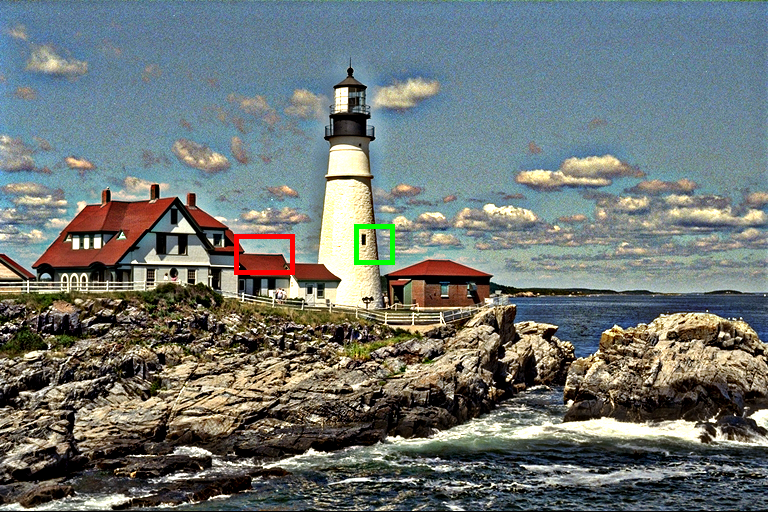}
	\includegraphics[width=0.15\linewidth]{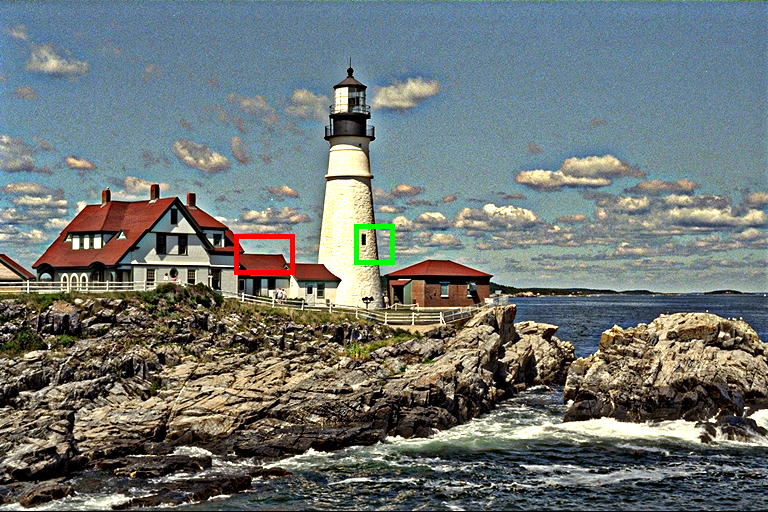}
	\subfigure[]{\includegraphics[width=0.15\linewidth]{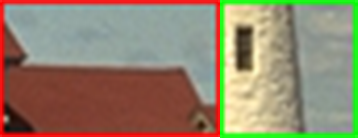}}
	\subfigure[]{\includegraphics[width=0.15\linewidth]{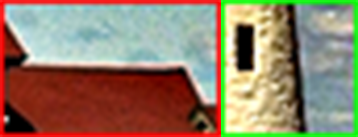}}
	\subfigure[]{\includegraphics[width=0.15\linewidth]{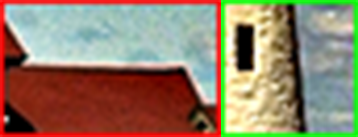}}
	\subfigure[]{\includegraphics[width=0.15\linewidth]{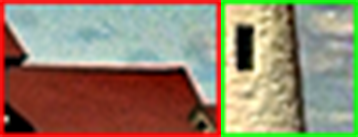}}
	\subfigure[]{\includegraphics[width=0.15\linewidth]{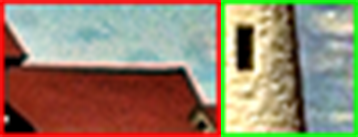}}
	\subfigure[]{\includegraphics[width=0.15\linewidth]{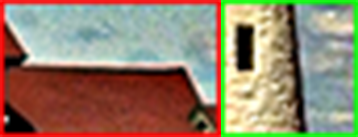}}

	\hspace{0.15\linewidth}
	\includegraphics[width=0.15\linewidth]{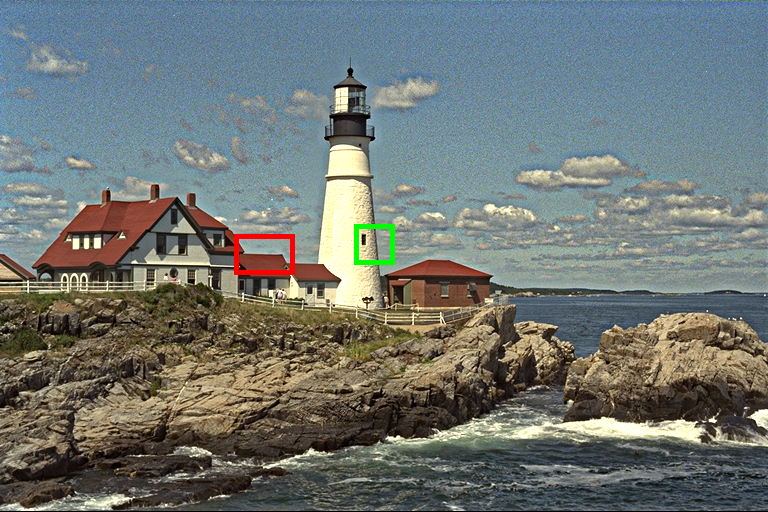}
	\includegraphics[width=0.15\linewidth]{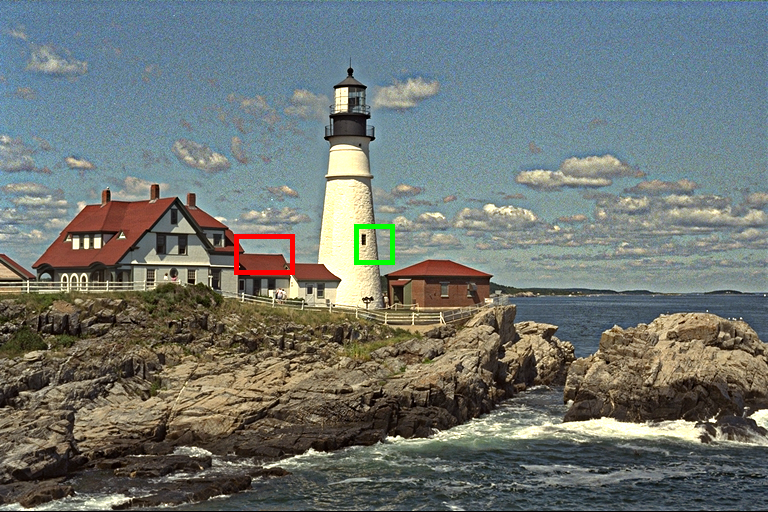}
	\includegraphics[width=0.15\linewidth]{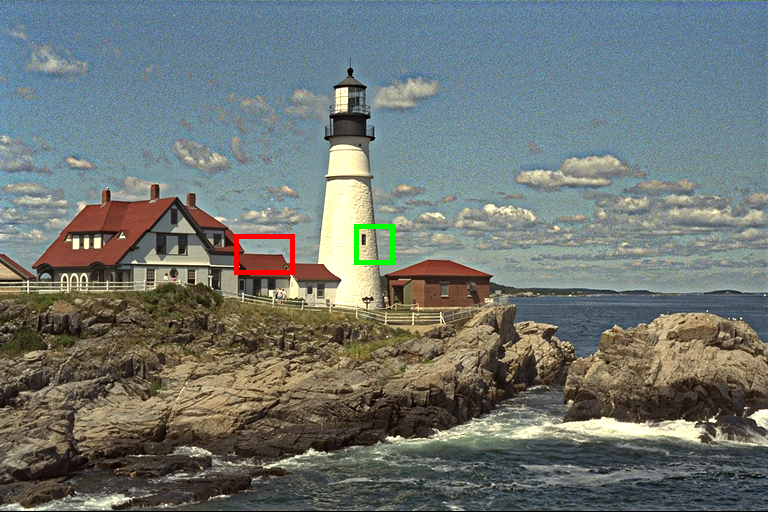}
	\includegraphics[width=0.15\linewidth]{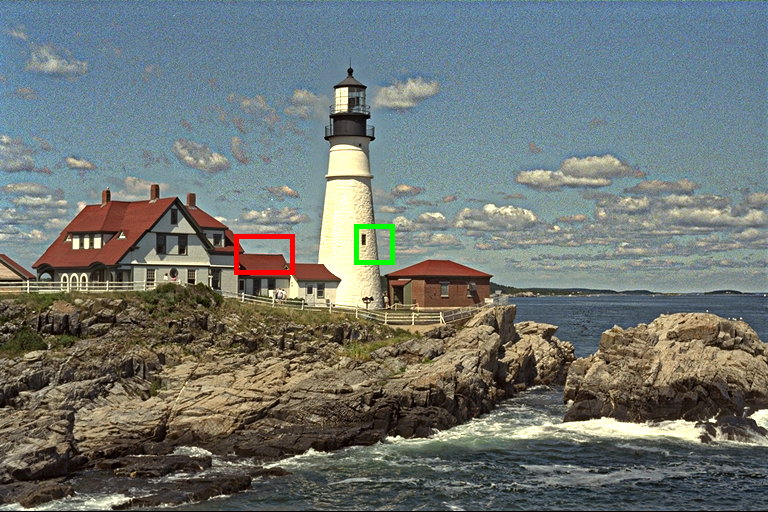}
	\includegraphics[width=0.15\linewidth]{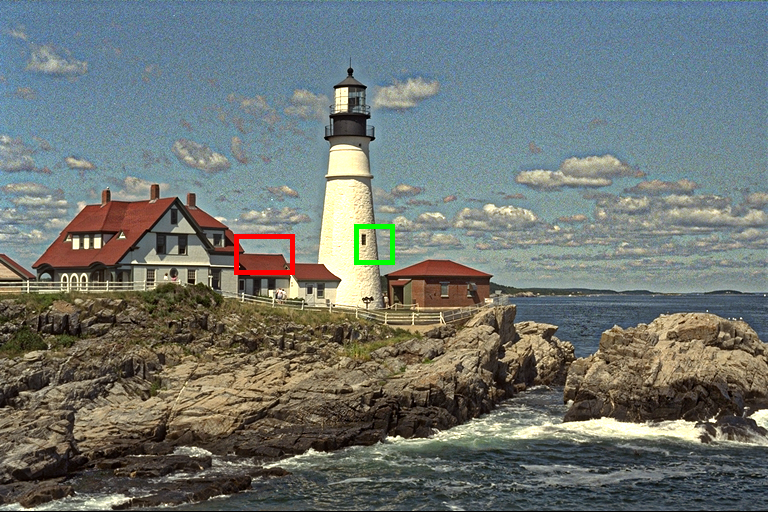}
	
	\hspace{0.15\linewidth}
	\subfigure[]{\includegraphics[width=0.15\linewidth]{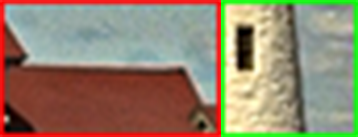}}
	\subfigure[]{\includegraphics[width=0.15\linewidth]{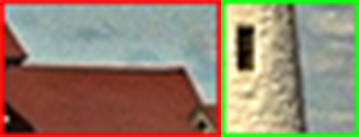}}
	\subfigure[]{\includegraphics[width=0.15\linewidth]{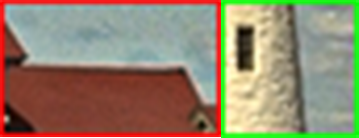}}
	\subfigure[]{\includegraphics[width=0.15\linewidth]{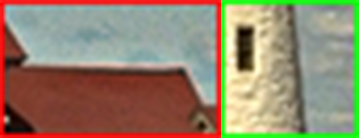}}
	\subfigure[]{\includegraphics[width=0.15\linewidth]{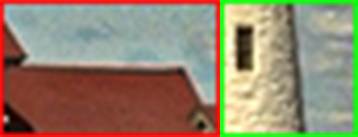}}

	\caption{Comparison of enhanced results by different filters. (a) Original image. Enhanced images: (b)-(f) GIF, WGIF, GGIF, SKWGIF, RDWGIF. (g)-(k) GH-GIF, GH-WGIF, GH-GGIF, GH-SKWGIF, GH-RDWGIF.}		
	\label{fig:4.3}
\end{figure}

Fig. \ref{fig:4.3} shows the detail enhanced images of ten guided filters performed on an image from Kodak24 dataset. Table \ref{tab:4.3} summarizes the average BIQI, PIQE, and NIQE values of ten guided filters on Kodak24 dataset. As shown in Fig. \ref{fig:4.3}, whether it's the horizontal roof edges or the vertical lighthouse edges, PM-GF-based GIFs demonstrate fewer artifacts and sharper edges compared to their counterparts.  Moreover, they get better BIQI, PIQE and NIQE values compared to LAM-based GIFs from Table. \ref{tab:4.3}.

\begin{table}[!htbp]
	\centering
	\caption{Average BIQI\cite{moorthy2010two}, PIQI\cite{venkatanath2015blind} and NIQE\cite{mittal2012making} of different filters performed on Kodak24 dataset. Best values are highlighted in bold. (↑): the higher, the better. (↓): the lower, the better.}
	\begin{tabular}{rllllll}
		\toprule
		\multicolumn{1}{l}{Metric} & Model & GIF   & WGIF  & GGIF  & SKWGIF & RDWGIF \\
		\midrule
		\multicolumn{1}{c}{BIQI ↑} & LAM & 40.45 & 40.80 & \textbf{43.33} & 42.59 & 44.99 \\
		& PM-GF & \textbf{41.53} & \textbf{41.81} & 41.98 & \textbf{43.41} & \textbf{45.08} \\
		\cmidrule{2-7}    \multicolumn{1}{c}{PIQE ↓} & LAM & 45.96 & 45.87 & 42.13 & 45.31 & 45.20 \\
		& PM-GF & \textbf{38.76} & \textbf{38.86} & \textbf{37.85} & \textbf{38.82} & \textbf{39.43} \\
		\cmidrule{2-7}    \multicolumn{1}{c}{NIQE ↓} & LAM & 3.958 & 3.944 & 3.559 & 3.816 & 3.999 \\
		& PM-GF & \textbf{3.393} & \textbf{3.394} & \textbf{3.323} & \textbf{3.381} & \textbf{3.486} \\
		\bottomrule
	\end{tabular}%
	\label{tab:4.3}%
\end{table}%

\subsection{Tone mapping of high dynamic range images}
High dynamic range (HDR) images capture a wider range of brightness, compared to regular images. Due to the limited dynamic range of displayers or printers, HDR images need to be mapped to low dynamic range (LDR) images for proper viewing or printing. This process is known as tone mapping. In this section, we apply   guided filtering to tone mapping of HDR image for evaluating the performance of ten filters.

In \cite{durand2002fast}, Durand et al. proposed a technique for displaying HDR images, which reduces the contrast while preserving detail. Similar to image detail enhancement, this technique uses edge-preserving ﬁlter to decompose an HDR image $I$ into a base layers $\bar I$ (encoding large-scale variations) and a detail layer $I-\bar I$. The base layer is compressed by a scale factor of $c \in (0,1)$, while the detail layer remains unchanged. The tone-mapped image is produced by adding the compressed base layer back to the unchanged detail layer; i.e., $I_{tone\_mapping}=c \bar I+( I-\bar I)$.

All experiments were conducted using the same  parameter setting as WGIF \cite{li2014weighted}, i.e., $r=16$ and $\varepsilon=1/4$.  We adopt two metrics TMQI \cite{yeganeh2012objective} and BTMQI \cite{gu2016blind} for evaluation. A higher TMQI value and a lower BTMQI value typically indicate higher quality.

\begin{figure}[!htbp]
	\centering
	\subfigure[]{\includegraphics[width=0.18\linewidth]{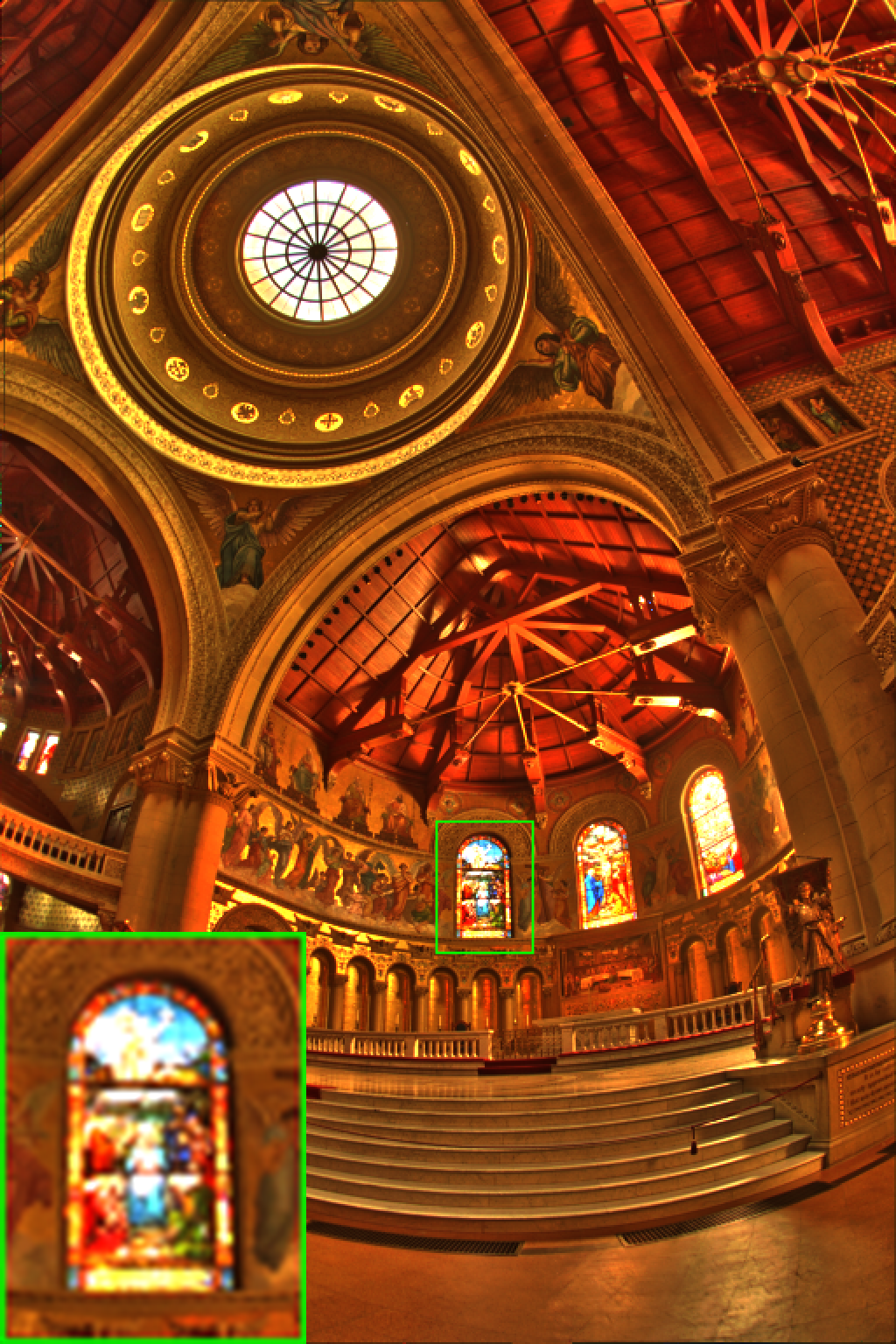}}
	\subfigure[]{\includegraphics[width=0.18\linewidth]{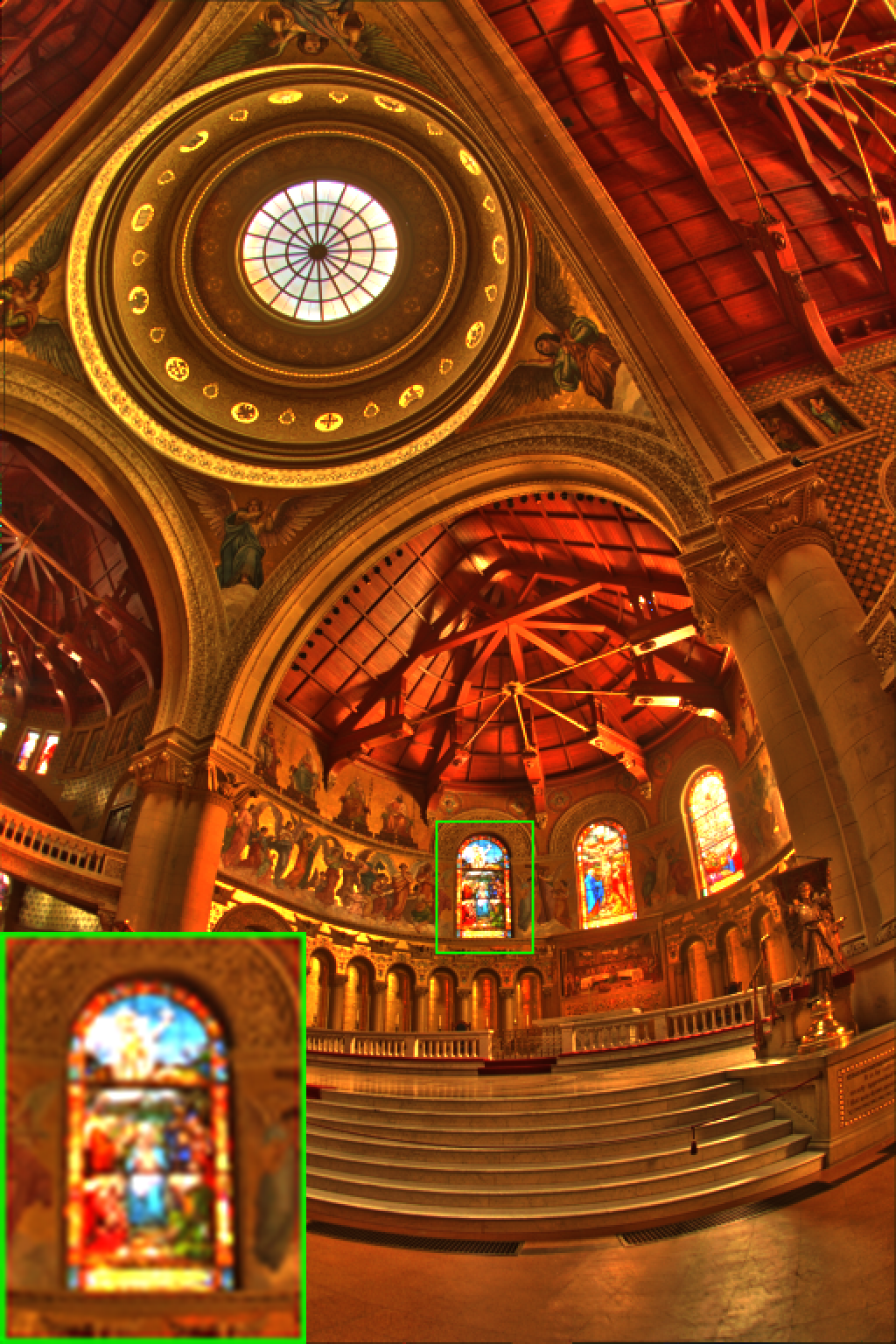}}
	\subfigure[]{\includegraphics[width=0.18\linewidth]{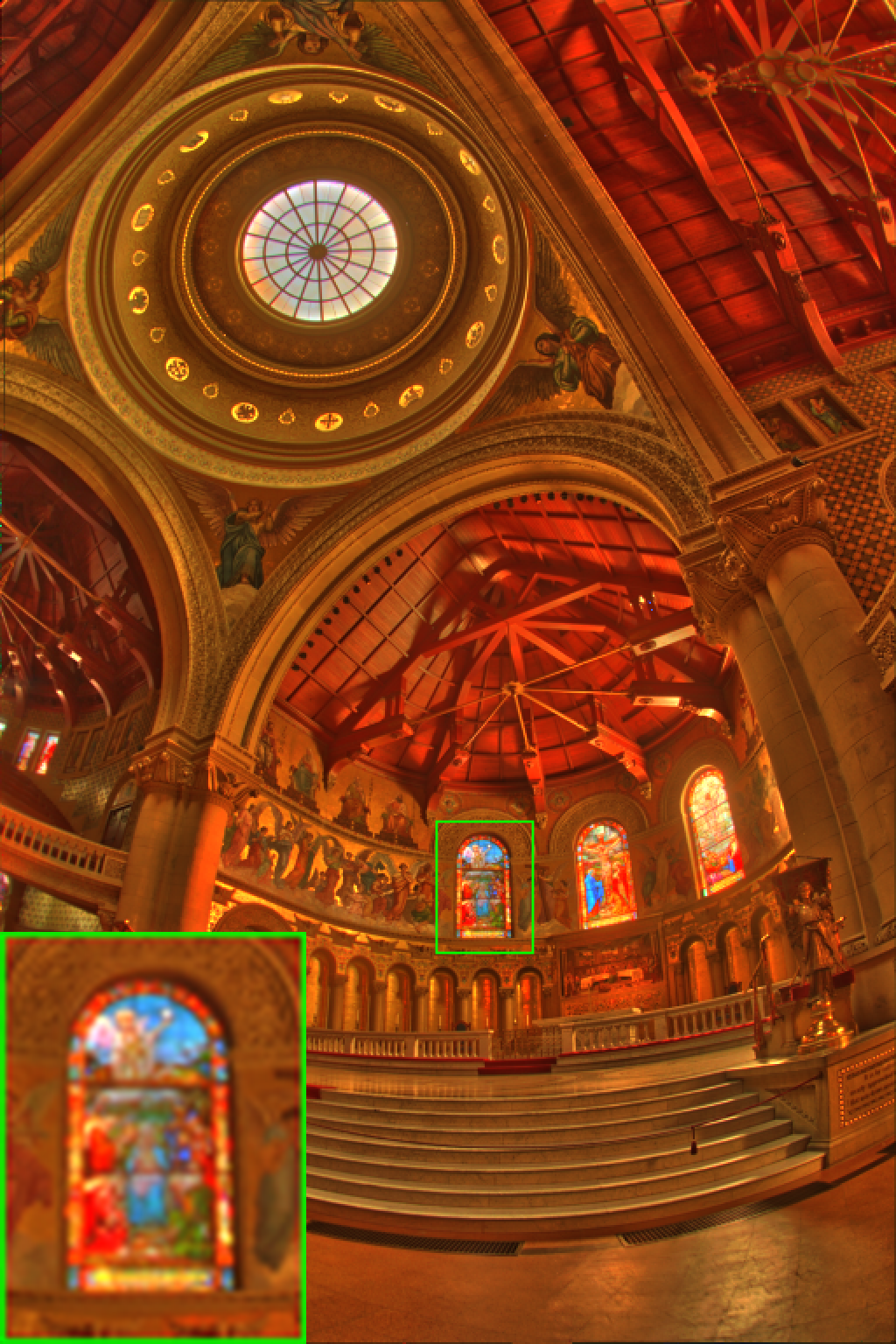}}
	\subfigure[]{\includegraphics[width=0.18\linewidth]{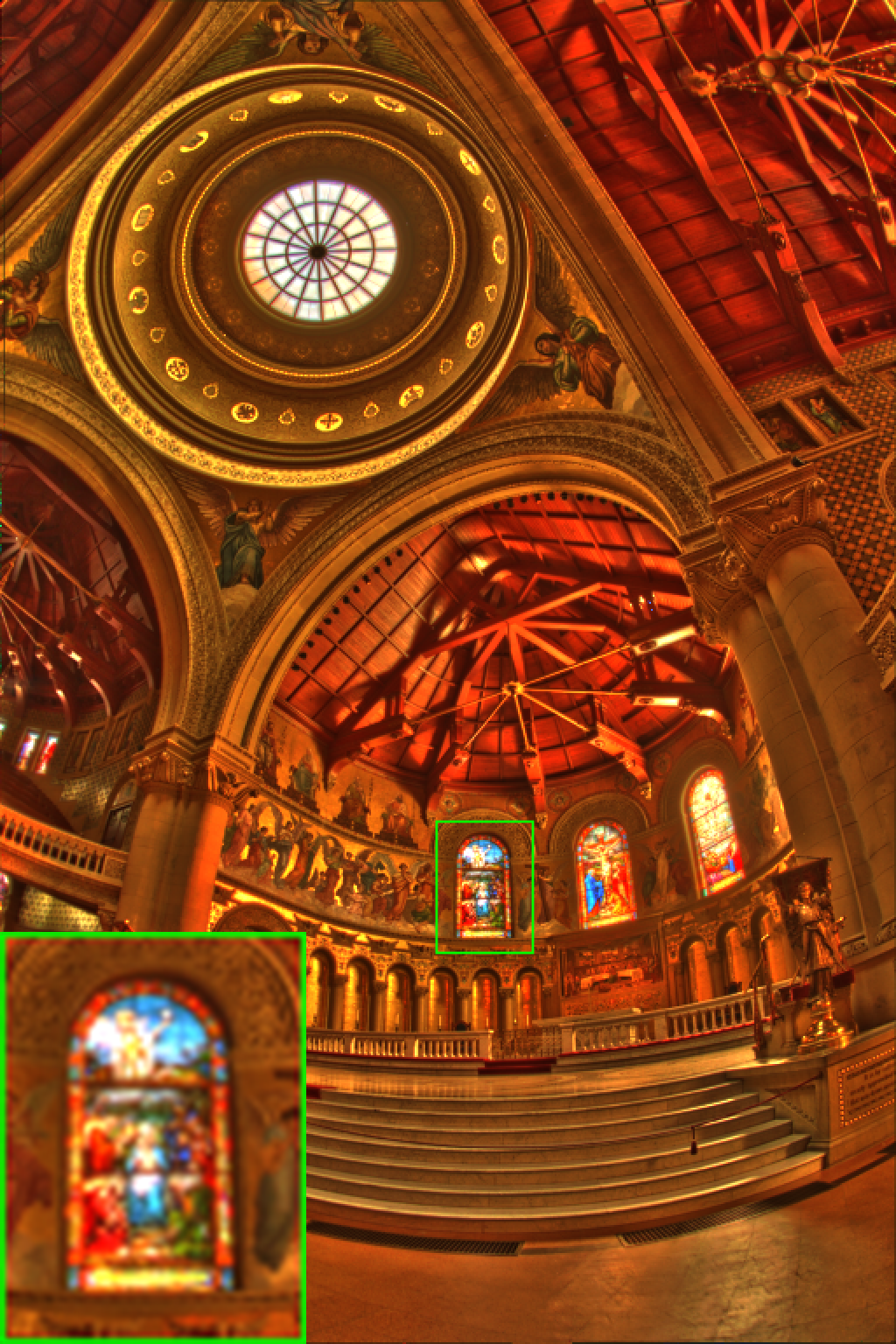}}	
	\subfigure[]{\includegraphics[width=0.18\linewidth]{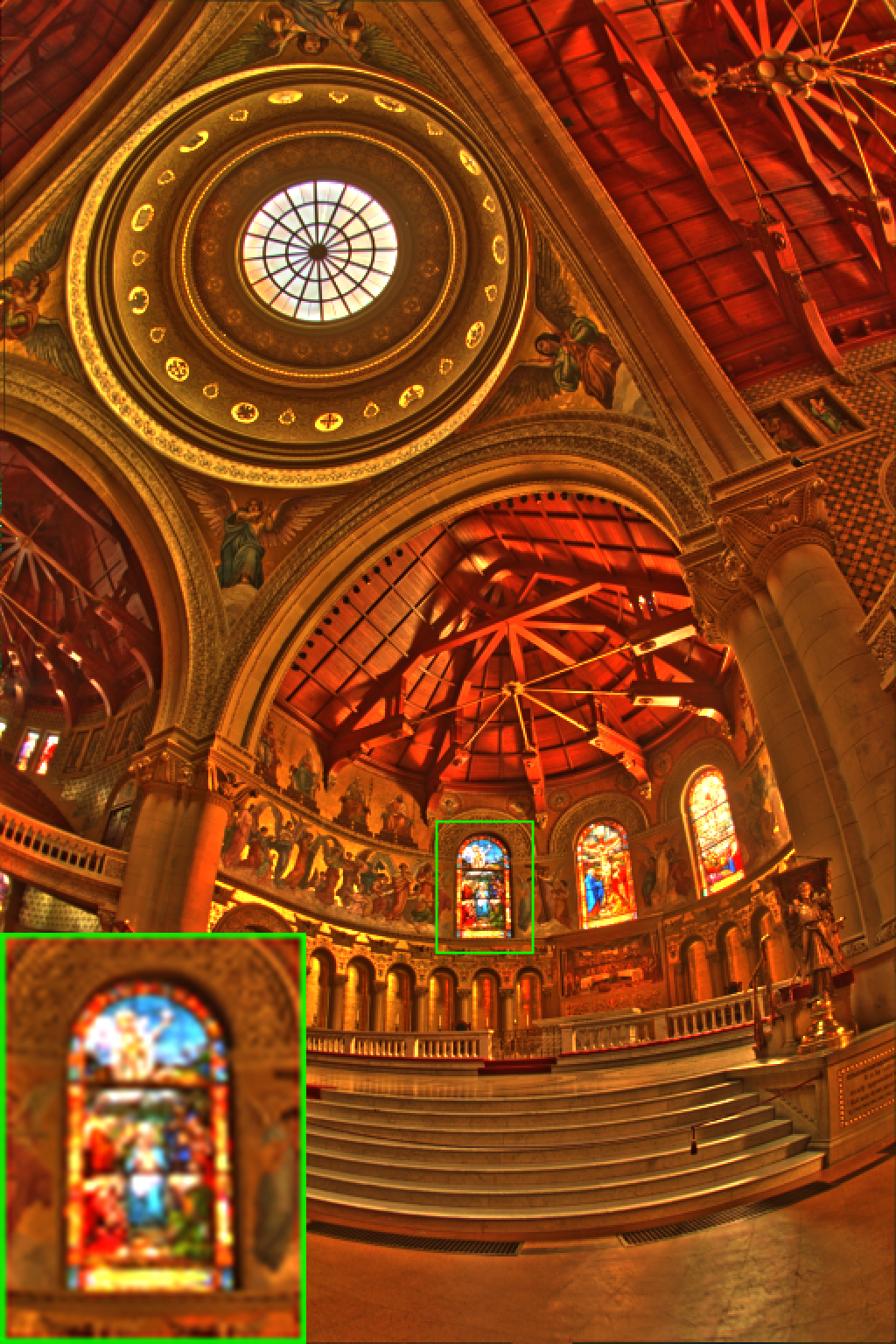}}
	\subfigure[]{\includegraphics[width=0.18\linewidth]{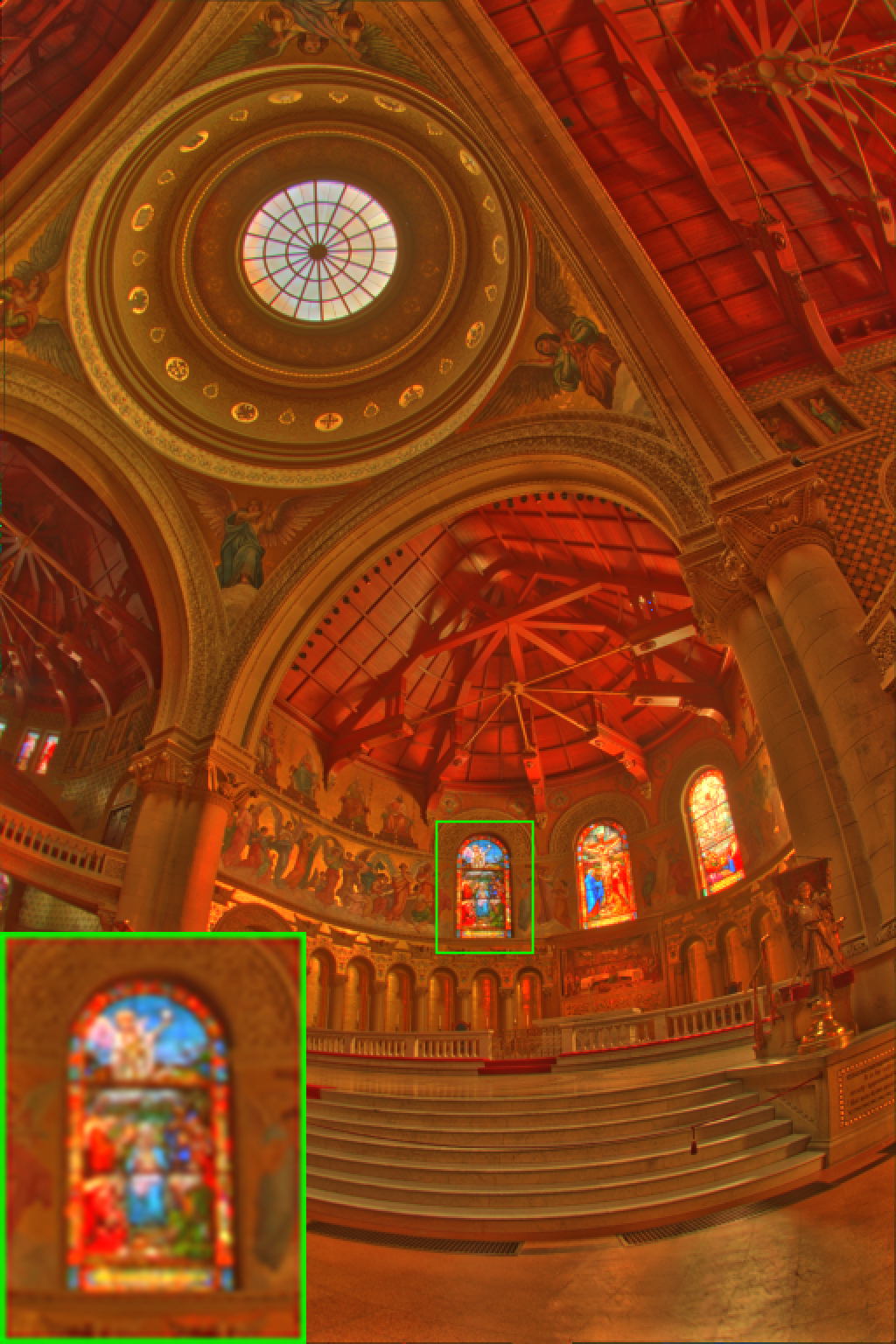}}
	\subfigure[]{\includegraphics[width=0.18\linewidth]{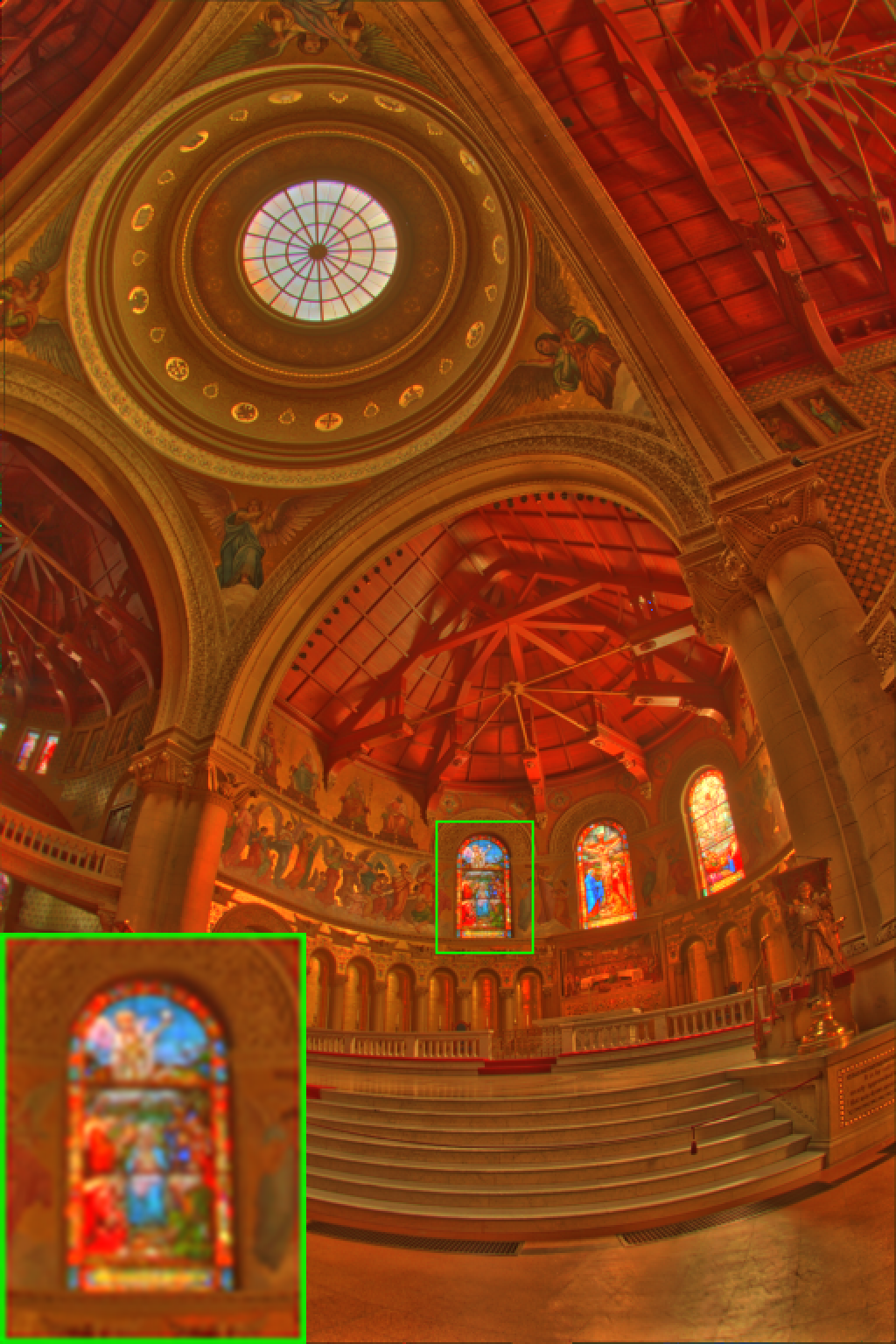}}
	\subfigure[]{\includegraphics[width=0.18\linewidth]{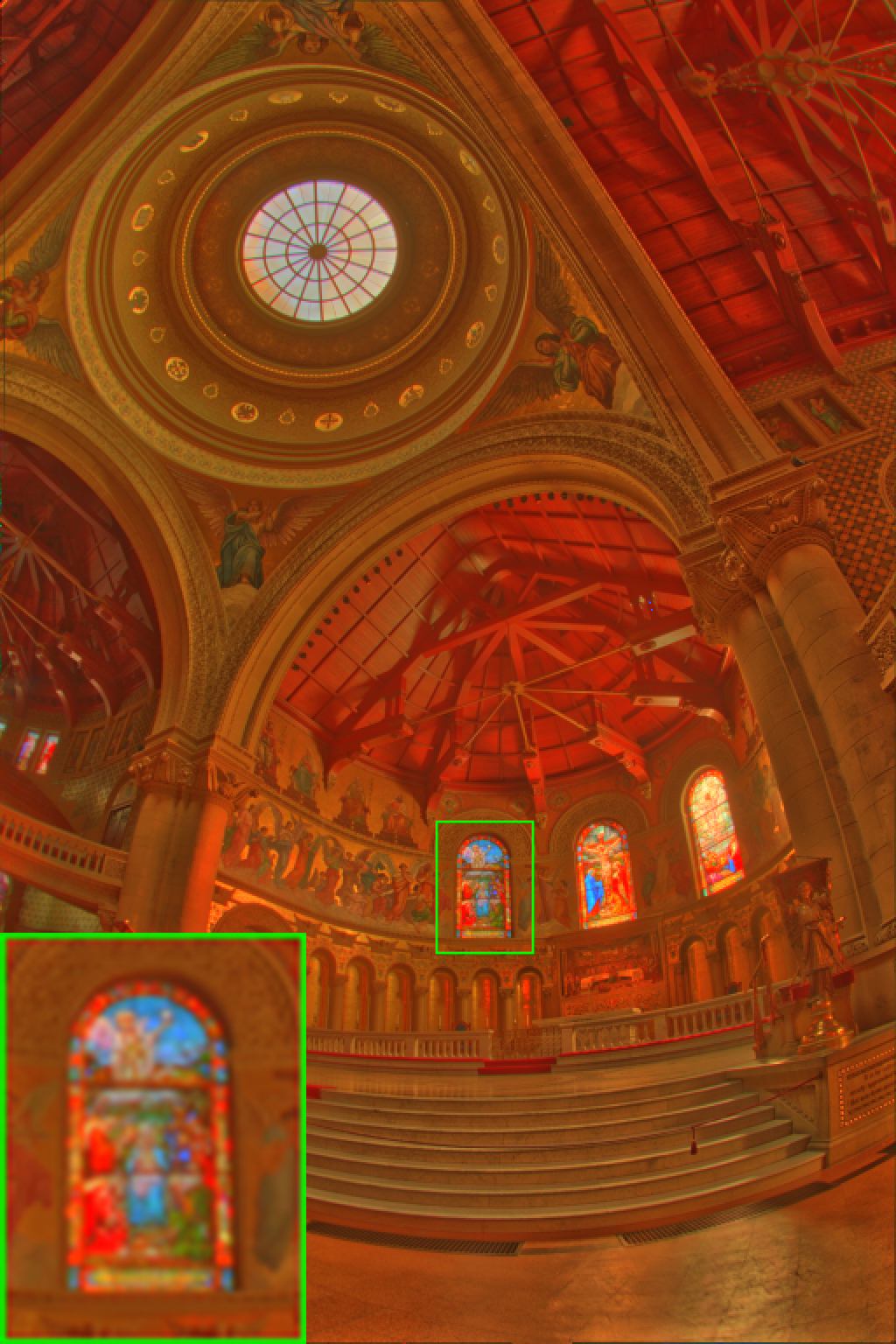}}
	\subfigure[]{\includegraphics[width=0.18\linewidth]{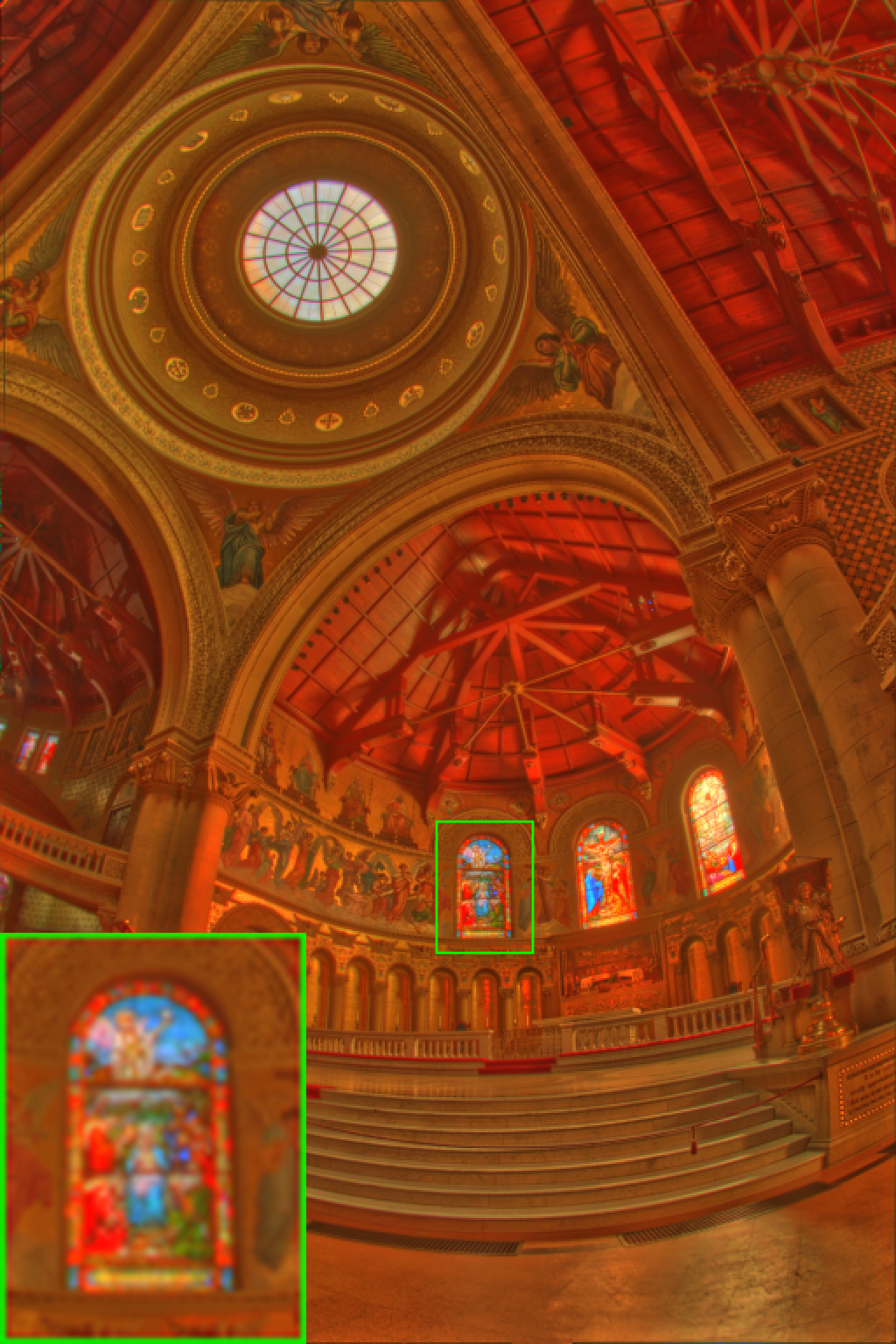}}	
	\subfigure[]{\includegraphics[width=0.18\linewidth]{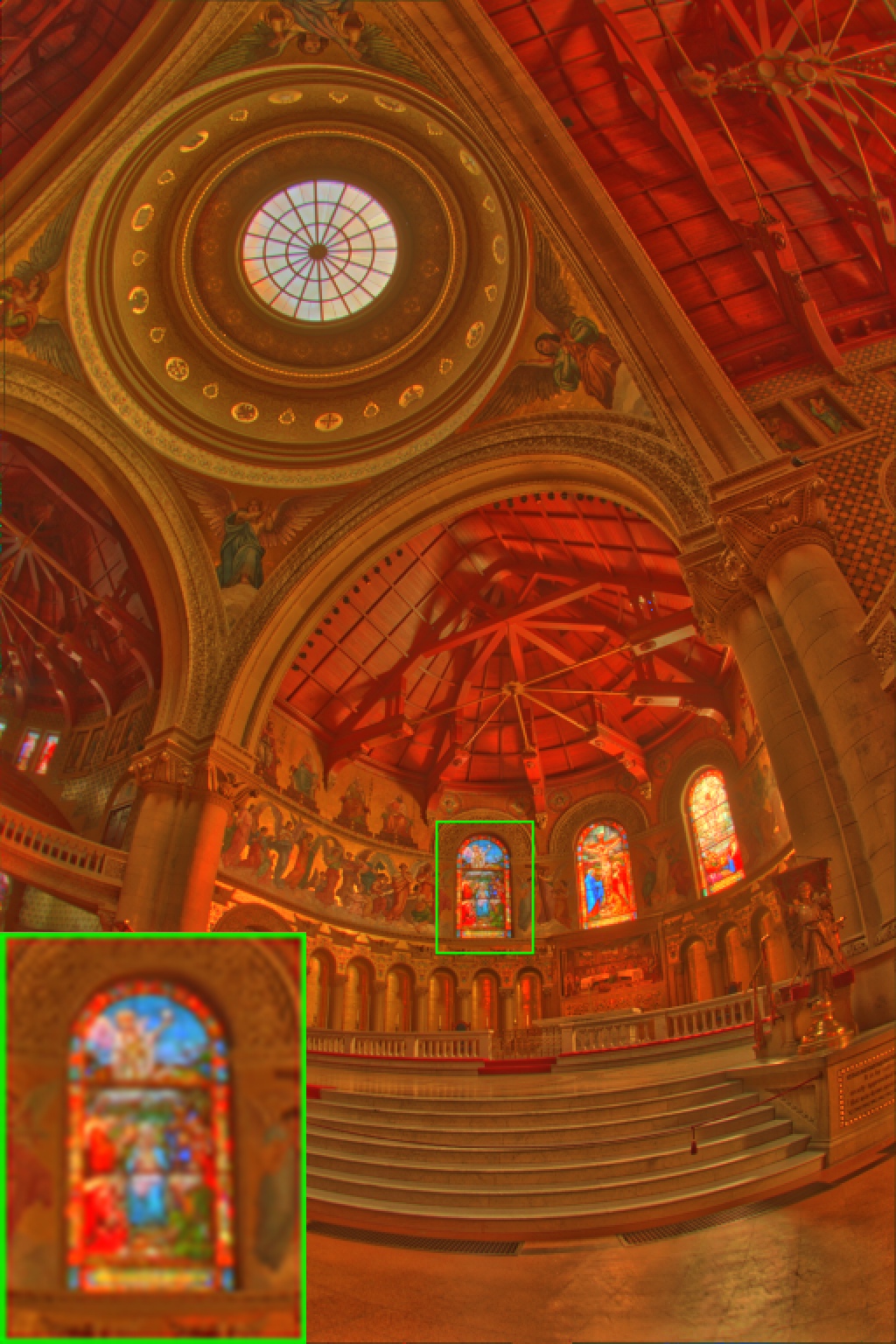}}
	\caption{Comparison of HDR image ``memorial'' tone mapping results by different filters. LDR images: (a)-(e) GIF, WGIF, GGIF, SKWGIF, RDWGIF. (f)-(j) GH-GIF, GH-WGIF, GH-GGIF, GH-SKWGIF, GH-RDWGIF.}
	\label{fig:4.4-1}
\end{figure}

\begin{figure}[!htbp]
	\centering
	\subfigure[]{\includegraphics[width=0.18\linewidth]{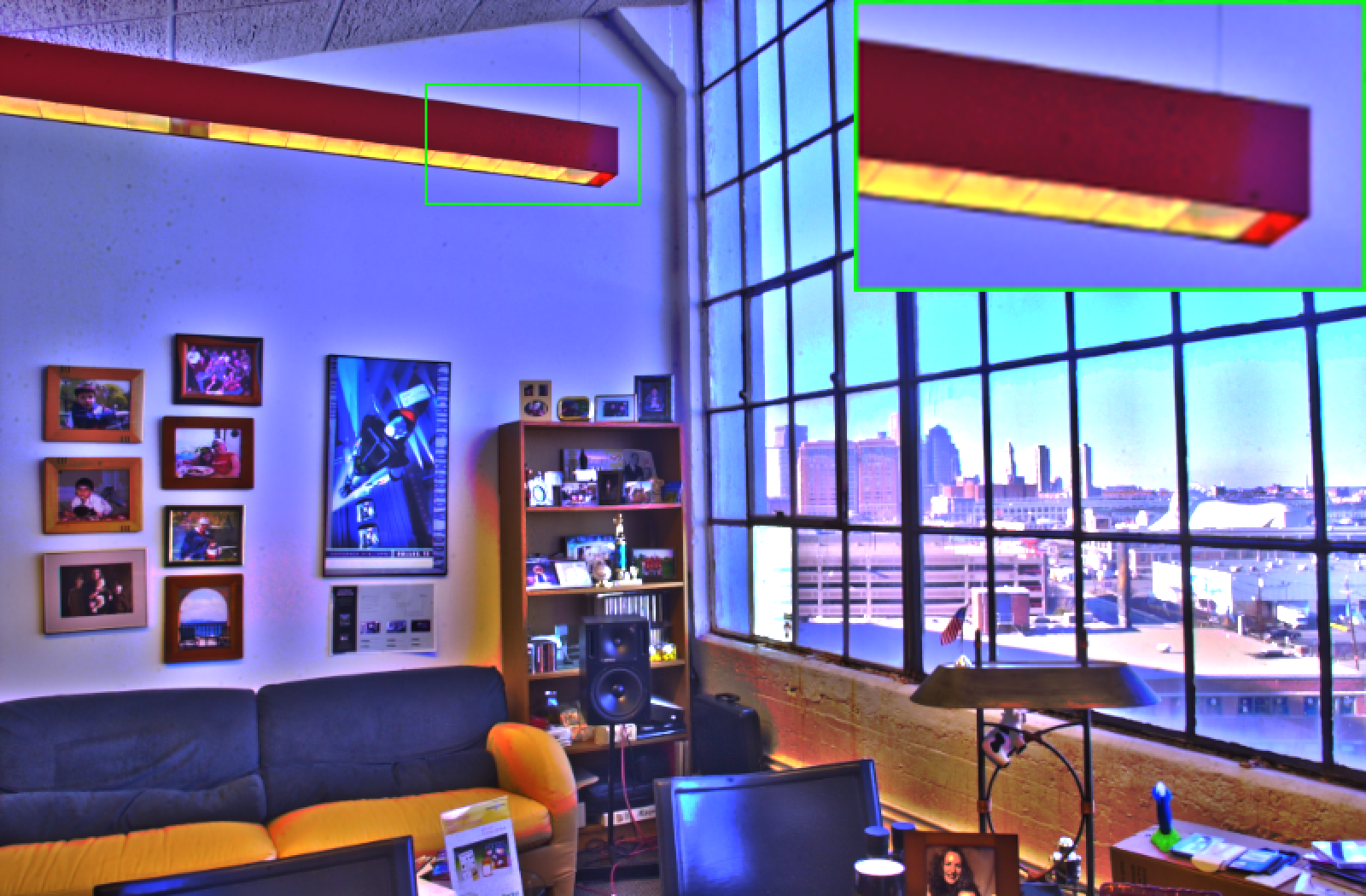}}
	\subfigure[]{\includegraphics[width=0.18\linewidth]{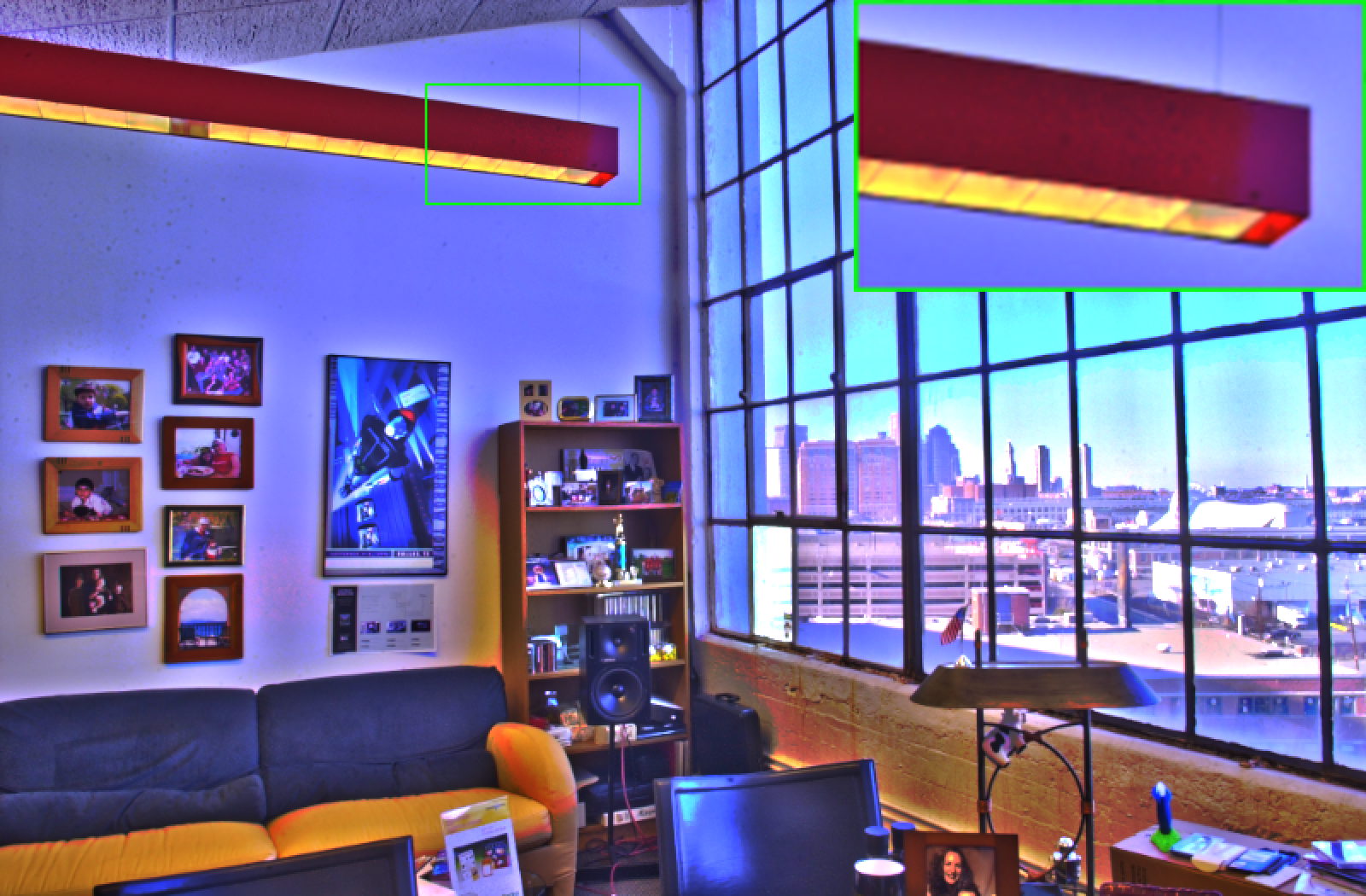}}
	\subfigure[]{\includegraphics[width=0.18\linewidth]{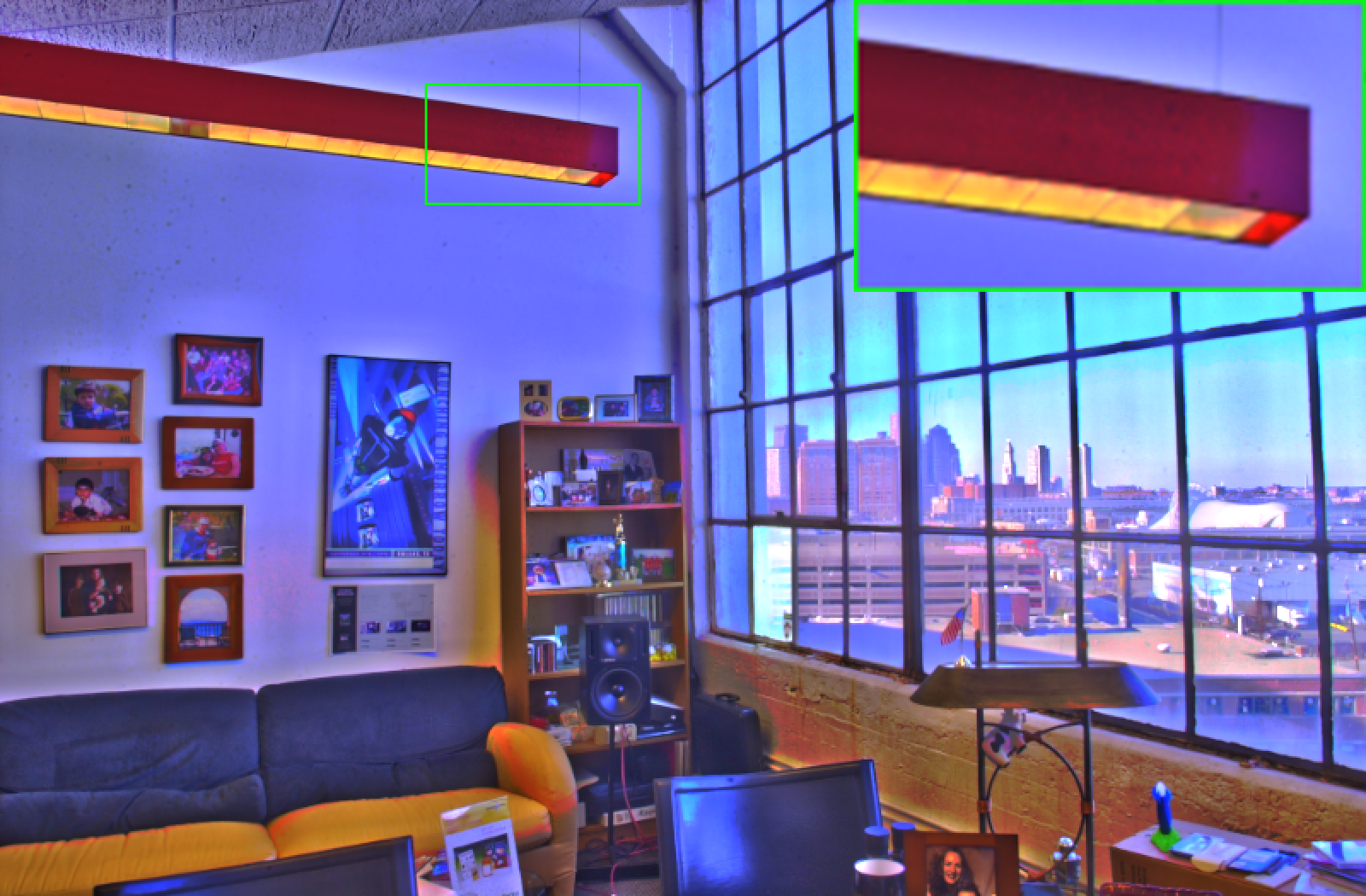}}
	\subfigure[]{\includegraphics[width=0.18\linewidth]{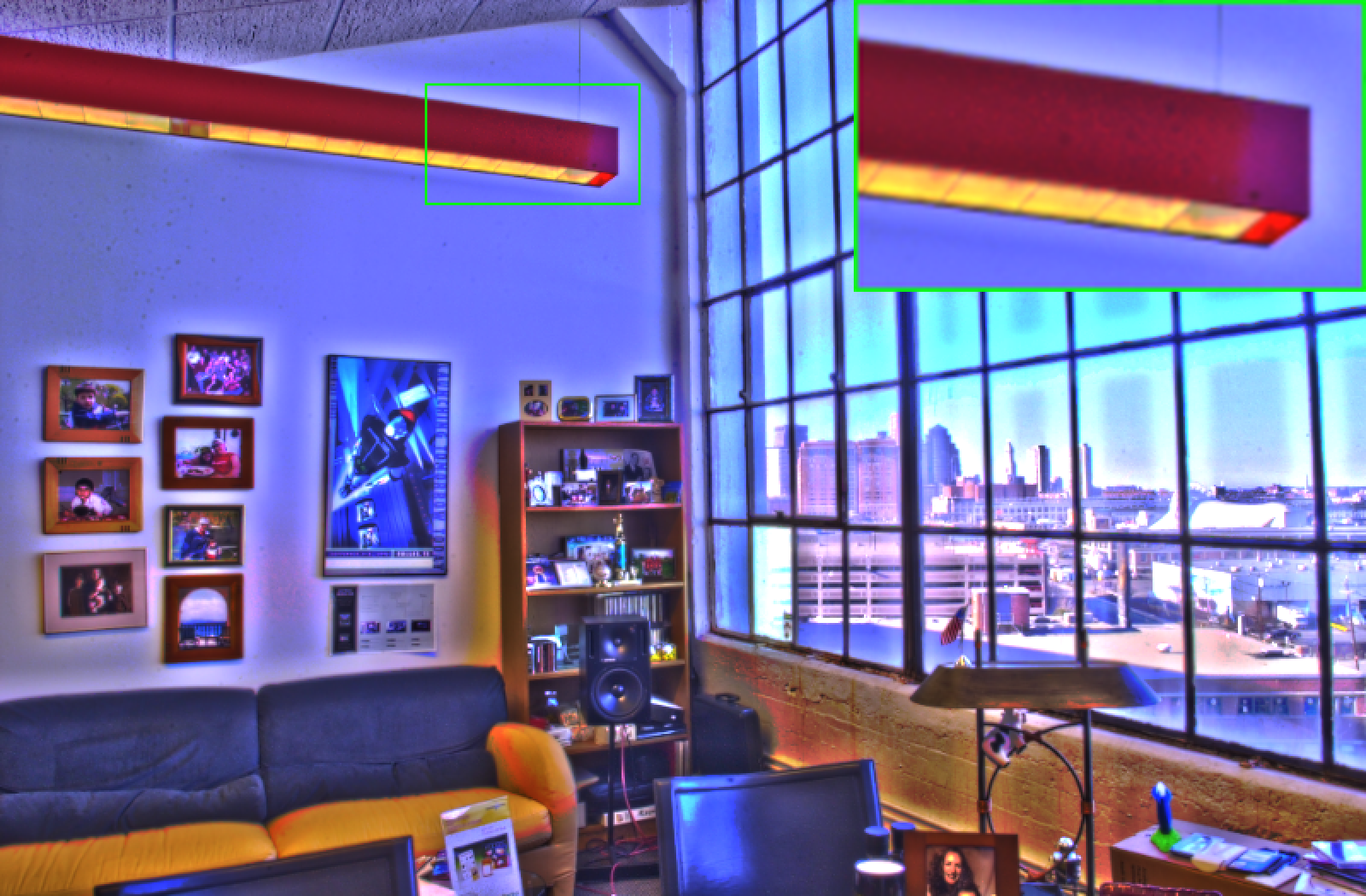}}	
	\subfigure[]{\includegraphics[width=0.18\linewidth]{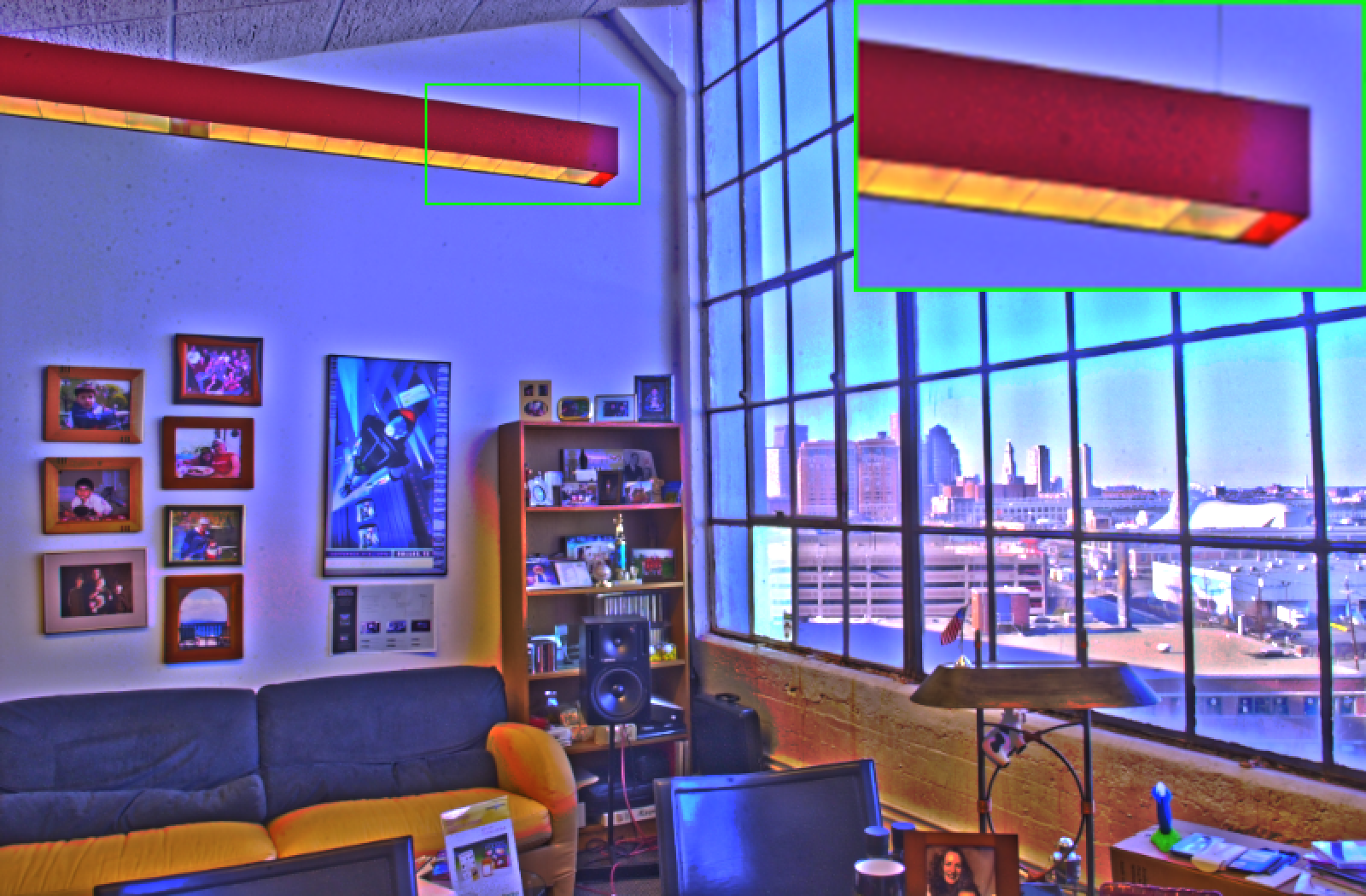}}
	\subfigure[]{\includegraphics[width=0.18\linewidth]{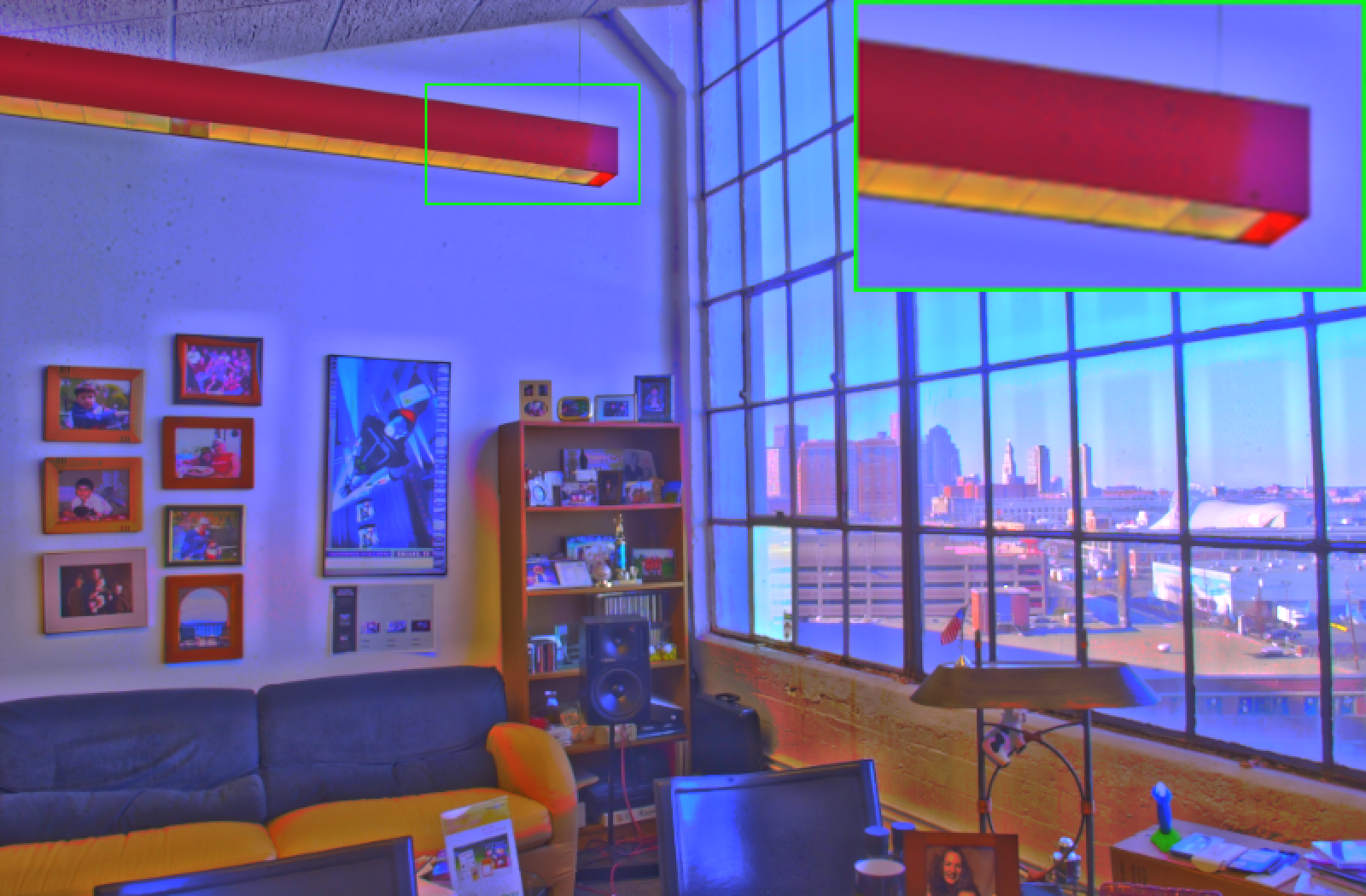}}
	\subfigure[]{\includegraphics[width=0.18\linewidth]{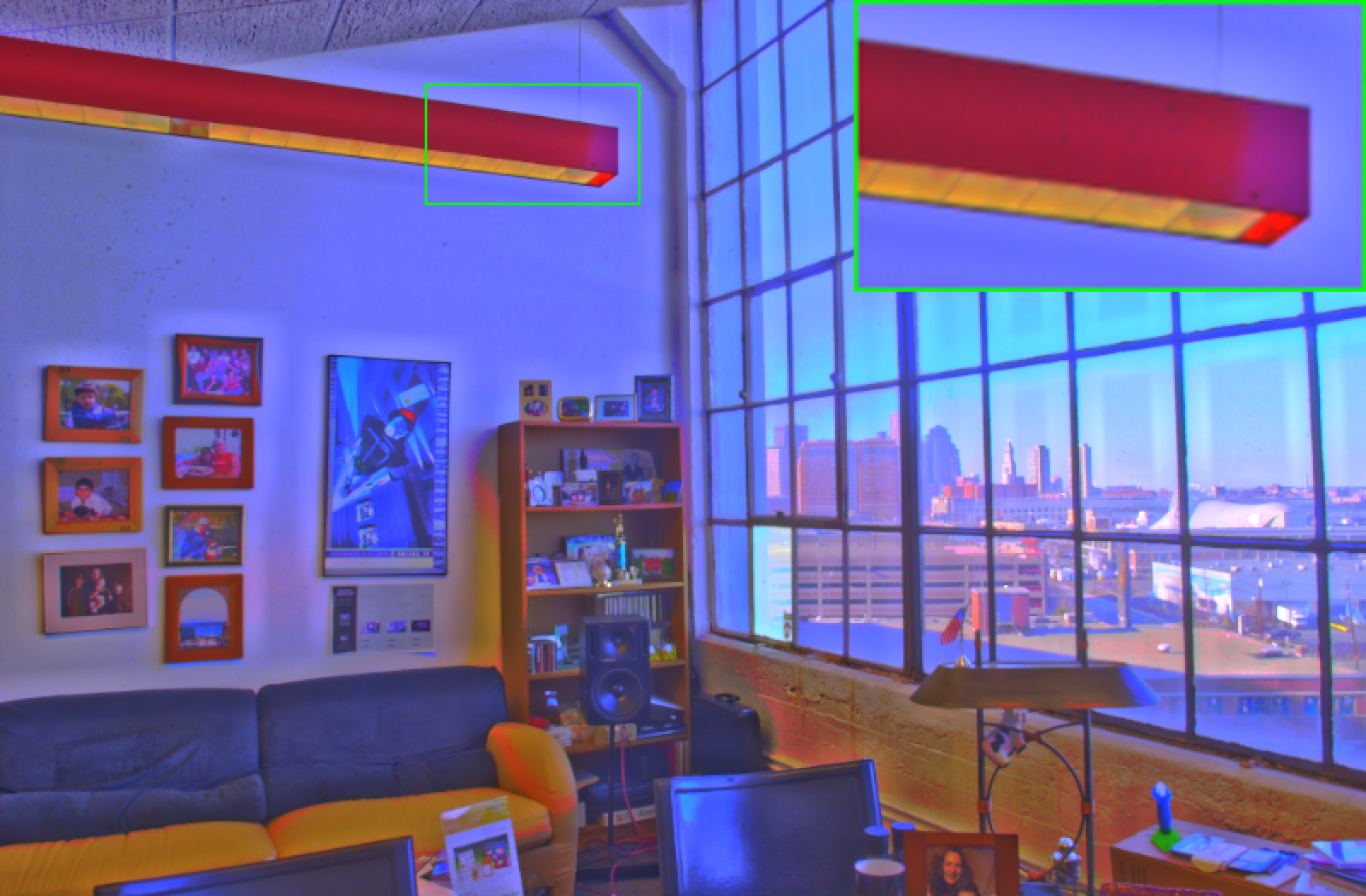}}
	\subfigure[]{\includegraphics[width=0.18\linewidth]{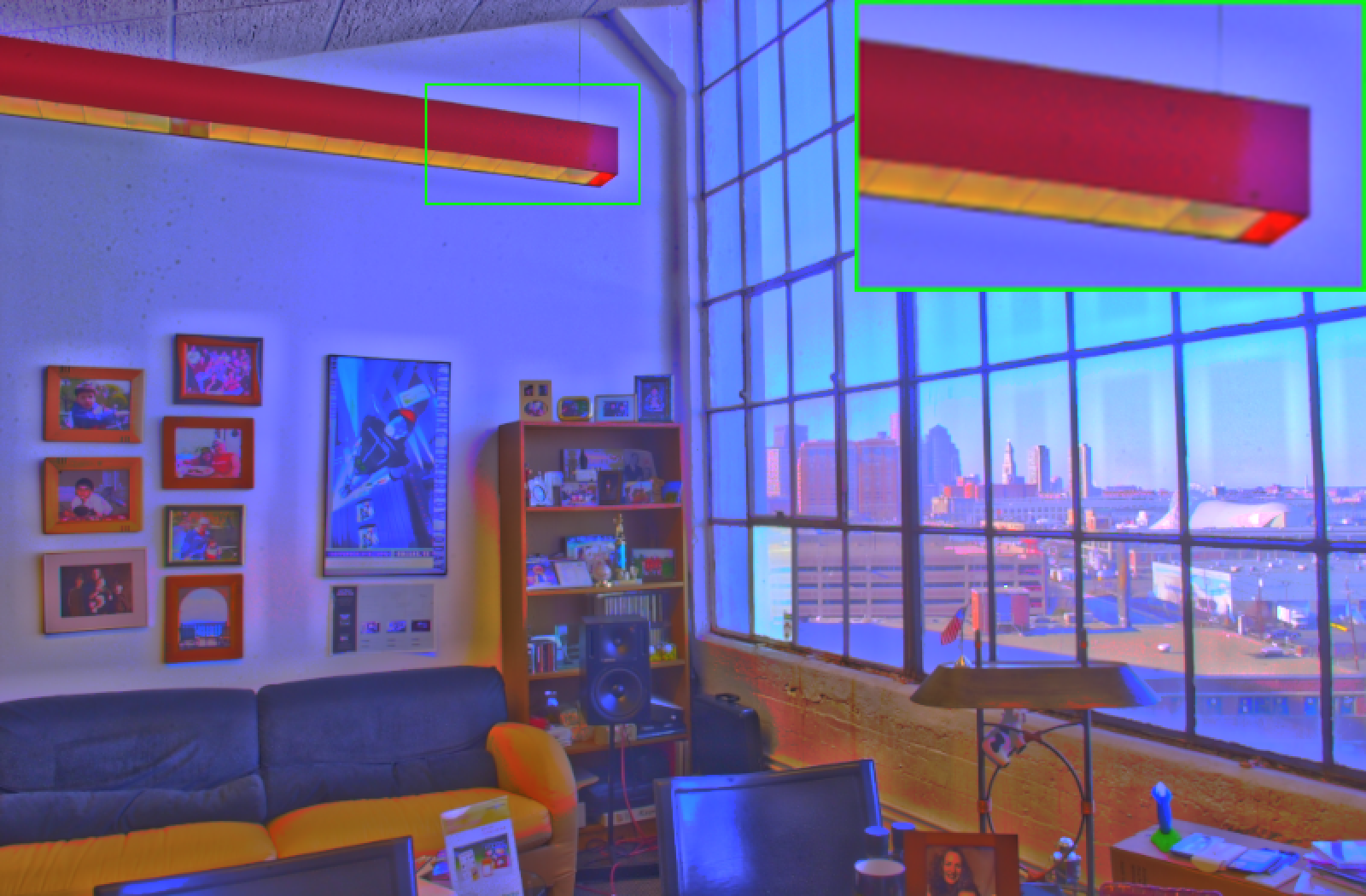}}
	\subfigure[]{\includegraphics[width=0.18\linewidth]{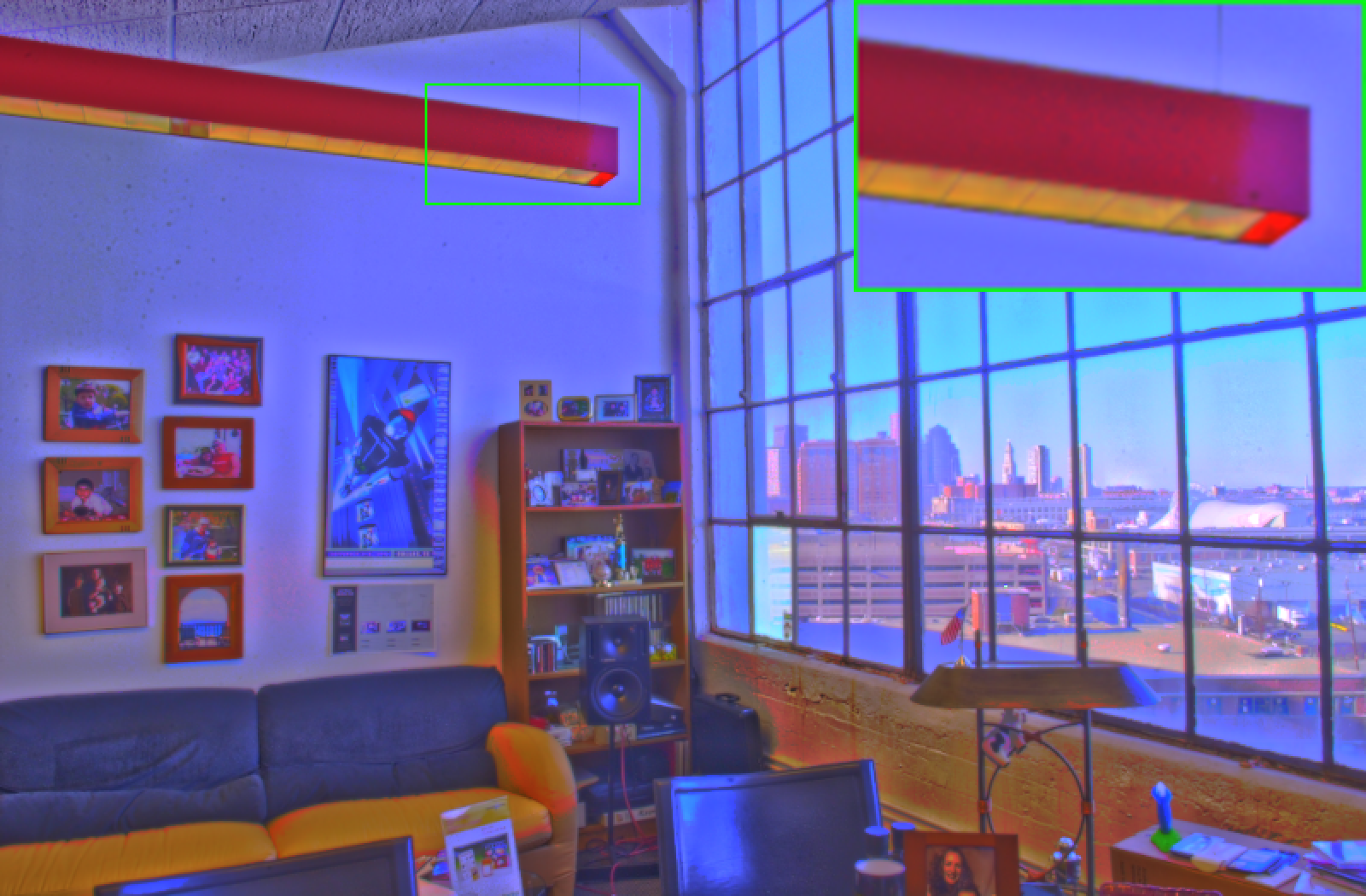}}	
	\subfigure[]{\includegraphics[width=0.18\linewidth]{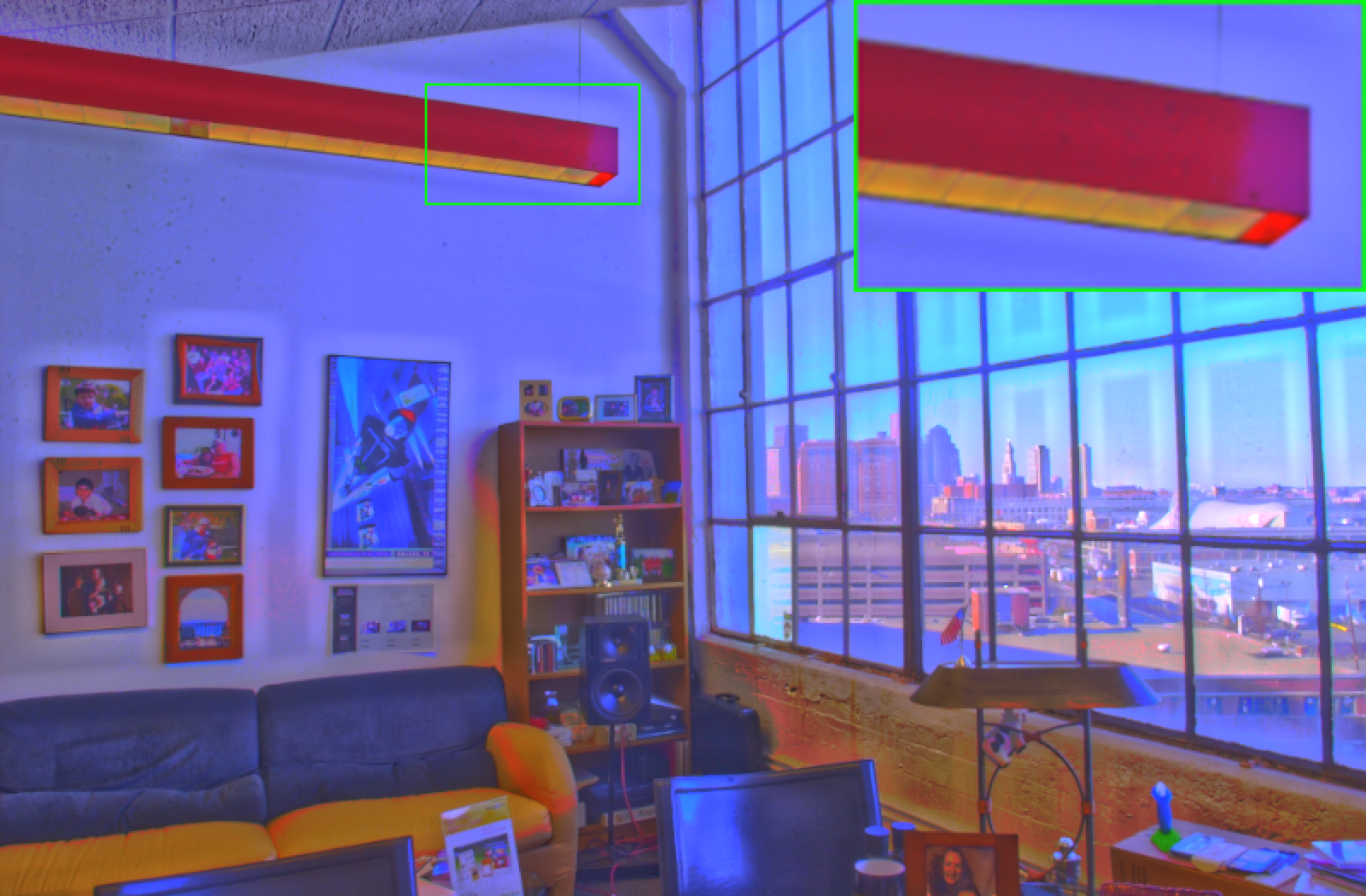}}
	\caption{Comparison of HDR image "smalloffice" tone mapping results by different filters. LDR images: (a)-(e) GIF, WGIF, GGIF, SKWGIF, RDWGIF. (f)-(j) GH-GIF, GH-WGIF, GH-GGIF, GH-SKWGIF, GH-RDWGIF.}
	\label{fig:4.4-2}
\end{figure}

Fig. \ref{fig:4.4-1} and Fig. \ref{fig:4.4-2} display the tone mapped results by using ten GIFs for two HDR images. As shown in the local magnifications of Fig. \ref{fig:4.4-1} and \ref{fig:4.4-2}, PM-GF-based GIFs exhibit more natural tone mapping and richer detail features in comparison to their counterparts. Table \ref{tab:4.4} presents the corresponding TMQI and BTMQI values. It is evident that PM-GF-based GIFs outperform their counterparts, achieving better TMQI and BTMQI scores.

\begin{table}[!htbp]
	\centering
	\caption{The TMQI \cite{yeganeh2012objective}, and BTMQI\cite{gu2016blind} values of different filters in "memorial" and "smalloffice". Best values are highlighted in bold, (↑): the higher, the better. (↓): the lower, the better.}
	\begin{tabular}{llllllll}
		\toprule
		Image & Metric & Model & GIF   & WGIF  & GGIF  & SKWGIF & RDWGIF \\
		\midrule
		memorial & TMQI ↑ & LAM   & 0.8464 & 0.8474 & \textbf{0.8537} & 0.8407 & 0.8420 \\
		&       & PM-GF & \textbf{0.8471} & \textbf{0.8446} & 0.8347 & \textbf{0.8436} & \textbf{0.8463} \\
		\cmidrule{3-8}          & BTMQI ↓ & LAM   & 4.141 & 4.145 & 4.058 & 4.193 & 4.246 \\
		&       & PM-GF & \textbf{3.333} & \textbf{3.350} & \textbf{3.330} & \textbf{3.026} & \textbf{3.001} \\
		\midrule
		smalloffice & TMQI ↑ & LAM   & 0.8695 & 0.8866 & \textbf{0.9443} & 0.8764 & 0.9108 \\
		&       & PM-GF & \textbf{0.9447} & \textbf{0.9427} & 0.9338 & \textbf{0.9381} & \textbf{0.937} \\
		\cmidrule{3-8}          & BTMQI ↓ & LAM   & 4.004 & 3.695 & 2.128 & 3.930  & 2.744 \\
		&       & PM-GF & \textbf{1.675} & \textbf{1.722} & \textbf{2.002} & \textbf{1.888} & \textbf{1.868} \\
		\bottomrule
	\end{tabular}%
	\label{tab:4.4}%
\end{table}%

\subsection{Single image haze removal}
In \cite{he2010single}, He et al. proposed a haze removal technique based on the atmospheric scattering model and the dark channel prior. In \cite{he2012guided}, the GIF was used to refine the transmission map in dehazing algorithms. By using the original image as a guidance one, the GIF adjusts the transmission map estimation based on the structural information of the original image, thereby capturing changes in transmission rates more accurately. Since haze typically results in loss of image details and blurring, the GIF contributes to preservation of the details in the image, maintaining its clarity and quality while removing haze.

In this experiment, we use the GIFs to refine the transmission map  in single image haze removal for performance comparison between two types of GIFs (LAM-based and PM-GF-based). A natural foggy image from the LIVE image defogging dataset \footnote{\href{https://live.ece.utexas.edu/research/fog/fade\_defade.html}{https://live.ece.utexas.edu/research/fog/fade\_defade.html}} is chosen as the test data. Following \cite{he2012guided}, the radius $r$ of local window and the regularization parameter $\varepsilon$ are set to 20 and $10^{-3}$,  respectively.

\begin{figure}[!htbp]
	\centering		
	
	\includegraphics[width=0.15\linewidth]{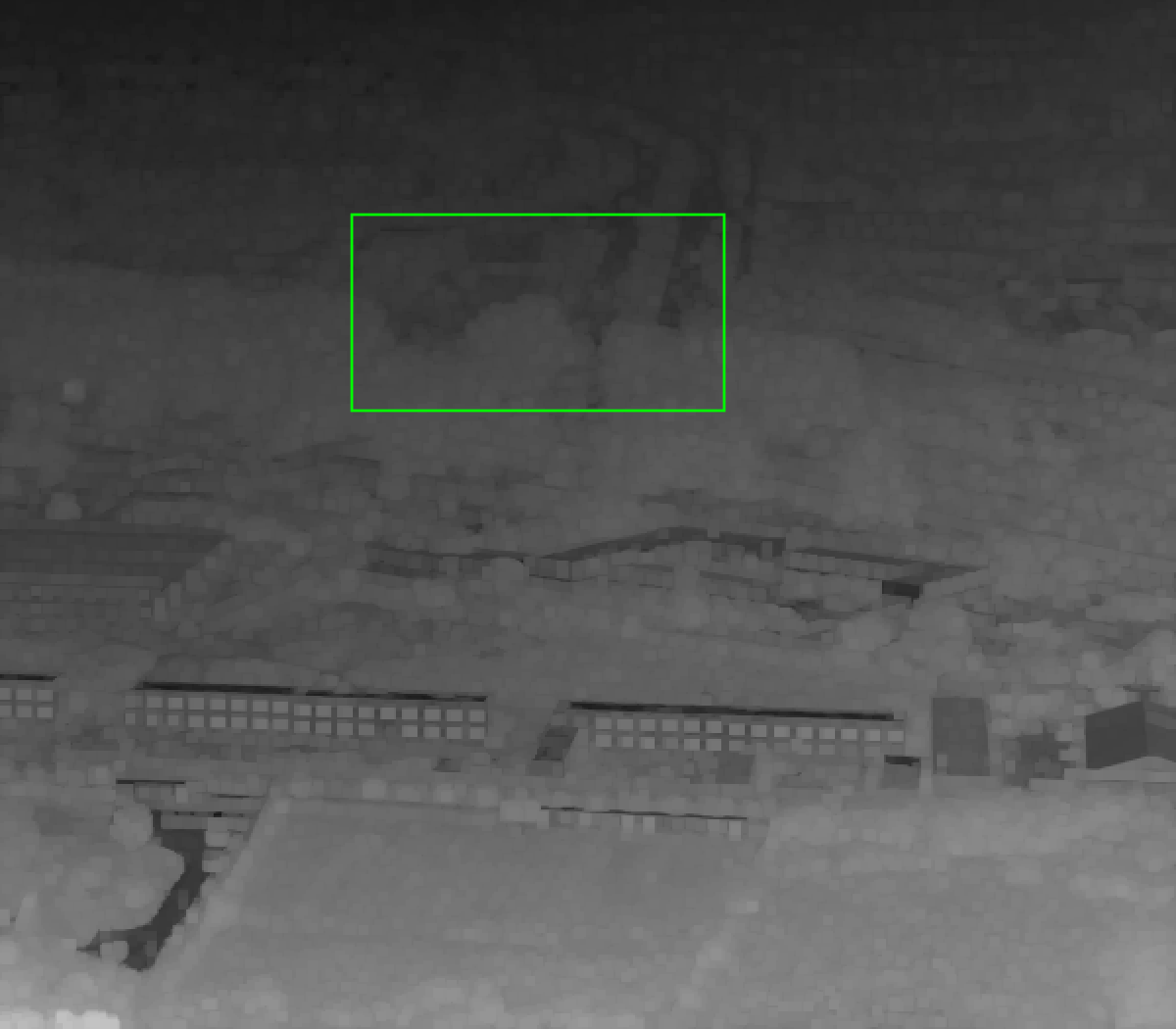}
	\includegraphics[width=0.15\linewidth]{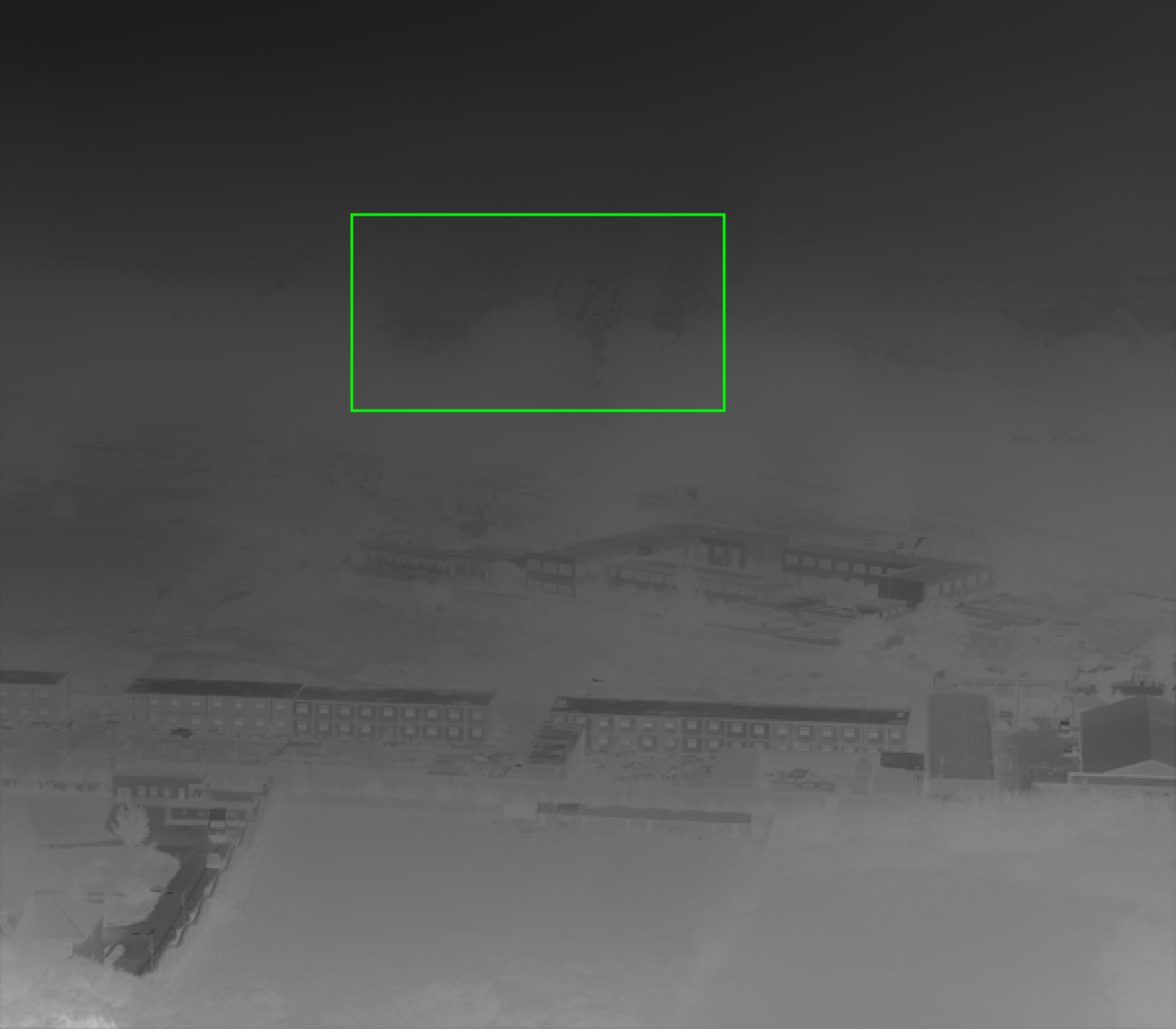}
	\includegraphics[width=0.15\linewidth]{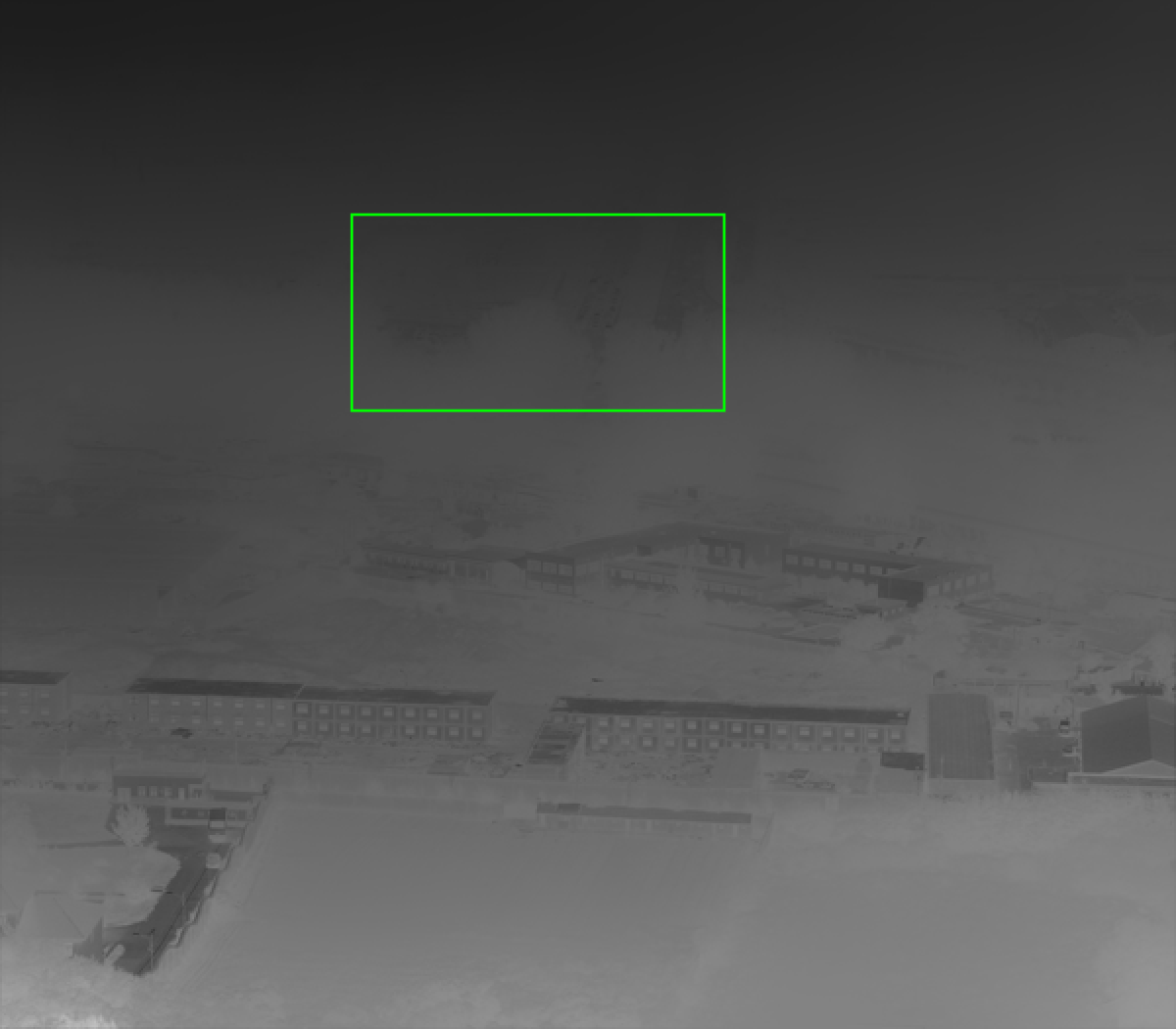}
	\includegraphics[width=0.15\linewidth]{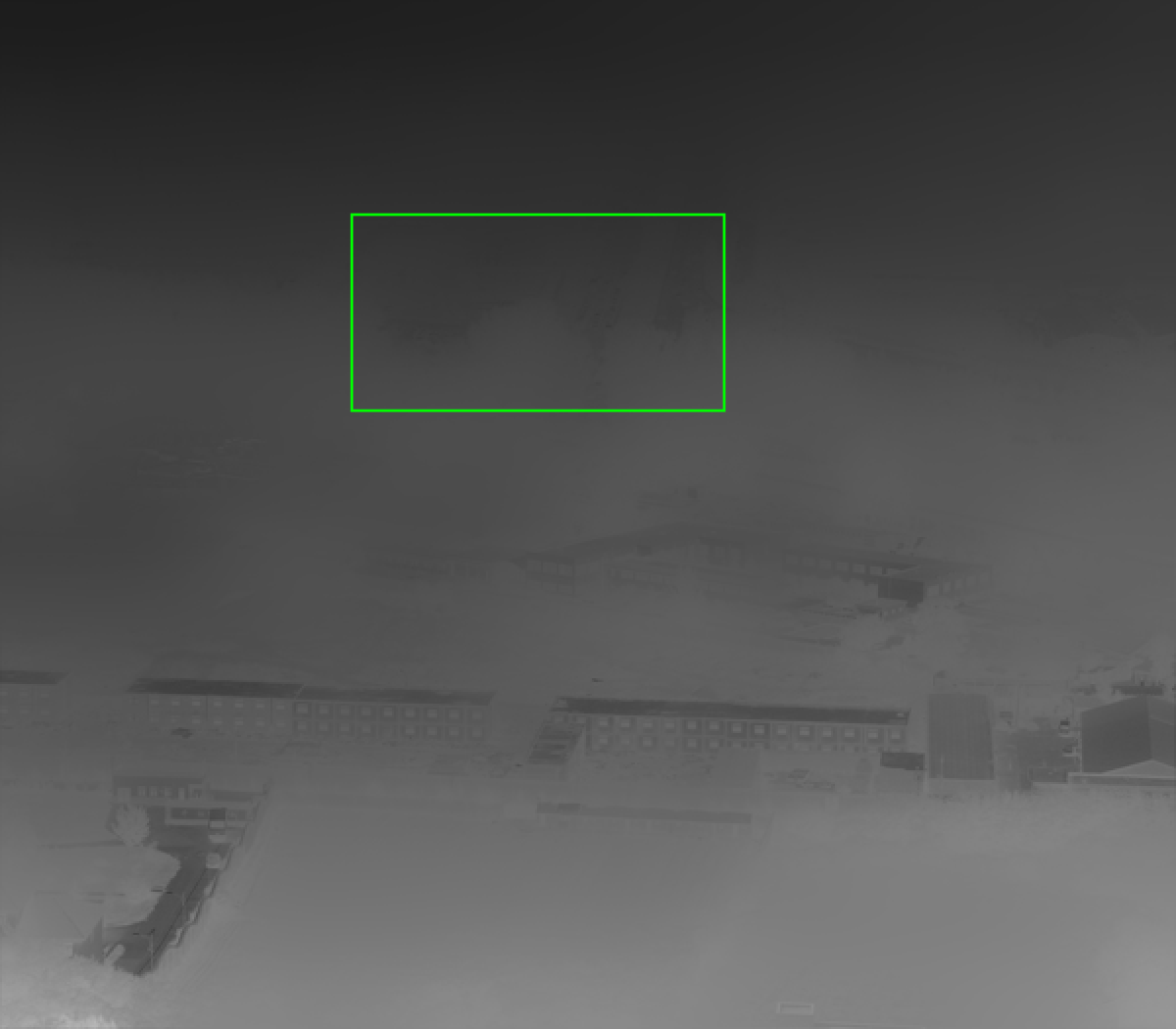}
	\includegraphics[width=0.15\linewidth]{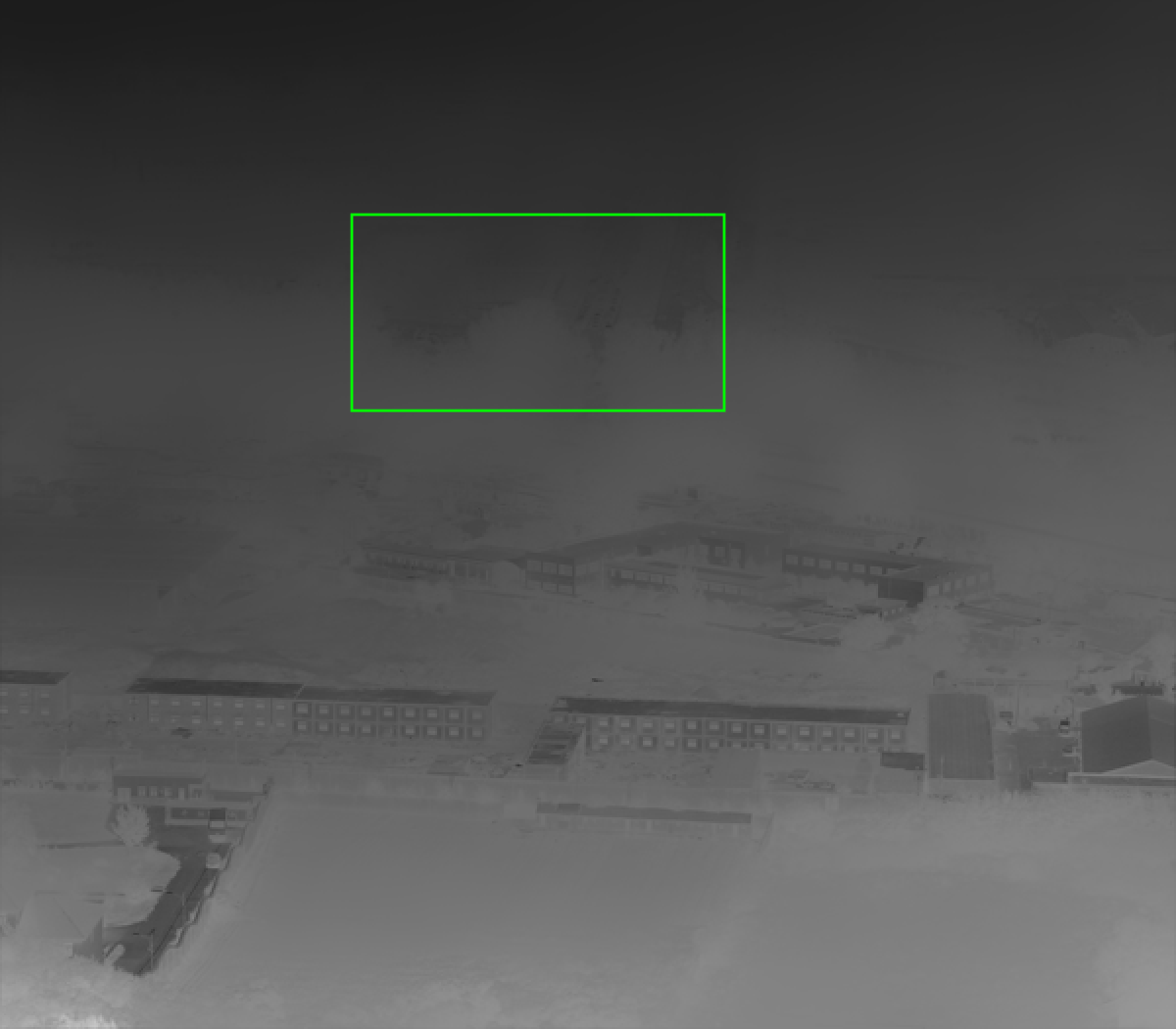}
	\includegraphics[width=0.15\linewidth]{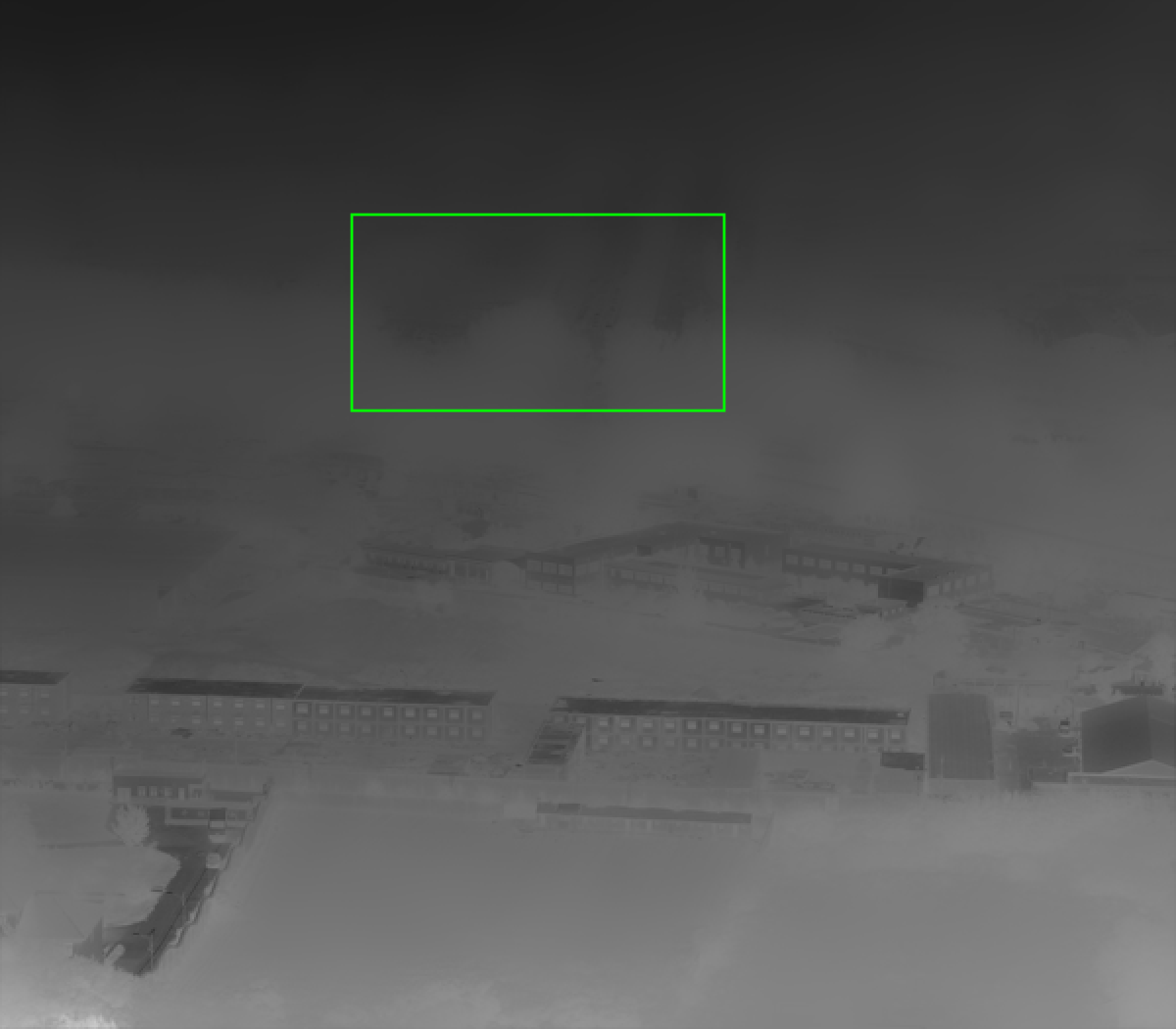}
	
	\subfigure[]{\includegraphics[width=0.15\linewidth]{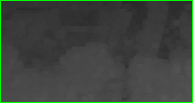}}
	\subfigure[]{\includegraphics[width=0.15\linewidth]{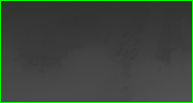}}
	\subfigure[]{\includegraphics[width=0.15\linewidth]{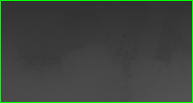}}
	\subfigure[]{\includegraphics[width=0.15\linewidth]{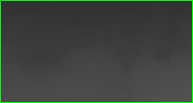}}
	\subfigure[]{\includegraphics[width=0.15\linewidth]{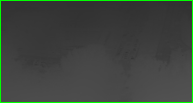}}
	\subfigure[]{\includegraphics[width=0.15\linewidth]{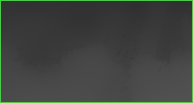}}

	\hspace{0.15\linewidth}
	\includegraphics[width=0.15\linewidth]{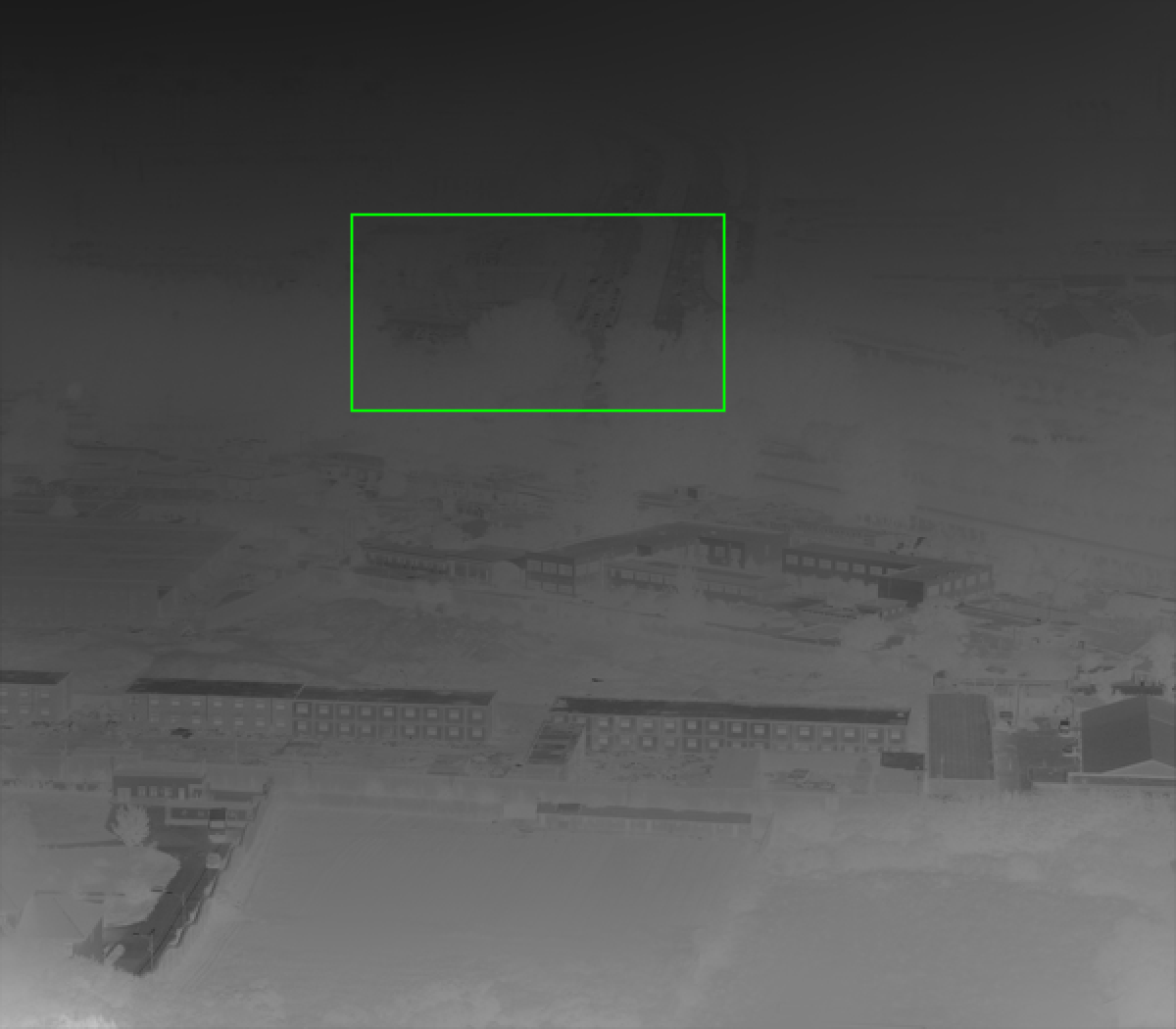}
	\includegraphics[width=0.15\linewidth]{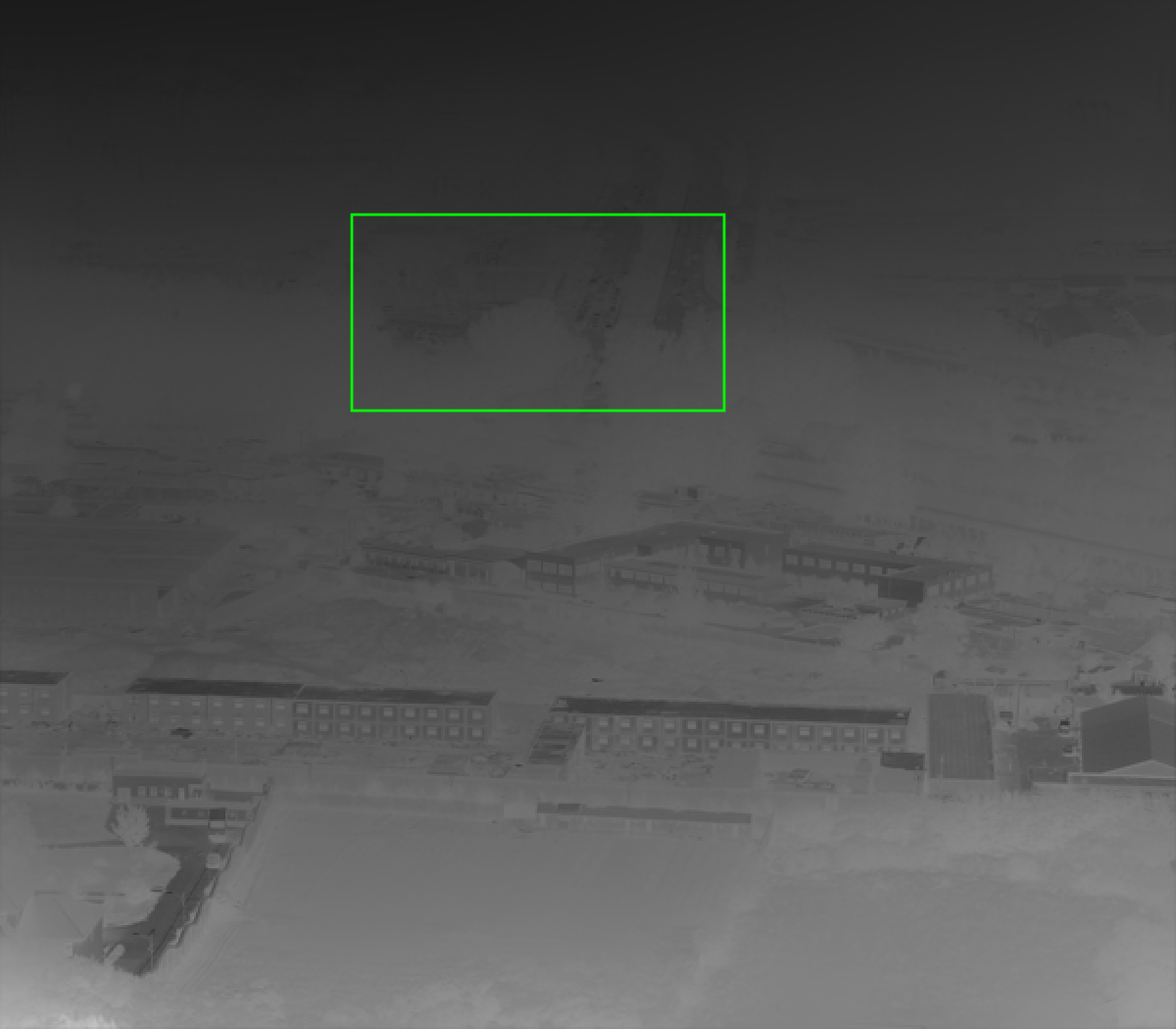}
	\includegraphics[width=0.15\linewidth]{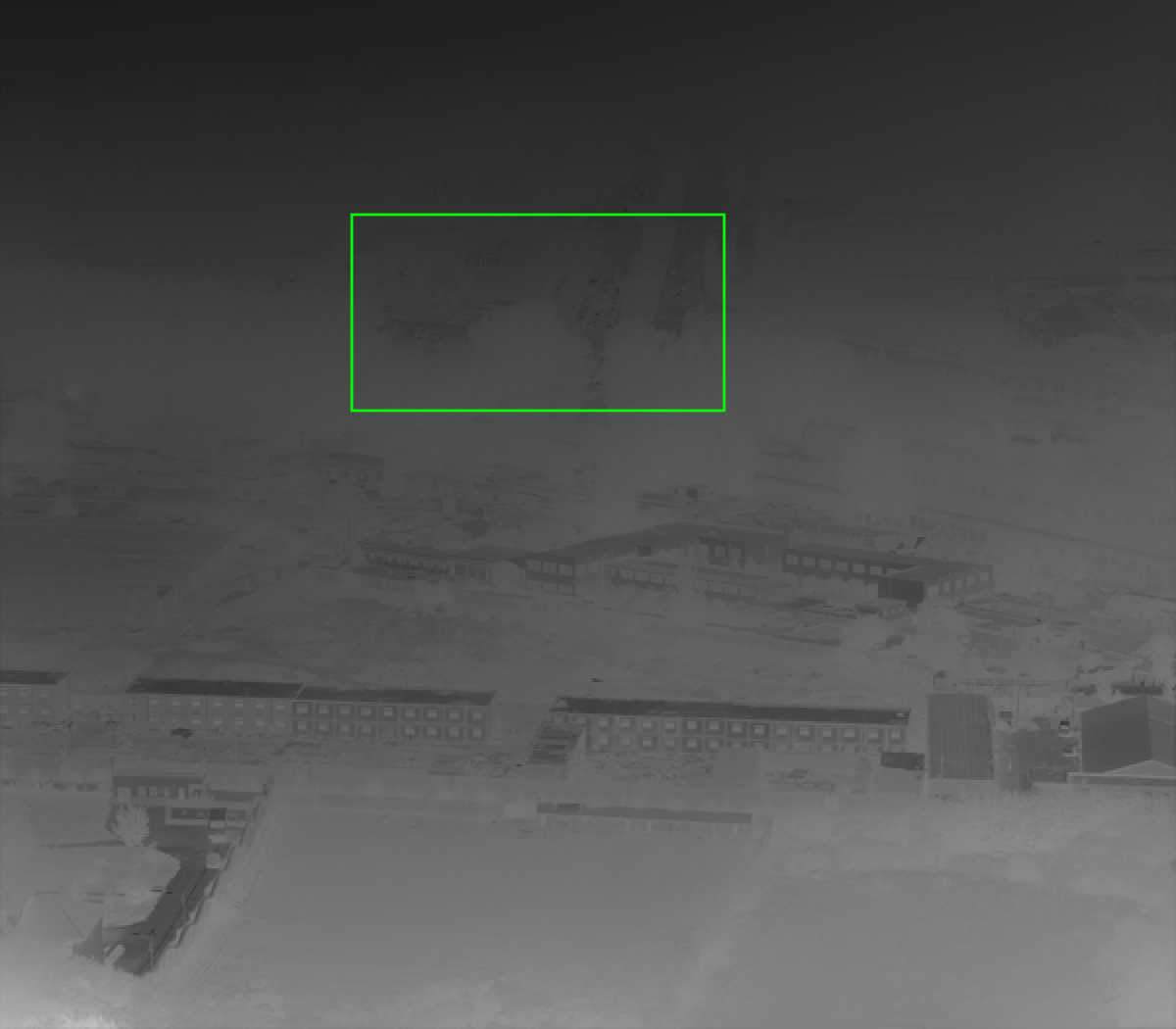}
	\includegraphics[width=0.15\linewidth]{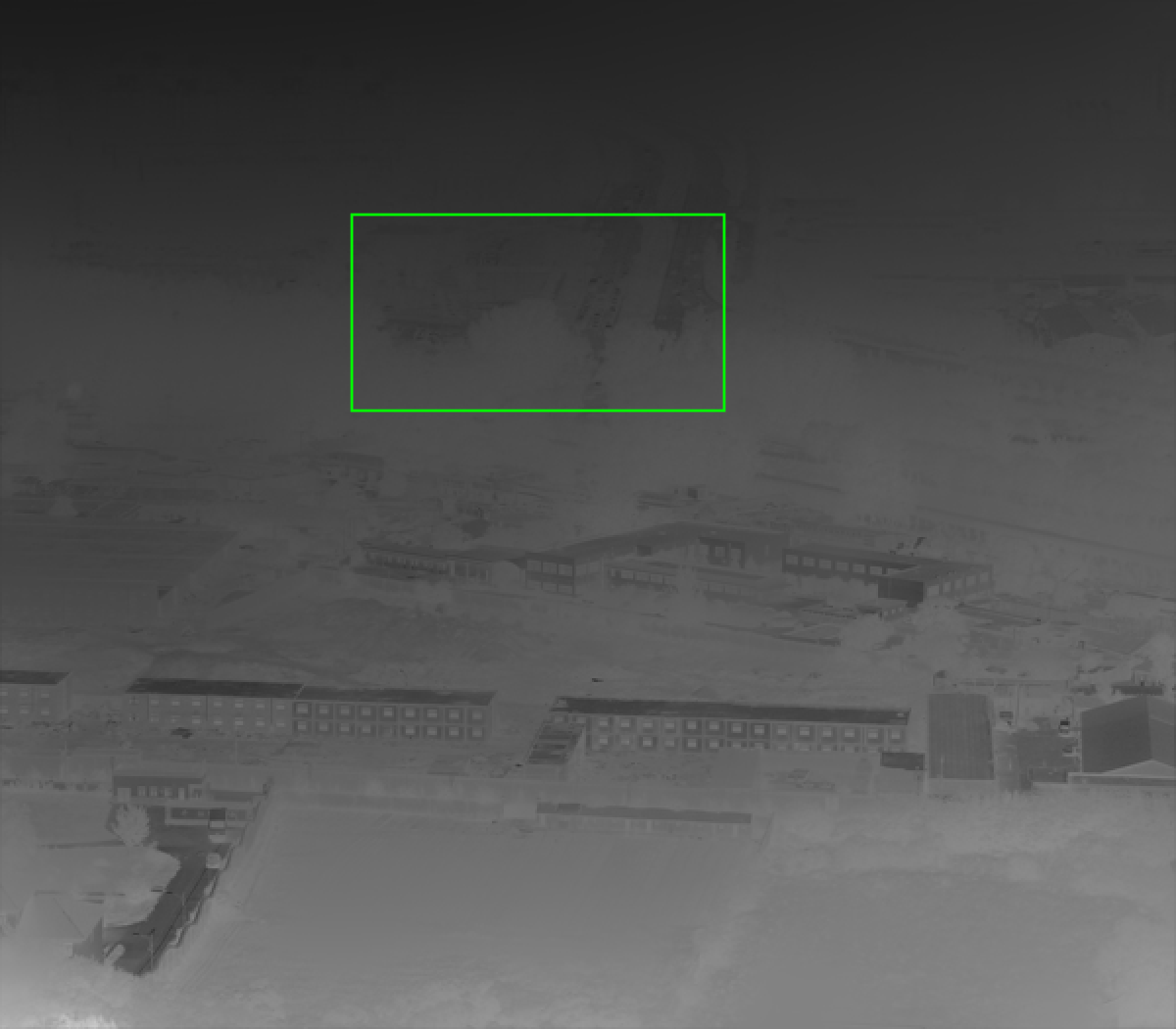}
	\includegraphics[width=0.15\linewidth]{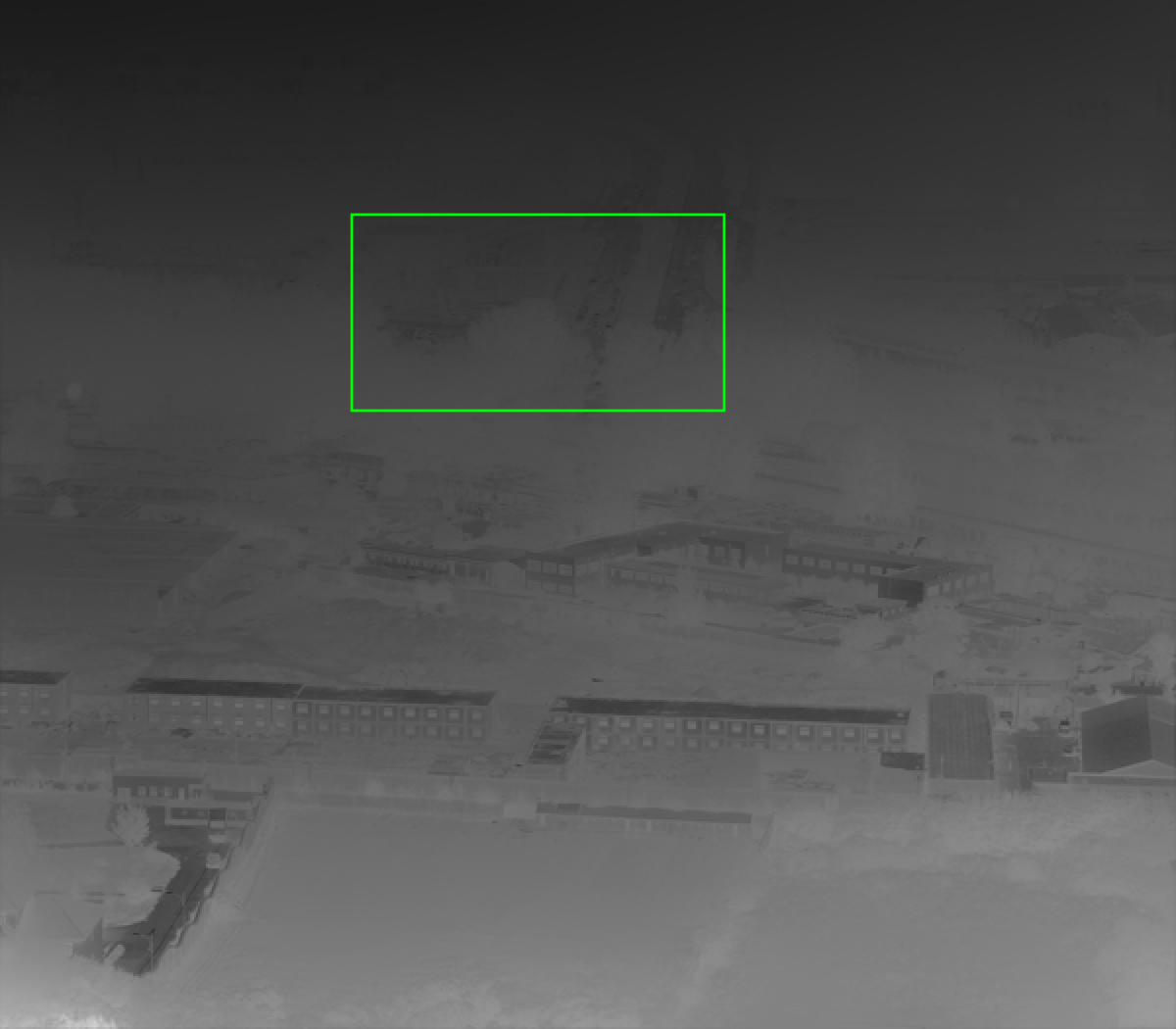}
	
	\hspace{0.15\linewidth}
	\subfigure[]{\includegraphics[width=0.15\linewidth]{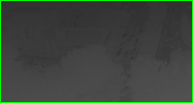}}
	\subfigure[]{\includegraphics[width=0.15\linewidth]{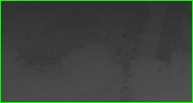}}
	\subfigure[]{\includegraphics[width=0.15\linewidth]{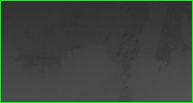}}
	\subfigure[]{\includegraphics[width=0.15\linewidth]{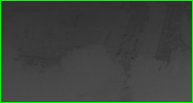}}
	\subfigure[]{\includegraphics[width=0.15\linewidth]{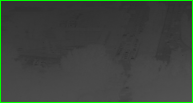}}

	\caption{Comparison of raw transmission map and  refined ones by different filters.  (a) Raw map.  Refined maps: (b)-(f) GIF, WGIF, GGIF, SKWGIF, RDWGIF. (g)-(k) GH-GIF, GH-WGIF, GH-GGIF, GH-SKWGIF, GH-RDWGIF.}		
	\label{fig:4.6-1}
\end{figure}

\begin{figure}[!htbp]
	\centering
	\subfigure[]{\includegraphics[width=0.15\linewidth]{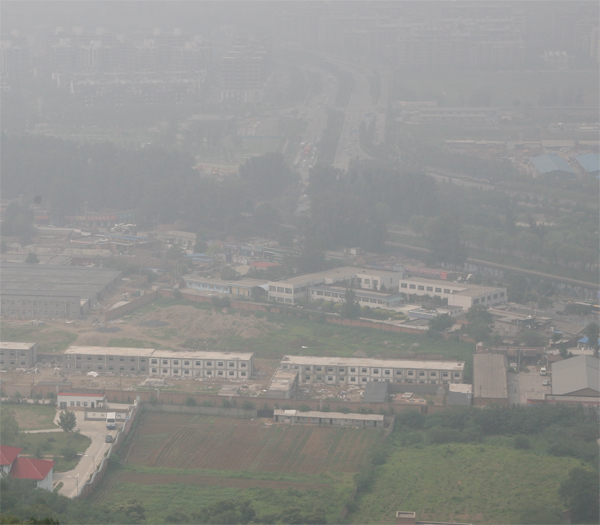}}
	\subfigure[]{\includegraphics[width=0.15\linewidth]{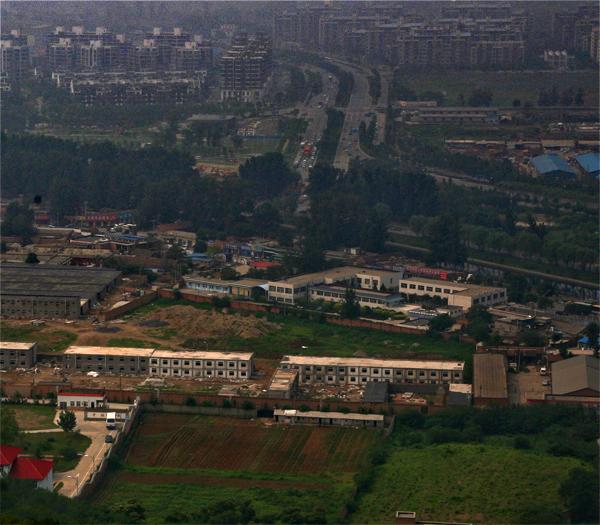}}
	\subfigure[]{\includegraphics[width=0.15\linewidth]{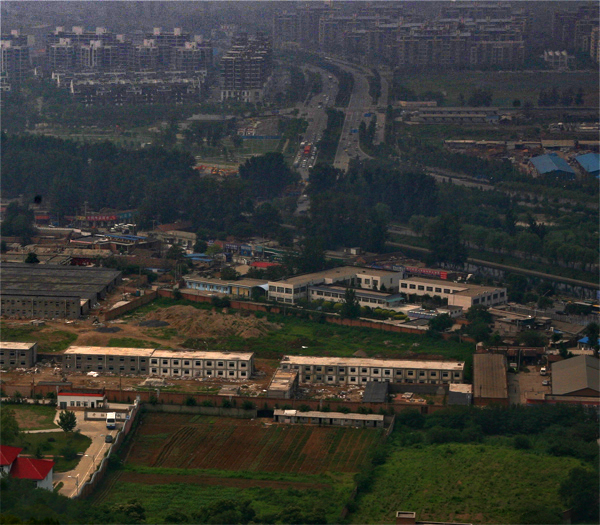}}
	\subfigure[]{\includegraphics[width=0.15\linewidth]{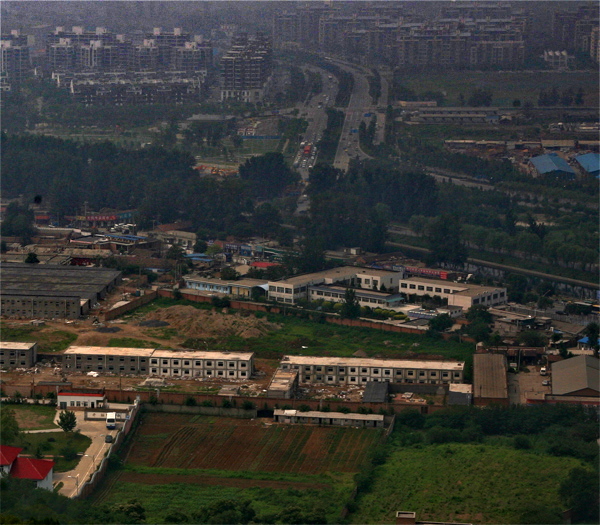}}
	\subfigure[]{\includegraphics[width=0.15\linewidth]{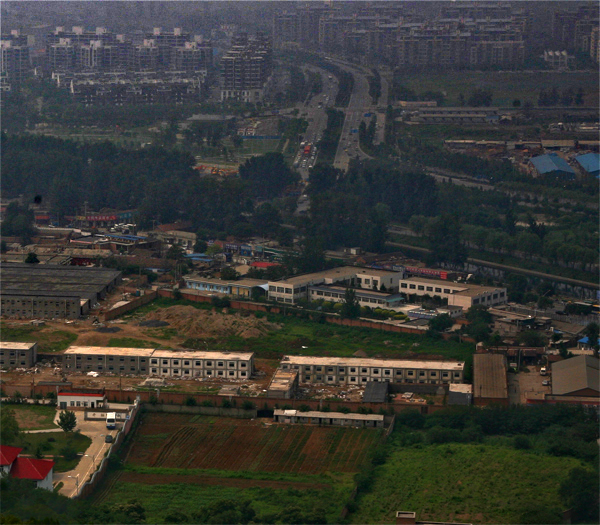}}	
	\subfigure[]{\includegraphics[width=0.15\linewidth]{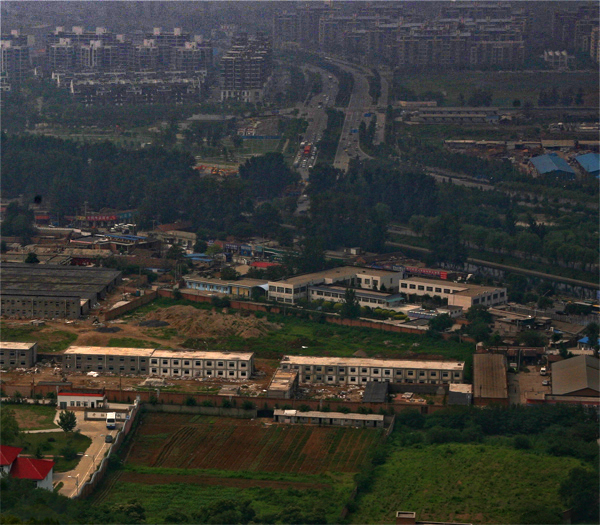}}
	
	\hspace{0.15\linewidth} \subfigure[]{\includegraphics[width=0.15\linewidth]{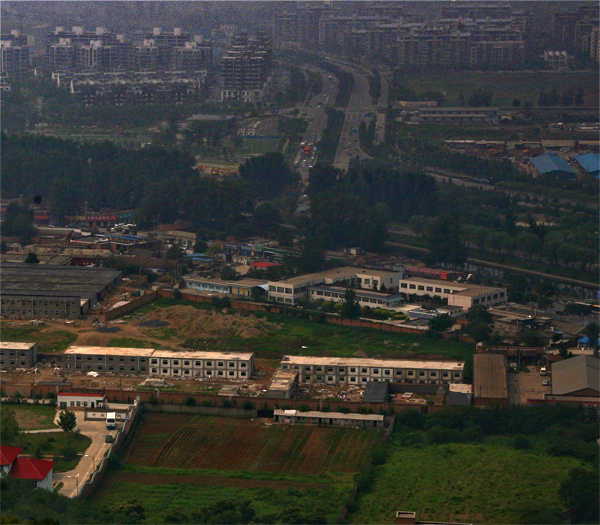}}
	\subfigure[]{\includegraphics[width=0.15\linewidth]{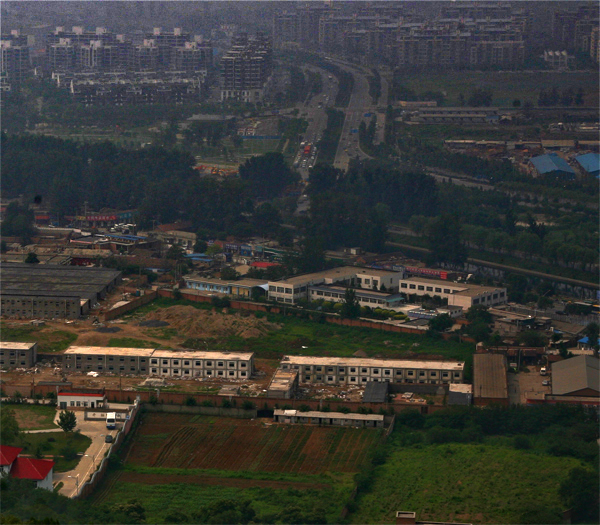}}
	\subfigure[]{\includegraphics[width=0.15\linewidth]{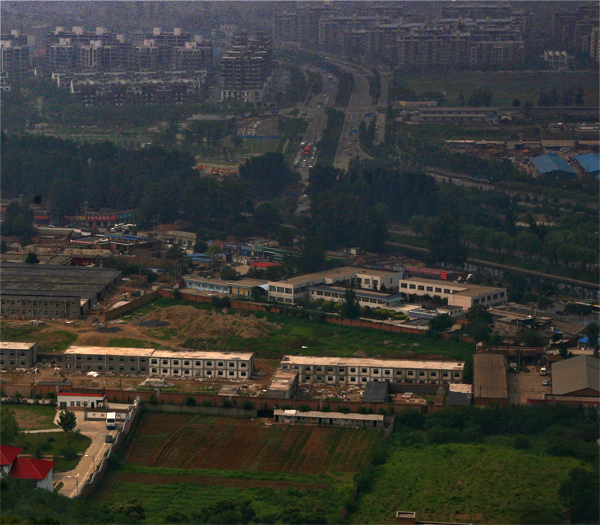}}
	\subfigure[]{\includegraphics[width=0.15\linewidth]{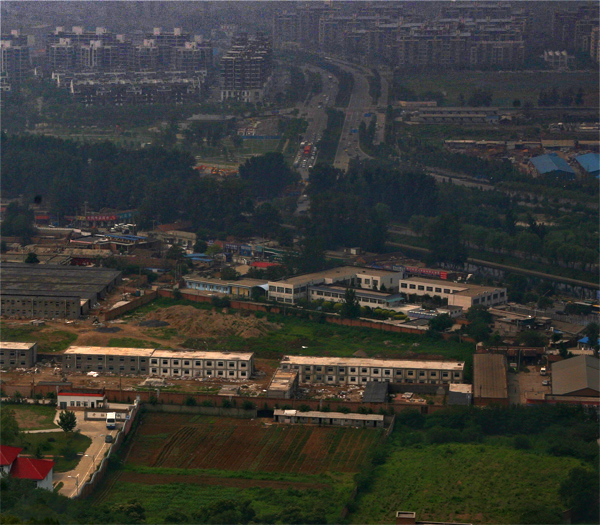}}	
	\subfigure[]{\includegraphics[width=0.15\linewidth]{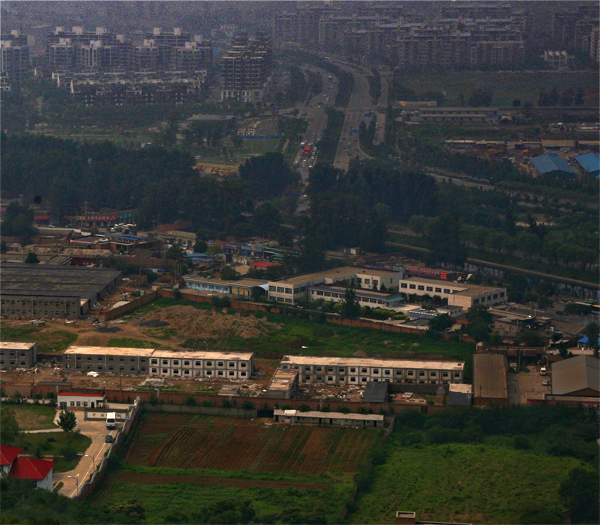}}
	\caption{Comparison of dehazing results by different filters.  (a) Hazy image. Dehazed images: (b)-(f) GIF, WGIF, GGIF, SKWGIF, RDWGIF. (g)-(k) GH-GIF, GH-WGIF, GH-GGIF, GH-SKWGIF, GH-RDWGIF.}
	\label{fig:4.5-1}
\end{figure}

As shown in the local magnification of Fig. \ref{fig:4.6-1}, the transmission map refined by PM-GF-based GIFs exhibits clearer edges and richer detail features. The results of the experiment are displayed in Fig. \ref{fig:4.5-1} , which show the hazy image and the corresponding dehazed ones. Subjective evaluations reveal that the guided filters based on PM-GF exhibit better visual effects in haze removal, compared to the GIFs based on LAM.

\subsection{Image texture removal smoothing}
Texture removal smoothing is expressed as the problem of seeking a suitable structure for an observed image by means of removing texture detail. The objective is to suppress small-scale structures (even when they include strong edges) and meanwhile, to preserve large-scale structures (even when they contain weak edges). The techniques for texture removal smoothing are thus different from smoothing ones aiming at preserving strong edges.

Zhang et al. \cite{zhang2014rolling} introduced a technique for texture removal smoothing, termed rolling guidance filter (RGF), which smooths small structures of different scales in an iterative way. The RGF initially removes small structures in the image by Gaussian filtering. Subsequently, the filtering output undergoes iterative refinement via joint bilateral filtering \cite{yu2009image}. The RGF can be regarded as a rolling-guided framework, in which the used bilateral filtering can be optionally substituted with other edge-preserving filtering techniques, such as the GIFs.

\begin{figure}[!htbp]
	\centering
	\subfigure[]{\includegraphics[width=0.15\linewidth]{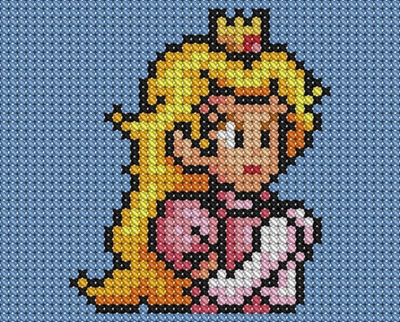}}
	\subfigure[]{\includegraphics[width=0.15\linewidth]{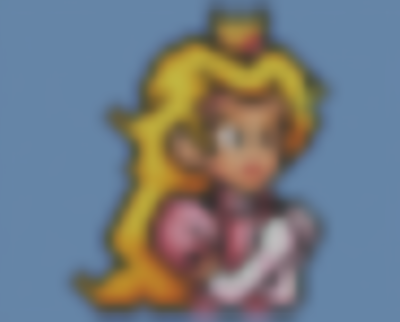}}
	\subfigure[]{\includegraphics[width=0.15\linewidth]{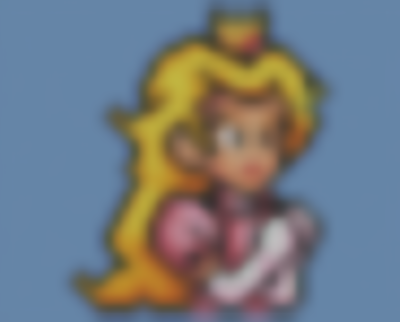}}
	\subfigure[]{\includegraphics[width=0.15\linewidth]{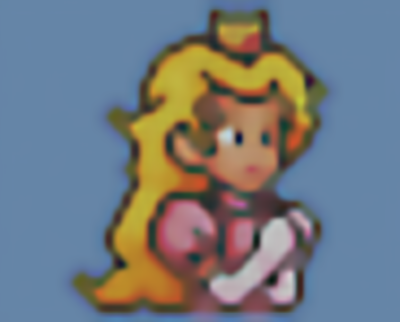}}
	\subfigure[]{\includegraphics[width=0.15\linewidth]{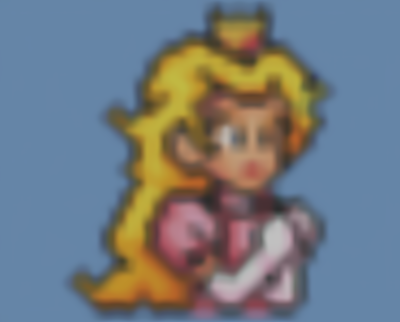}}	
	\subfigure[]{\includegraphics[width=0.15\linewidth]{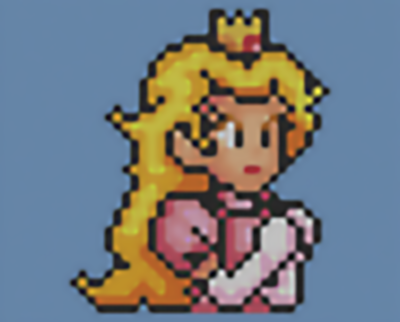}}
	
	\hspace{0.15\linewidth} 	\subfigure[]{\includegraphics[width=0.15\linewidth]{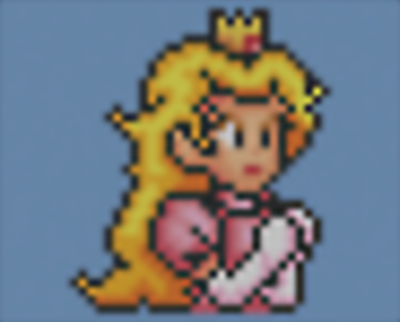}}
	\subfigure[]{\includegraphics[width=0.15\linewidth]{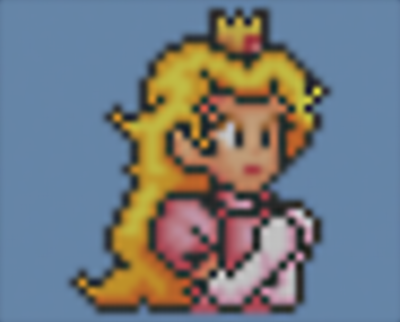}}
	\subfigure[]{\includegraphics[width=0.15\linewidth]{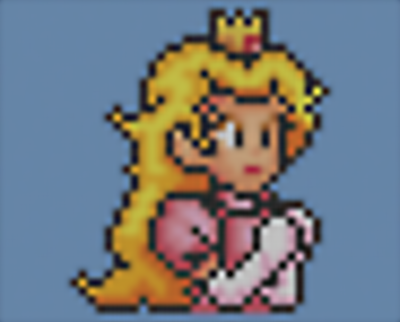}}
	\subfigure[]{\includegraphics[width=0.15\linewidth]{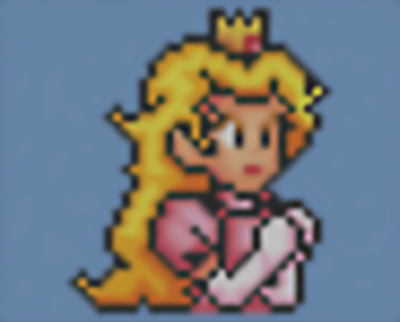}}	
	\subfigure[]{\includegraphics[width=0.15\linewidth]{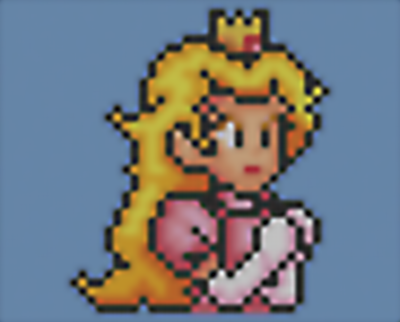}}
	\caption{Comparison of texture removal smoothing results by different filters. (a) Original image. Texture removal smoothed results: (b)-(f) GIF, WGIF, GGIF, SKWGIF, RDWGIF. (g)-(k) GH-GIF, GH-WGIF, GH-GGIF, GH-SKWGIF, GH-RDWGIF.}
	\label{fig:4.7}
\end{figure}

\begin{figure}[!htbp]
	\centering
	\subfigure[]{\includegraphics[width=0.15\linewidth]{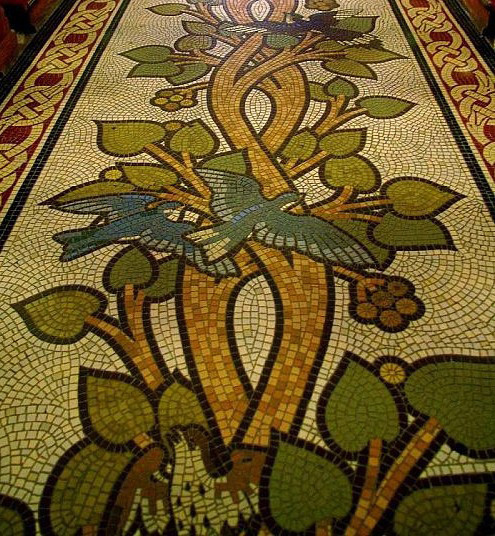}}
	\subfigure[]{\includegraphics[width=0.15\linewidth]{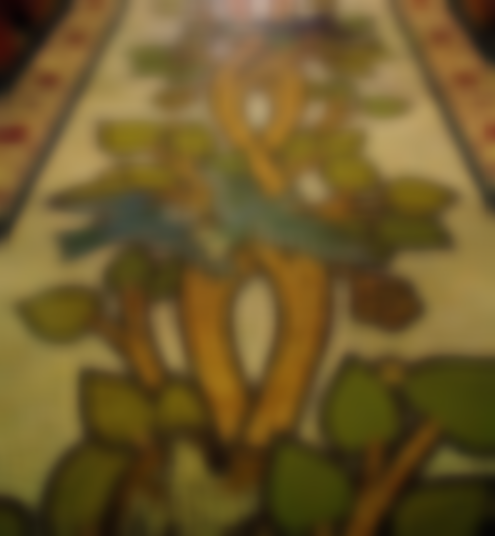}}
	\subfigure[]{\includegraphics[width=0.15\linewidth]{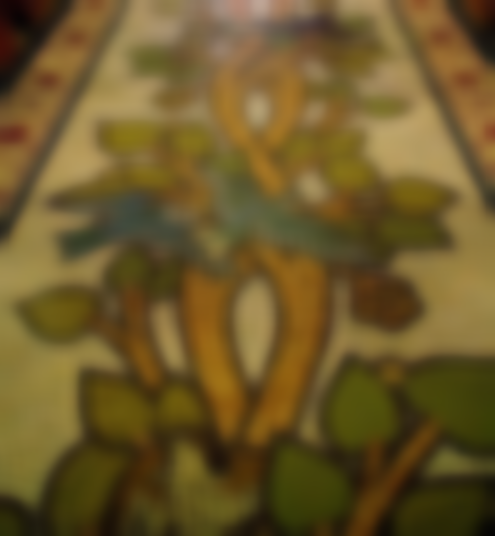}}
	\subfigure[]{\includegraphics[width=0.15\linewidth]{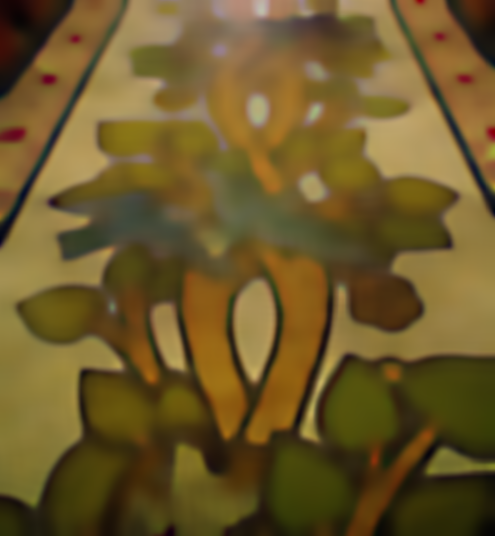}}
	\subfigure[]{\includegraphics[width=0.15\linewidth]{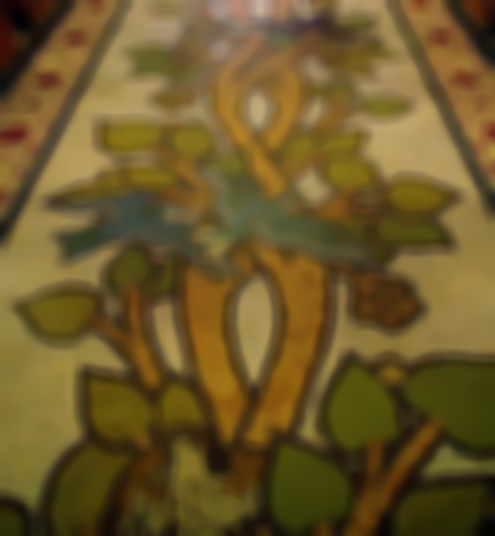}}	
	\subfigure[]{\includegraphics[width=0.15\linewidth]{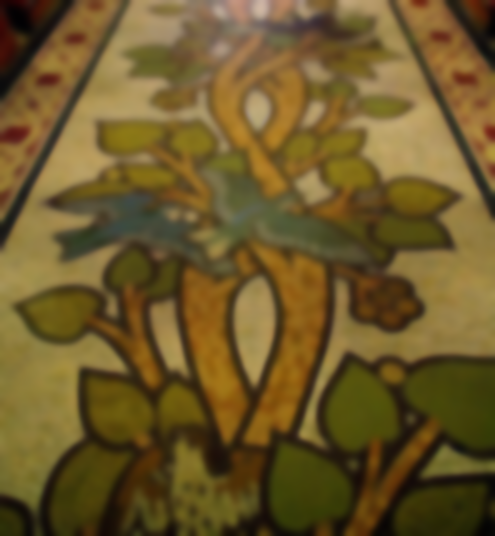}}
	
	\hspace{0.15\linewidth} \subfigure[]{\includegraphics[width=0.15\linewidth]{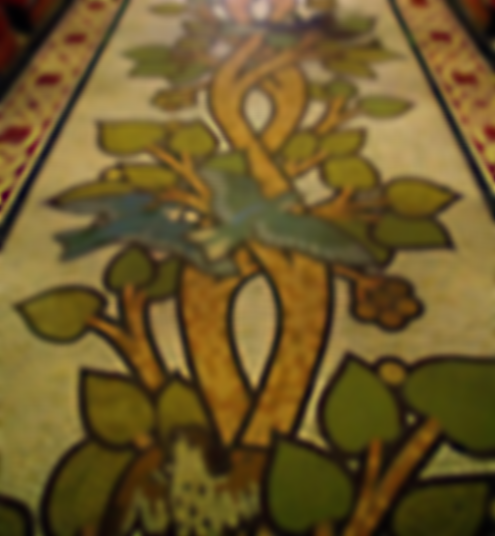}}
	\subfigure[]{\includegraphics[width=0.15\linewidth]{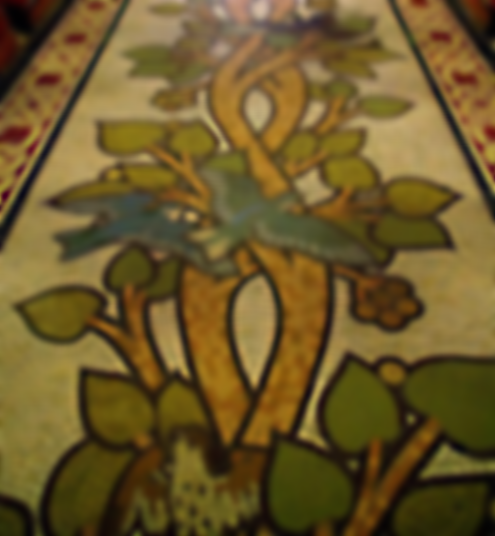}}
	\subfigure[]{\includegraphics[width=0.15\linewidth]{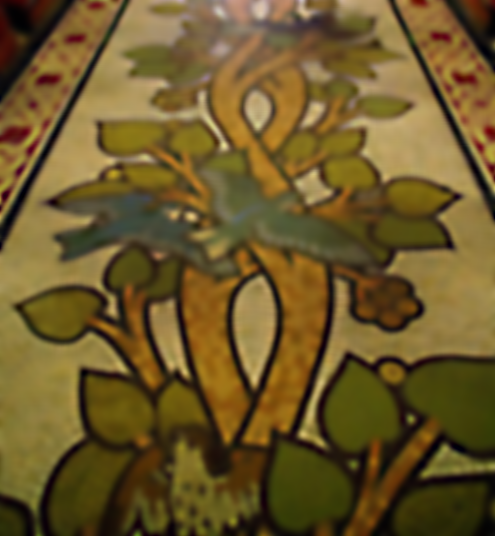}}
	\subfigure[]{\includegraphics[width=0.15\linewidth]{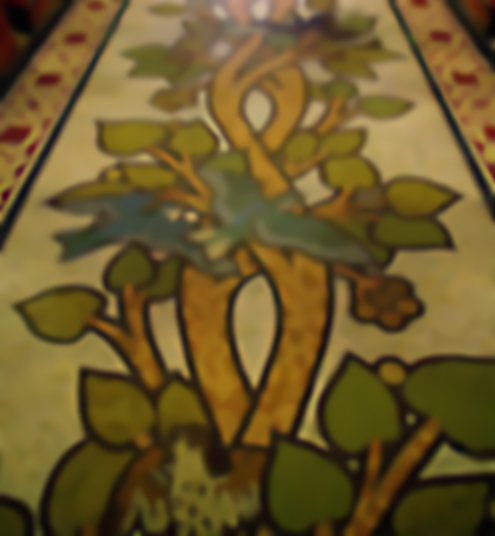}}	
	\subfigure[]{\includegraphics[width=0.15\linewidth]{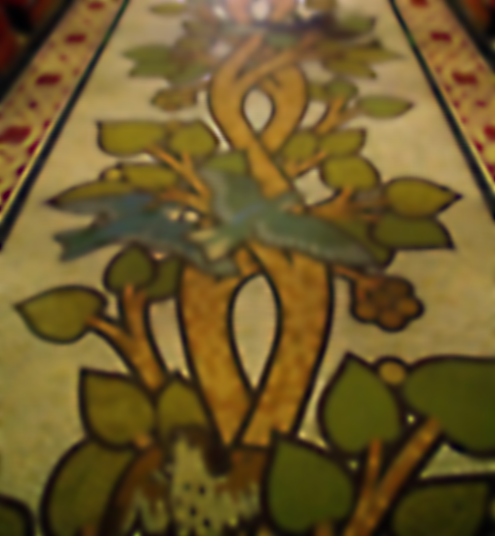}}
	\caption{Comparison of texture removal smoothing results by different filters. (a) Original image. Texture removal smoothed results: (b)-(f) GIF, WGIF, GGIF, SKWGIF, RDWGIF. (g)-(k) GH-GIF, GH-WGIF, GH-GGIF, GH-SKWGIF, GH-RDWGIF.}
	\label{fig:4.8}
\end{figure}

In this experiment, we replace joint bilateral filtering in RGF with GIFs based on LAM and PM-GF to form ten variants of RGF. We set $r=8$ and $\varepsilon=0.2^2$ for ten guided filters, following RDWGIF \cite{zhang2022robust}. The smoothing results for ten new RGFs  are obtained after five iterations. In Fig. \ref{fig:4.7} and Fig. \ref{fig:4.8}, we display the texture removal smoothing results for two images with textures. It can be observed that RGFs with PM-GF-based GIFs achieve a higher performance in both removing texture and preserving structure, compared to RGFs with LAM-based GIFs.

\section{Conclusion}
In this paper, we introduce a novel prior model (i.e., PM-GF) with a single parameter instead of the original local affine model (LAM) with two parameters for guided filtering. In PM-GF, the filtering output in a local window is obtained by adding a weighted portion of the filtered guidance image with a Gaussian highpass ﬁlter to a Gaussian lowpass ﬁltered version of the input image, where the weight coefﬁcient is the unique parameter to be estimated for deciding the extent to which edges are preserved. Therefore, PM-GF gives an explicit interpretation of the mechanism of transferring the desired structure from the guidance image to the filtering output. Based on PM-GF, we propose the Gaussian highpass guided image filter (GH-GIF) as well as several new guided filters, derived from the existing guided filters by utilizing PM-GF instead of LAM used in these filters. Finally, a large number of experiments are conducted to compare the performance of the new guided filters and their counterparts in six real applications. Our observations are that PM-GF-based guided filters all outperform the LAM-based guided filters in terms of subjective assessment and objective evaluation.

The PM-GF-based guided filters can smooth the image while well preserving primary image structures; it indicates their great application potentials in many tasks, which can be served as the pre-processing or post-processing techniques to enhance the primary image information (main structures and edges), thereby improving the accuracy. In addition, our new filters only involve a single parameter (weight coefﬁcient) in the formula for the final output, which can be learned by a single network adaptively, thereby being more suitable for joint/guided filtering with deep convolutional networks \cite{li2019joint}, compared to the LAM-based guided filters. 

However, like existing guided filters, our filters are also implicitly based on a priori assumption that presumes the guidance and input images share the identical structural information, without accounting for potential inconsistencies in edges. Accordingly, when there are inconsistent structures between both images, our filters would introduce substantial errors by directly transferring the structure from the guidance image to the Gaussian smoothed version of the input image.
 
In the future work, we shall pay attention to deep guided filtering, attempting to learn adaptively the weight coefﬁcient in the formula for the final output by means of deep learning approaches for improving PM-GF-based GIFs. Moreover, we shall explore our new filters to broaden the application.

\newpage
\bibliographystyle{elsarticle-num} 
\bibliography{reference}

\end{document}